\documentclass{article}

\usepackage[nonatbib, final]{neurips_2025}

\usepackage[
    verbose,
    version=5,
    shorthands,hidemsgs,
    savespace,
]{preamble}

\usepackage{placeins}
\usepackage{annotate-equations}

\usepackage{algorithm}
\usepackage{algorithmic}

\everypar=\expandafter{\the\everypar\looseness=-1 }
\title{DeCaFlow: A deconfounding causal generative model}

\author{%
  \textbf{Alejandro Almod\'ovar}\textsuperscript{1, *} \quad
  \textbf{Adri\'an Javaloy}\textsuperscript{2,*} \\ [2ex]
  \textbf{Juan Parras}\textsuperscript{1} \quad
  \textbf{Santiago Zazo}\textsuperscript{1} \quad
  \textbf{Isabel Valera}\textsuperscript{3} \\ [4ex]
  \textsuperscript{1}Information Processing and Telecommunications Center, Universidad Polit\'ecnica de Madrid, ES \\
  \textsuperscript{2}School of Informatics, University of Edinburgh, UK \\
  \textsuperscript{3} Department of Computer science, Saarland University, DE \\
}

\newcommand{\ExternalLink}{%
	\tikz[x=1.2ex, y=1.2ex, baseline=-0.05ex]{%
		\begin{scope}[x=1ex, y=1ex]
			\clip (-0.1,-0.1) 
			--++ (-0, 1.2) 
			--++ (0.6, 0) 
			--++ (0, -0.6) 
			--++ (0.6, 0) 
			--++ (0, -1);
			\path[draw, 
			line width = 0.9, 
			rounded corners=0.5] 
			(0,0) rectangle (1,1);
		\end{scope}
		\path[draw, line width = 0.9] (0.5, 0.5) 
		-- (1, 1);
		\path[draw, line width = 0.9] (0.6, 1) 
		-- (1, 1) -- (1, 0.6);
	}
}
\newcommand{\urlicon}[1]{\href{#1}{\large\ExternalLink}}

\begin{document}
	\maketitle

    \begingroup
    \renewcommand\thefootnote{*}
    \footnotetext{Equal contribution. Correspondence to \href{mailto:<alejandro.almodovar@upm.es>}{\methodname{alejandro.almodovar@upm.es}} and \href{mailto:<ajavaloy@ed.ac.uk>}{\methodname{ajavaloy@ed.ac.uk}}.}
    \endgroup
	
    \doparttoc %
	\faketableofcontents %

\begin{abstract} 
    We introduce \ours, a deconfounding causal generative model. %
    Training once per dataset using just observational data and the underlying causal graph, \ours enables accurate causal inference on continuous variables under the presence of hidden confounders.  
    Specifically, we extend previous results on causal estimation under hidden confounding to show that a single instance of \ours provides correct estimates for all causal queries identifiable with do-calculus,
    leveraging proxy variables to adjust for the causal effects when do-calculus alone is insufficient.
    Moreover, we show that counterfactual queries are identifiable as long as their interventional counterparts are identifiable, and thus are also correctly estimated by \ours. %
    Our empirical results on diverse settings---including the Ecoli70 dataset, with \num{3} independent hidden confounders, tens of observed variables and hundreds of causal queries---show that \ours outperforms existing approaches, while demonstrating its out-of-the-box applicability to any given causal graph.

\end{abstract}

    \section{Causal generative models and hidden confounding}
\label{sec:intro}
Causal inference %
seeks to determine how changes in one variable affect others, which is crucial to evaluate the effects of interventions in fields such as healthcare~\citep{feuerriegel2024causal}, marketing policing~\citep{varian2016causal} or education~\citep{zhao2017estimating}. 
In real-world scenarios, where empirical trials often are infeasible due to ethical, financial, or practical constraints, answering causal queries from observational data becomes essential.
However, this is a challenging task, especially in the presence of  unmeasured or hidden confounders affecting a subset of the observed variables~\citep{greenland1996basic, adams2021identification}.

In this work, we aim to propose a practical approach for accurate causal inference on continuous variables under the presence of hidden confounders. %
To this end,  we build on two key concepts: %
\itemi~\emph{causal generative  models} (CGMs)~\citep{javaloy2024causal,chao2023interventional,khemakhem2021causal}, a class of generative models that  can generate samples not only from the observational distribution, but also from interventional and (in some cases) counterfactual distributions; and \itemii~\emph{proxy variables}, \ie, conditionally independent variables that yield information about the hidden confounders~\citep{miao2018identifying, miao2023identifying, wang2021proxy, miao2023identifying}.
Consequently, we introduce the
\emph{\underline{de}confounding \underline{ca}usal normalizing \underline{flow}} (\ours), a CGM which provides correct estimates of a broad class of interventional and counterfactual queries under hidden confounding, requiring only observational data, the causal graph, and training once per dataset.
Architecturally, \ours resembles variational autoencoders \citep{kingma2013auto} as it is trained with the ELBO and comprises: \itemi~a causal normalizing flow (CNF)~\citep{javaloy2024causal} as ``decoder'', adapted to be conditioned on the (potentially many) hidden confounders; and \itemii a conditional normalizing flow~\citep{winkler2019learning} as ``encoder'', computing the modeled posterior distribution of the hidden confounders given the observations.

We proved theoretically that
\emph{\ours yields correct estimates for both interventional and counterfactual queries that  are identifiable with do-calculus,
leveraging the information of proxy variables when do-calculus alone is insufficient.}
To that end, we first extend recent advances in proximal causal inference by \citet{miao2018identifying} and \citet{wang2021proxy} to
include
counterfactual causal queries. 
Then, we integrate proximal-identifiability with do-calculus, expanding
the number of identifiable queries of which \ours is shown to provide correct estimates.

\begin{wrapfigure}[19]{R}{0.51\linewidth}
	\centering
    \vspace{-10pt}
	    \pgfdeclarelayer{background}
    \pgfdeclarelayer{foreground}
    \pgfsetlayers{background,main,foreground}

  \begin{tikzpicture}
  \tikzset{mynode/.style={style=fairnessobs, fill=white}}
    \tikzset{hnode/.style={style=fairnesshidden}}
      \draw
        (-2.0, 0) node[hnode] (b1191){\methodname{b1191}} %
        (-2.4, -1) node[mynode] (fixC){\methodname{fixC}} %
        (-1.75, -2) node[mynode] (traA){\methodname{traA}} (-2.25, -2)
        (2, -2) node[mynode] (ygcE){\methodname{ygcE}} %
        (2.0, 0) node[hnode] (eutG){\methodname{eutG}} %
        (-3.75, -2) node[mynode] (yceP){\methodname{yceP}}
        (-4.0, -3) node[mynode] (ibpB){\methodname{ibpB}}
        (-2.0, -3) node[mynode] (yfaD){\methodname{yfaD}}
        (-2.5, -5) node[mynode] (lacY){\methodname{lacY}}
        (2.5, -1) node[mynode] (sucA){\methodname{sucA}} %
        (-5.15, -2) node[mynode] (ygbD){\methodname{}}%
        (-5.30, -2) node[mynode] (yjbO){\methodname{}}%
        (-5.45, -2) node[mynode] (cchB){\methodname{cchB}}
        (2.0, -6) node[mynode] (ycgX){\methodname{ycgX}}
        (4.75, -2) node[mynode] (dnaJ){\methodname{}}
        (4.90, -2) node[mynode] (flgD){\methodname{}}
        (5.05, -2) node[mynode] (gltA){\methodname{}}
        (5.20, -2) node[mynode] (sucD){\methodname{}}%
        (5.35, -2) node[mynode] (yhdM){\methodname{}}%
        (4.0, -3) node[mynode] (atpD){\methodname{atpD}}
        (5.5, -2) node[mynode] (atpG){\methodname{atpG}} %
        (-5.0, -7) node[mynode] (b1583){\methodname{b1583}}
        (-5.0, -4) node[mynode] (icdA){\methodname{icdA}}
        (0.4, -3) node[mynode] (asnA){\methodname{asnA}}
        (-5.5, -6) node[mynode] (lacZ){\methodname{lacZ}}
        (-3.25, -4) node[mynode] (lacA){\methodname{lacA}}
        (2.15, -3) node[hnode](cspG){\methodname{cspG}}
        (-3.0, -7) node[mynode] (yaeM){\methodname{yaeM}}
        (-1.0, -4) node[mynode] (cspA){\methodname{cspA}}
        (1.5, -4) node[mynode] (yeoC){\methodname{yeoC}}
        (0.0, -6) node[mynode] (pspA){\methodname{pspA}}
        (3.0, -5) node[mynode] (pspB){\methodname{pspB}} %
        (3.3, -4) node[mynode] (yedE){\methodname{yedE}}
        (5.0, -5) node[mynode] (yhel){\methodname{yhel}}
        (-5.5, -5) node[mynode] (aceB){\methodname{aceB}}
        (-2.4, -6) node[mynode] (hupB){\methodname{hupB}}
        (-1.1, -5) node[mynode] (yfiA){\methodname{yfiA}}
        (0.8, -5) node[mynode] (lpdA){\methodname{lpdA}} %
        (-3.7, -6) node[mynode] (nuoM){\methodname{nuoM}}
        (5.0, -7) node[mynode] (dnaG){\methodname{dnaG}}
        (5.5, -4) node[mynode] (b1963){\methodname{b1963}}
        (3.6, -6) node[mynode] (folK){\methodname{folK}}
        (5.6, -6) node[mynode] (dnaK){\methodname{dnaK}}
        (-1.0, -7) node[mynode] (mopB){\methodname{mopB}}
        (3.0, -7) node[mynode] (nmpC){\methodname{nmpC}}
        (1.0, -7) node[mynode] (ftsJ){\methodname{ftsJ}};
      \begin{scope}[->, style={>={Stealth[scale=0.8]}}]
        \begin{pgfonlayer}{background}
            \draw [hiddenarrow] (b1191) to (fixC);
            \draw [hiddenarrow] (b1191) to (traA);
            \draw [hiddenarrow] (b1191) to (ygcE);
            \draw (fixC) to[bend right=10] (ygbD);
            \draw (fixC) to[bend right=10] (yjbO);
            \draw (fixC) to[bend right=10] (cchB);
            \draw (fixC) to (yceP);
            \draw [style=nonidentifiable](fixC) to (traA);
            \draw (fixC) to[bend left=10] (ycgX);
            \draw (ygcE) to (icdA);
            \draw (ygcE) to (asnA);
            \draw (ygcE) to (atpD);
            \draw [dashed, draw=gray] (eutG) to (yceP);
            \draw [dashed, draw=gray] (eutG) to[bend right=10] (ibpB);
            \draw [dashed, draw=gray] (eutG) to (yfaD);
            \draw [dashed, draw=gray] (eutG) to (lacY);
            \draw [dashed, draw=gray] (eutG) to (sucA);
            \draw (yceP) to (b1583);
            \draw [style=nonidentifiable](yceP) to (ibpB);
            \draw [style=identifiable](yceP) to (yfaD);
            \draw (lacY) to (lacZ);
            \draw (lacY) to (nuoM);
            \draw (sucA) to (traA);
            \draw [style=nonidentifiable] (sucA) to (yfaD);
            \draw (sucA) to (ygcE);
            \draw (sucA) to[bend left=10] (flgD);
            \draw (sucA) to[bend left=10] (gltA);
            \draw (sucA) to[bend left=10]  (sucD);
            \draw (sucA) to[bend left=10] (yhdM);
            \draw (sucA) to[bend left=10] (atpD);
            \draw (sucA) to[bend left=10] (atpG);
            \draw (sucA) to[bend left=10]  (dnaJ);
            \draw (ycgX) to[bend left=0] (dnaG);
            \draw (atpD) to (yhel);
            \draw (icdA) to (aceB);
            \draw (asnA) to (icdA);
            \draw (asnA) to[bend right=5] (lacZ);
            \draw (asnA) to (lacA);
            \draw [style=identifiable](asnA) to[bend left=23] (lacY);
            \draw (lacZ) to (b1583);
            \draw [style=identifiable](lacZ) to (yaeM);
            \draw (lacZ) to (mopB);
            \draw (lacA) to (b1583);
            \draw [style=identifiable](lacA) to (yaeM);
            \draw (lacA) to (lacZ);
            \draw [style=identifiable](lacA) to (lacY);
            \draw [dashed, draw=gray] (cspG) to (lacA);
            \draw [dashed, draw=gray] (cspG) to[bend left=5] (lacY);
            \draw [dashed, draw=gray] (cspG) to (yaeM);
            \draw [dashed, draw=gray] (cspG) to (cspA);
            \draw [dashed, draw=gray] (cspG) to (yeoC);
            \draw [dashed, draw=gray] (cspG) to[bend right=25] (pspA);
            \draw [dashed, draw=gray] (cspG) to (pspB);
            \draw [dashed, draw=gray] (cspG) to (yedE);
            \draw (cspA) to (hupB);
            \draw (cspA) to (yfiA);
            \draw (pspA) to (nmpC);
            \draw [style=identifiable](pspB) to (pspA);
            \draw [style=identifiable](yedE) to (pspB);
            \draw [style=identifiable](yedE) to (pspA);
            \draw (yedE) to (lpdA);
            \draw (yedE) to (yhel);
            \draw (yhel) to (ycgX);
            \draw (yhel) to (dnaG);
            \draw (yhel) to (b1963);
            \draw (yhel) to (folK);
            \draw (yhel) to (dnaK);
            \draw (yfiA) to (hupB);
            \draw (dnaK) to (mopB);
            \draw (mopB) to (ftsJ);
        \end{pgfonlayer}
      \end{scope}
    \end{tikzpicture}
	
	\caption{\textbf{\Ours can be effortlessly applied to highly complex causal graphs},  as that of the Ecoli70 dataset \citep{schafer2005ecoli}, with multiple  hidden confounders and dozens of variables.
		We dash \textcolor{hiddencolor}{hidden confounders}, and highlight direct \emph{hidden-confounded} effects as \textcolor{idcolor}{identifiable} (and thus correctly estimated by \ours), or \textcolor{secondcolor}{unidentifiable}.%
	}
	\label{fig:ecoli_graph_custom}
\end{wrapfigure}
As proof of the flexibility that \ours offers, \cref{fig:ecoli_graph_custom} illustrates the causal graph of the Ecoli70 dataset~\citep{schafer2005ecoli}, comprising 43 observed variables and 3 hidden confounders, showing that \ours can effortlessly scale to complex settings and accurately recover diverse causal effects after a single training process.
Remarkably, green edges in the figure represent direct causal effects that \ours can identify, despite the  presence of hidden confounders.
 We additionally, provide algorithms to help practitioners easily check in the given the causal graph whether a particular query of interest can be correctly estimated by \ours.

Finally, 
we empirically validate all our claims on semi-synthetic and real-world experiments, demonstrating that \ours outperforms existing alternatives while being widely applicable. 
An implementation of \ours can be found in \href{https://github.com/aalmodovares/DeCaFlow}{\methodname{github.com/aalmodovares/DeCaFlow}}.

    \subsection{Related works}
\label{sec:related-work}
 We briefly discuss relevant works in the literature, and defer the reader to \cref{app:sec:related-work} for further details.

\paragraph{Causal generative models.} As mentioned above, we refer as CGMs to the class of generative models that can generate samples from the observational,  interventional and, in some cases, counterfactual distributions. 
A  common recipe to build causally consistent CGMs consists of modeling each variable as a function of its causal parents with an independent model. %
In terms of the choice for modeling these functions, prior works range from simple yet well-established additive noise models~\citep{hoyer2008nonlinear}, to more complex but powerful diffusion-based causal models~\citep{chao2023interventional}, among others~\citep{rahman2024modular,kocaoglu2017causalgan,yang2020causalvae,pawlowski2020deep, parafita2022estimand}.
Due to their sequential nature, these approaches %
can overfit and propagate errors to downstream variables.
Alternatively,
recent works have explored using a single (structurally-constrained) network to model the SCM at once, \eg, 
using normalizing flows~\citep{khemakhem2021causal, javaloy2024causal}, or graph neural networks~\citep{zevcevic2021relating, sanchez2022vaca}.
However, all the aforementioned approaches  assume \textit{causal sufficiency}, \ie the absence of hidden confounders, limiting their applicability in settings with hidden confounding. 

\paragraph{Causal inference hidden  confounding.}
When dealing with hidden confounding, many  approaches  handle only interventional queries and are tailored to a specific causal graph and a single treatment-outcome pair, requiring us to train one model for each query we want to answer.
Prior works exploit instrumental variables~\citep{angrist2009mostly},  mediators~\citep{pearl2009causality}, and, more recently, 
proxy variables to account for hidden confounding~\citep{allman09, kuroki2014measurement, kallus2018causal,miao2023identifying, miao2018identifying, mastouri2021proximal, kompa2022deep, wang2021proxy}, from which we build upon later in \cref{sec:theoretica_results}.
{Recent works have aimed to unify causal inference and generative modeling under hidden confounding~\citep{xia2021causal, pmlr-v206-xi23a, nasr2023counterfactual}. 
In particular,  Neural Causal Models perform causal identifiability and estimation under hidden confounding on discrete variables~\citep{xia2021causal, xia2023neural} by,
given a causal query, training two ``adversarial'' models and returning their estimation if they coincide. %
The model by \citet{nasr2023counterfactual}, instead, focuses on counterfactual queries for simple causal graphs where adjustment sets or instrumental variables are available. %
Our work thus aims to complement this line of work by providing a \emph{practical and scalable} GCM to solve a broad class of causal queries, interventional or counterfactual, on continuous variables and large causal graphs with a single end-to-end training. %

    \section{Confounded structural causal models} 
\label{sec:background}

\begin{definition} \label{def:scm}
    A \emph{(confounded)  Structural Causal Model (SCM)} is a triplet $\scm \coloneqq (\funcb, \distribution[\varexo], \distribution[\varhidden])$ describing a data-generating process over $\obssize$ observed (endogenous) variables $\varobs\coloneqq \vect{\evarobs}{\obssize} \in\gX$:
    \begin{equation}
        \ervobs_\indexone \coloneqq \func[\indexone](\varparents[\ervobs]{\indexone}, \ervu_\indexone, \varhidden)  \quad \text{for}\quad  \indexone = \range{\obssize} \,, 
        \; \text{with} \; \varexo\coloneqq \vect{\evarexo}{\obssize} \sim \distribution[\varexo]\,, \; 
        \varhidden \sim \distribution[\varhidden]\,,
        \label{eq:scm}
    \end{equation}%
    where $\func[\indexone]$ represents the structural equation to compute the \nth{\indexone} endogenous variable, $\evarobs_\indexone$, from its observed \emph{causal parents}, \varparents[\ervobs]{\indexone}, the \nth{\indexone} exogenous variable, $\evarexo_\indexone$, and the  \emph{hidden confounders},
    \varhidden
    $\in\gZ$.
\end{definition}

While we make the dependence on the hidden confounders explicit for all observed variables in \cref{eq:scm}, we assume \wlogg that a subset of them may not be directly affected by the hidden confounders. 
Furthermore, given a SCM \scm, we denote by \graph the \emph{faithful}
causal graph that it induces, representing a direct causal relationship between pairs of 
endogenous and hidden variables \emph{only} if it exists. %

A core idea of causal inference is the do operator~\citep{Pearl2012TheDR}, denoted by $\doop(\vartreat)$, which formalizes the action of externally intervening on the variable \vartreat, \ie, to fix \vartreat to a value independently of its parents. 
The do operator enables the computation of interventional and counterfactual queries in SCMs~\citep{peters2017elements}:%

\begin{definition} A \emph{causal query} $Q(\scm)\coloneqq p_\scm(\giventhat{\varoutcome}{\doop(\vartreat), \varcov})$ is a distribution over $\varoutcome \in \varobs$ (the \emph{outcome} variable), as a result of intervening upon the variable $\vartreat\in\varobs$ (the \emph{treatment} variable). %
Additionally,  $Q(\scm)$ denotes an \emph{interventional} or \emph{counterfactual} query if the variable $\varcov$ is, respectively, the empty set or the vector of observed factual values, \factual\varobs. 
\end{definition}%

We call a causal query \textit{identifiable} if it can be expressed as a function of the observational distribution, $p_\scm(\varobs)$, and the causal graph \graph~\citep{pearl2009causality}. As a result, any SCM inducing the same graph and matching the observational distribution produces correct estimates of that causal query.
Moreover, \textit{any} identifiable query can be rewritten this way using a set of three rules, the \emph{do-calculus}~\citep{981e909a-69b6-346e-bd9d-1ef8e392bda3}. Yet, in the presence of \emph{hidden confounders}, this may not be possible and even applying the do-operator to evaluate causal queries would produce incorrect estimates, as unaccounted confounders would bias the results.

    \section{Deconfounding causal normalizing flows}
\label{sec:deconfounding-causal-flows}

In this work, we assume the existence of an underlying confounded SCM, \scm, as in \cref{def:scm}, of which we have access to $\samplesize$ \iid observations as well as to the faithful causal graph, \graph. 
Our objective is to design and learn a CGM that can \emph{accurately estimate as many causal queries from the original SCM as possible}, despite the presence of unobserved hidden confounding. 
In other words, we seek a substitute model of \scm that we can use to accurately perform causal inference.

\paragraph{Assumptions.}  

In addition, we assume all variables to be continuous, and the SCM \scm to: \itemi have $C^1$-diffeomorphic causal equations conditioned on \varhidden, %
and \itemii induce an acyclic causal graph \graph.

Note that assumption \itemi implies that $\funcb:\gU\times\gZ\rightarrow\gX$ is invertible from \varobs to \varexo, given \varhidden. This is not a limiting assumption, since we never observe \varexo and we can always find an invertible mapping by merging all \varexo producing the same observations and taking their Kn\"othe-Rosenblatt transport \citep{Knothe1957ContributionsTT, 10.2307/2236692},  while remaining causally equivalent to the original SCM assuming all other assumptions hold.

\subsection{(Unconfounded) Causal normalizing flows}
\label{subsec:causal-flows}

Causal normalizing flows (CNFs)
\citep{javaloy2024causal} play an important role in this work, as they form the foundations of \ours, given their identifiability guarantees despite a mild set of assumptions.
Given a causal graph \graph, a CNF, \flow, is a masked autoregressive normalizing flow~\citep{papamakarios2021normalizing} built such that, paired with a distribution \distribution[\varexo], defines an unconfounded SCM $\scm_\thetab = (\flow, \distribution[\varexo])$ that induces graph \graph by design.

As demonstrated by \citet{javaloy2024causal}, CNFs represent a remarkable family of CGMs, as they not only form a parametric family of \emph{identifiable SCMs}, 
but they can provably approximate the underlying SCM in the three rungs of Pearl's ladder of causation~\citep{pearl2009causality} %
simply by maximizing the observed joint evidence, \ie, $\max_\thetab \log p_\thetab(\varobs)$.
Furthermore, CNFs are also equipped with an \emph{exact do-operator} for efficient sampling of any causal query, enabling their use for complex causal-inference tasks.

Their main downside, as discussed in \cref{sec:intro}, is that CNFs need to assume causal sufficiency---on top of the assumptions made above---to guarantee the aforementioned capabilities, thus limiting their application. %
Next, we attempt to address this limitation %
without losing theoretical guarantees.

\subsection{Deconfounding causal normalizing flows}
\label{sec:decaf}

\begin{figure}[t]
    \centering
    \begin{subfigure}[t]{0.5\linewidth}
    	\centering
		\def\dx{1.}
		\def\dy{0.4}
		\begin{tikzpicture}[
			every path/.style={semithick}
			]
			\pic (A) {
				architecture={%
					{\varproxyo, \varproxyt, \varblock, \vartreat, \varoutcome, $\epsilon$}
				}{6}{%
					{1, 1, 1, 1, 1, 0}
				}{%
					{$\varhidden$} %
				}{1}{%
					{{"draw",},{"draw",},{"draw",},{"draw",},{"draw",},{"draw",}}
				}{%
					0
				}{%
					{1, 1, 1, 1, 1, 1}	
				}
			};
			
			\path (Az0.east) ++ (0.5, 0) pic (B) {
				architecture={%
					{, $\varexo_\varproxyo$, $\varexo_\varproxyt$, $\varexo_\varblock$, $\varexo_\vartreat$, $\varexo_\varoutcome$}
				}{6}{%
					{1, 0, 0, 0, 0, 0}
				}{%
					{\varproxyo, \varproxyt, \varblock, \vartreat, \varoutcome}
				}{5}{%
					{%
						{"hidden","hidden","hidden","hidden","hidden"},%
						{"draw","","","draw",""},%
						{"","draw","","","draw"},%
						{"","","draw","draw","draw"},%
						{"","","","draw","draw"},%
						{"","","","","draw"},%
					}%
				}{%
					1
				}{%
					{0, 1, 1, 1, 1, 1}	
				}
			};
            \draw[myarrow] (Az0.90) |- (Bx0.west -| Bbox.west);
			\path (Ax0 -| Ay0) node {\encoder};
			\path (Bx0 -| By0) ++ (-0.5mm, 0) node {\small$\flow^{-1}$};
			
			\coordinate (m) at ($(Abox)!0.5!(Bbox)$);
			\path (Abox.south -| m) ++ (0, -0.75*\dy) node (G) {\graph};
			\draw[myarrow] (G) -| (Abox.south);
			\draw[myarrow] (G) -| (Bbox.south);
            
            \path 
                (Abox.north) node[above] {$\substack{\text{Deconfounding network}}$} --
                (Bbox.north) node[above] {$\substack{\text{Generative network}}$};
		\end{tikzpicture}
        \caption{\ours training computations.}
        \label{fig:architecture}
    \end{subfigure}%
    \begin{subfigure}[t]{0.5\linewidth}
    \centering
        \begin{tikzpicture}
            \path pic (I-) {
				architecture={%
					{, $\varexo_\varproxyo$, $\varexo_\varproxyt$, $\varexo_\varblock$,, $\varexo_\varoutcome$}
				}{6}{%
					{1, 0, 0, 0, 0, 0}
				}{%
					{\varproxyo, \varproxyt, \varblock, \color{secondcolor}\vartreat, \varoutcome}
				}{5}{%
					{%
						{"hidden","hidden","hidden","","hidden"},%
						{"draw","","","",""},%
						{"","draw","","","draw"},%
						{"","","draw","","draw"},%
						{"","","","","draw"},%
						{"","","","","draw"},%
					}%
				}{%
					1
				}{%
					{0, 1, 1, 1, 0, 1}	
				}
			};
            \path[secondcolor] (I-a4.center) node (I-a4) {\color{secondcolor}$\varexo_\vartreat'$};
            \draw[myarrow,secondcolor,thick]
            (I-a4.east) -- (I-a4 -| I-box.west);
            \draw[myarrow, secondcolor,thick] (I-x4) node[input] {} -- (I-y3) node[output] {};
            \draw[myarrow, secondcolor,thick] (I-z3 -| I-box.east) -- (I-z3);
            \path (I-z3.center) node[right] {\color{secondcolor}$= \alpha$};
            \path (I-x0 -| I-y0) ++ (-0.5mm, 0) node {\small$\flow^{-1}$};
            \path (I-box.north) node[above] {$\substack{\text{{\color{secondcolor}Intervened} generative network}}$};

            \path (-0.5, 1) node[left] (z) {$\left.\begin{aligned}\varhidden&\sim p(\varhidden)\\\varhidden &\sim q_\phib(\giventhat{\varhidden}{\factual\varobs})\end{aligned}\right\rbrace$} -- ($(z.east)!0.1!(I-box.west)$) coordinate (c1);
            \draw[myarrow] (z.east) -- (c1 |- z.east) |- (I-a0 -| I-box.west);

            \path (-0.75, -0.4) node[left] (u) {$\left.\begin{aligned}\varexo&\sim p(\varexo)\\\varexo &= \flowz(\factual\varobs)\end{aligned}\right\rbrace$} -- ($(u.east)!0.3!(I-a0.west)$) coordinate (c2);
            \foreach \i in {1,2,3,5}
                \draw[myarrow] (u.east) -- (c2 |- u.east) |- (I-a\i);
            \path (I-box.south) ++ (0, -0.75*\dy) node (G) {};
        \end{tikzpicture}
        \caption{Computation of causal queries.}
        \label{fig:do-operator}
    \end{subfigure}  
    \caption{\textbf{Example of \ours computations} for the causal graph \graph in \cref{app:fig:miao-block-graph}. Circles represent input/output variables of the masked conditional normalizing flows, and black dots conditional inputs. \captiona Steps performed during training (\cref{eq:elbo}), where $\epsilonb$ is a non-causal random variable needed to model $\varhidden$ with \encoder. \captionb Steps performed to compute an interventional or counterfactual query with CNFs~\citep{javaloy2024causal}, see \cref{app:sec:do-operator}, where $\varexo_\vartreat'$ is the value for which $\vartreat = \alpha$ always, \ie, $\varexo_\vartreat' = \flowz(\flowz^{-1}(\varexo)_{\varobs\setminus\vartreat}, \alpha)_\vartreat$\equationPunctuation{.}}
    \label{fig:model}
\end{figure}

We now introduce the \emph{\underline{de}confounding \underline{ca}usal normalizing \underline{flow}} (\ours), a family of models which extend CNFs~\citep{javaloy2024causal} to account for hidden confounding while retaining all their theoretical properties.
To achieve this, \ours follows the structure of a variational autoencoder~\citep{kingma2013auto}, \ie, \ours comprises two main components: \itemi~\emph{a generative network} that exploits structural constraints to accurately model the underlying SCM, given a substitute of \varhidden; and \itemii~an \emph{inference network} which approximates the \emph{intractable} posterior distribution of \varhidden as modeled by the generative network, given the observed endogenous variables.
In the following, we provide further details on both networks. 

\paragraph{Generative network.}
We adapt CNFs \citep{javaloy2024causal} to take the hidden confounders as conditional inputs by using conditional masked autoregressive normalizing flows~\citep{winkler2019learning}, instead of unconditional ones.
The resulting model, \flow, is thus an invertible transformation, conditioned on~\varhidden, describing a data-generating process that maps a set of exogenous variables \varexo %
to endogenous ones and vice versa, \ie,
$\flowz (\varobs) = \varexo \sim P_{\varexo}$ and $\varobs = \flowz^{-1}(\varexo)$\equationPunctuation{,}
where we further exploit the given causal graph \graph to ensure that the generative process is faithful, \ie, that
\begin{equation}
    p_\thetab(\giventhat{\varobs}{\varhidden}) = \prod_{\indexone=1}^\obssize p_\thetab(\giventhat{\ervobs_\indexone}{\varparents[\varobs]{\indexone},  \varhidden}) \equationPunctuation{,} \label{eq:factorization-decoder}
\end{equation}
defining a data-generating process similar to that of \cref{def:scm} and, just as in that definition, in \cref{eq:factorization-decoder} \emph{only the children of \varhidden are actually conditioned on \varhidden}.

\paragraph{Deconfounding network.}

To model the posterior distribution of \varhidden given our observations as modeled by \flow, \ie, the abduction step needed to compute counterfactuals \citep{pearl2009causality}, %
we use another masked autoregressive conditional normalizing flow \citep{winkler2019learning}, as it can approximate this distribution arbitrarily well.
Once again, we exploit knowledge of \graph and mask the resulting network, \encoder, such that it models \varhidden using only the strictly necessary variables to ease learning:
\begin{equation}
    \small q_\phib(\giventhat{\varhidden}{\varobs}) = q_\phi\left(\giventhat{\varhidden}{\varchildren{\varhidden}
    \cup     \varparents{\varchildren{\varhidden}}}\right) \equationPunctuation{.}
    \label{eq:factorization-encoder}
\end{equation}
We provide in \cref{app:sec:implementation} a more general version of \cref{eq:factorization-encoder} that accounts for several independent hidden confounders, and empirically validate the architecture and factorization choices in \cref{app:sec:encoder_ablation,app:sec:encoder_factorization}. %

\paragraph{Training process.}

We jointly train both networks as typically done in deep latent-variable models, \ie, we maximize the evidence lower bound (ELBO)~\citep{kingma2013auto}:
\begin{equation}
    \ELBO(\thetab, \phib) 
    = \Expect[q_\phib]{\log p_\thetab(\varobs, \varhidden)} + \entropy(q_\phib(\giventhat{\varhidden}{\varobs})) 
    = \Expect[q_\phib]{\log p_\thetab(\giventhat{\varobs}{\varhidden})} - \KLop{q_\phib(\giventhat{\varhidden}{\varobs})}{p(\varhidden)} 
    \equationPunctuation{,} \label{eq:elbo}
\end{equation}
where %
$\KLoperator$ is the Kullback-Leibler divergence \citep{kullback1951information}, $\entropy$ the differential entropy \citep{kolmogorov1956shannon} and $p(\varhidden)$ is the prior distribution of \varhidden which we set as a standard Gaussian. 
The motivation for this loss is three-fold: \itemi~we want the generative network to explain the observations given samples from $q_\phib$ (first term of \cref{eq:elbo}); \itemii~as we do not know the optimal size for \varhidden, we need to prevent the deconfounding network from allocating information exclusive of \varobs in \varhidden (entropy term in \cref{eq:elbo}); and \itemiii~all the results introduced next rely on \ours matching the observational distribution, $p_\scm(\varobs)$, which we encourage since
\begin{equation}
    \max_{\phib, \thetab} \; \ELBO(\phib, \thetab) = \min_{\phib, \thetab}  \; \KLop{p_\scm(\varobs)}{p_\thetab(\varobs)}  +\KLop{q_\phib(\giventhat{\varhidden}{\varobs})}{p_\thetab(\giventhat{\varhidden}{\varobs})} \equationPunctuation{.} \numberthis \label{eq:elbo-decomposition}
\end{equation}

To avoid posterior collapse, \ie, that the approximate posterior matches the prior and thus having uninformative latent variables~\citep{wang2021posterior}, we incorporate KL balancing terms \citep{vahdat2020nvae} to prevent the KL in \cref{eq:elbo} from vanishing, ensuring that the latent representation remains informative during training.
Other implementation details, \eg, the way masking both encoder and decoder with via the causal adjacency matrix, or how to adapt the do-operator of CNFs can be found in \cref{app:sec:implementation,app:sec:do-operator}, respectively.

\subsection{Inherited causal properties}
\label{subsec:disentanglement}

As a consequence of leveraging (causal) normalizing flows, \ours inherits many of the great properties of this family of models.
For once, both components of \ours are universal density approximators~\citep{papamakarios2021normalizing} meaning that, given enough resources, the generative network can perfectly match the observational distribution and the deconfounding network can perfectly learn the modeled posterior. In other words, \emph{we can perfectly minimize the two KL terms that appear in} \cref{eq:elbo-decomposition}.
Furthermore, note that the generative network, combined with two base distributions for \varexo and~\varhidden, defines a confounded SCM as in \cref{def:scm}, \ie, $\scm_\thetab = (\flow^{-1}, \distribution[\varexo], \distribution[\varhidden])$\equationPunctuation{.}

\paragraph{Causal consistency.} 
As we leverage the causal graph \graph to appropriately mask the conditional CNF of the generative network, we have that $\scm_\thetab$ respects all causal connections described by \graph. %
In other words, \emph{$\scm_\thetab$ induces the same causal graph \graph as \scm}.
As a result, we can ensure which variable affects which when we generate observations with \flow. 
\Cref{fig:model} depicts these structural constraints relating \varexo and \varhidden with \varobs, and we provide a detailed description in \cref{app:sec:implementation}.

Moreover, we prove in \cref{app:subsec:uncounded-identifiability} that \ours preserves one of the most crucial aspects of CNFs: Identifying (in the sense of \citet{pmlr-v206-xi23a}) the underlying SCM concerning those variables that are not directly caused by \varhidden:%

\begin{proposition}[Informal]
    If \ours induces the same causal graph and observational distribution as the underlying (confounded) SCM generating the data.  
    Then, \ours recovers the SCM for every variable not in \varchildren{\varhidden}, up to an element-wise transformation of their exogenous distributions.
\end{proposition}

\paragraph{\Ours do-operator.} 

While the above result ensures the causal equivalence of $\scm_\thetab$ and \scm for unconfounded variables, it is still unclear how to intervene on $\scm_\thetab$.
To this end, we adapt the do-operator of CNFs~\citep{javaloy2024causal}, represented in \cref{fig:do-operator} and detailed in \cref{app:sec:do-operator},  which provides an efficient \emph{and exact} way of sampling from any interventional and counterfactual  distribution.
Namely, to sample from an interventional
distribution $p(\giventhat{\varobs}{\doop(\vartreat\coloneqq\alpha)})$ over $\scm_\thetab$ we: \itemi sample $\varhidden \sim \distribution[\varhidden]$ and $\varexo\sim\distribution[\varexo]$; \itemii find the value of $\evarexo_\vartreat$ that yields $\vartreat = \alpha$ given \varexo, which we can easily do as \flow is invertible given \varhidden; and \itemiii return the sample $\varobs^{\doop(\vartreat)} \coloneqq \flowz^{-1}(\varexo_{\varobs\setminus\vartreat}, \evarexo_\vartreat)$\equationPunctuation{.}
The counterfactual case is quite similar, as the bijectivity of \flow implies that every counterfactual distribution is a delta distribution given \varhidden, and we can simply follow the process above but using $\varhidden \sim q_\phib(\giventhat{\varhidden}{\factual\varobs})$ and $\varexo = \flowz(\factual\varobs)$ for step one.
As a result, we can guarantee the correctness of \ours estimations on a number of causal queries:

\begin{corollary}[Informal] \label{cor:decaf-int-estimates}
	\ours provides correct estimates of any causal, interventional or counterfactual, query for which both the treatment and outcome variables are not direct children of a hidden confounder, \ie, $\vartreat, \varoutcome \notin \varchildren{\varhidden}$\equationPunctuation{.}
\end{corollary}

Thus far, we have shown that for causal queries over non-children of \varhidden, \ours inherits the theoretical guarantees of CNFs. In the following section, we investigate under which conditions \ours can also provide correct estimates of causal queries defined over children of \varhidden.

\section{Estimation of causal queries under hidden confounding}
\label{subsec:causal-query-ident}
\label{sec:theoretica_results}

By leveraging recent results in proximal-identifiability, we next show that \ours not only preserves the properties of CNFs, but expand them.
Namely, we characterize queries which \ours correctly estimates despite the hidden confounding.
While we present here an intuitive summary of our main theoretical results, formal statements and derivations can be found in \cref{app:sec:proofs}.
Throughout this section, we define \emph{proxy variables} relative to the causal query to estimate: if we are interested in the causal effect of \vartreat over \varoutcome, which are confounded by \varhidden, a proxy is an observed variable related to \varhidden and conditionally independent of \vartreat or \varoutcome (see \cref{prop:informal} next for precise definitions). Intuitively, proxy variables contain exploitable information about the hidden confounders, which we leverage in this section to provide accurate estimates of the causal queries of interest.

\subsection{Interventional queries}

First, we consider the identifiability of \emph{hidden-confounded} interventional queries, \ie, queries of the form $\query(\scm) = p_\scm(\giventhat{\varoutcome}{\doop(\vartreat)})$\,, where $\varoutcome,\vartreat\in\varchildren{\varhidden}$ are any two children of the hidden confounder.
We summarize our findings in the following proposition, which we properly formalize in \cref{app:subsec:general-case-proof}:

\begin{proposition}[Informal] 
\label{prop:informal}
    An %
    interventional query of the form $\query(\scm) = p_\scm(\giventhat{\varoutcome}{\doop(\vartreat)})$\,, where $\varoutcome,\vartreat\in\varchildren{\varhidden}$ are two different children of \varhidden, is identifiable if there exists a (potentially empty) subset of blocking variables $\varblock\subset\varobs \setminus\{\vartreat,\varoutcome\}$, and two other variables $\varproxyo,\varproxyt\in\varobs\setminus\{\vartreat,\varoutcome,\varblock\}$ such that:
    \begin{enumerate}
        \item $(\varblock, \varhidden)$ forms a valid adjustment set, \ie, $p(\giventhat{\varoutcome}{\doop(\vartreat)}) = \iint {p(\giventhat{\varoutcome}{\vartreat, \varblock, \varhidden})} {p(\varblock, \varhidden)} \diff\varblock \diff\varhidden$\,,
        
        \item \varproxyt is a proxy variable given \varblock, \ie, $\giventhat{\varproxyt \indep (\vartreat, \varproxyo)}{\varblock, \varhidden}$\,,
        
        \item \varproxyo is a null proxy variable given \varblock, \ie, $\giventhat{\varoutcome \indep \varproxyo}{\vartreat, \varblock, \varhidden}$\,, and
    
        \item both \varproxyt and \varproxyo yield enough information about the hidden confounder \varhidden.
    \end{enumerate}
\end{proposition}

\Cref{prop:informal} extends the results from \citet{miao2018identifying} and \citet{wang2021proxy} to prove identifiable of queries under hidden confounding \textit{even if treatment and outcome have observed parents in common}.
In turn, these results render causal queries identifiable in the infinite-data regime by leveraging proxy information, complementing classical do-calculus %
\citep{kuroki2014measurement}.
Intuitively, \varproxyt serves the purpose of building a function which ``substitutes'' the hidden confounder for that query, and \varproxyo is used to ensure that this substitute yields the correct estimate.
From this result, one natural step is then to extend the class of causal queries which are identifiable using do-calculus, where we introduce the queries identifiable with \cref{prop:informal} as an additional base case for the do-calculus recursive steps:
\begin{restatable}{corollary}{queryreduction} \label{cor:do-calculus}
    An interventional query is identifiable if, using do-calculus, it can be reduced to a combination of observational queries and identifiable interventional queries in the sense of \cref{prop:informal}.
\end{restatable}

\begin{wrapfigure}[11]{R}{0.3\linewidth}
    \centering
    \begin{tikzcd}[
     cells={nodes={main node}}, column sep=tiny, row sep=small
     ]
	     & \varproxyo \arrow[rr] & & \vartreat \arrow[secondcolor, d] \\
	     |[hidden]| \varhidden \arrow[hidden, ru] \arrow[hidden, rd] \arrow[hidden, rr] \arrow[hidden, rrrd] \arrow[hidden, rrru, secondcolor] & & \varblock \arrow[dr] \arrow[ur] & |[maincolor]| \rvm \arrow[secondcolor, d] & \varoutcome_1 \arrow[leftarrow, dl] \arrow[leftarrow, ul] \\
	     & \varproxyt \arrow[rr] & & \varoutcome_2
	 \end{tikzcd}%
    \caption{Illustrative causal graph where the  \textcolor{firstcolor}{presence} or \textcolor{secondcolor}{absence} of some parts render $p(\giventhat{\varoutcome_1}{\doop(\vartreat)})$ identifiable using do-calculus. Else, \cref{prop:informal} yields identifiability if \varproxyt and \varproxyo are informative proxies.}%
    \label{fig:example-confounded-scm}
\end{wrapfigure}

To understand the implications of \cref{prop:informal,cor:do-calculus}, consider the causal graph in \cref{fig:example-confounded-scm}, and suppose we want to compute $\query(\scm) = p(\giventhat{\varoutcome_1}{\doop(\vartreat)})$\equationPunctuation{.}
Then, we can proceed as usual and apply the rules of probability theory and do-calculus to rewrite $\query(\scm)$ as %
\begin{equation}
	\query(\scm) = p(\giventhat{\varoutcome_1}{\doop(\vartreat)}) %
	= \int p(\giventhat{\varoutcome_1}{\vartreat, \varoutcome_2}) p(\giventhat{\varoutcome_2}{\doop(\vartreat)}) \dif\varoutcome_2 \equationPunctuation{.}
\end{equation}
As a result, the identifiability of $p(\giventhat{\varoutcome_2}{\doop(\vartreat)})$ implies that of $\query(\scm)$.
We can then devise a few different scenarios:
\begin{enumerate}
	\item If there is no edge from \varhidden to \vartreat, \ie, $\vartreat \notin \varchildren{\varhidden}$, then the backdoor criterion holds for $\{\varproxyo, \varblock\} = \varparents{\vartreat} \subset \varobs$ and both $p(\giventhat{\varoutcome_1}{\doop(\vartreat)})$ and $p(\giventhat{\varoutcome_2}{\doop(\vartreat)})$ are identifiable.
	\item If there exists a mediator variable between \vartreat and $\varoutcome_2$, $\rvm$, we can apply the front-door adjustment and both $p(\giventhat{\varoutcome_1}{\doop(\vartreat)})$ and $p(\giventhat{\varoutcome_2}{\doop(\vartreat)})$ are identifiable.
\end{enumerate}

\vspace{-\parskip}%
\begin{enumerate}
\setcounter{enumi}{2}
	\item If $\varoutcome_2$ is not caused by $\vartreat$, then $p(\giventhat{\varoutcome_2}{\doop(\vartreat)}) = p(\varoutcome_2)$ and both queries are identifiable.
	\item Otherwise, we can still render $p(\giventhat{\varoutcome_2}{\doop(\vartreat)})$ (and thus $p(\giventhat{\varoutcome_1}{\doop(\vartreat)})$) identifiable if \varproxyt and \varproxyo yield sufficient information about \varhidden (intuitively, this means that the posterior of \varhidden changes enough as we change \varproxyt and \varproxyo; we formalize this notion in \cref{def:completeness}) and we can hence apply \cref{prop:informal}.
\end{enumerate}

The example above nicely illustrates how \cref{prop:informal} complements do-calculus: if we find a query unidentifiable due to reaching a dead end with do-calculus---in this case, $p(\giventhat{\varoutcome_2}{\doop(\vartreat)}))$---then \cref{prop:informal} provides an additional case for which the query can still be made identifiable.
Moreover, this case clearly shows how \cref{prop:informal} extends prior results as these \emph{did not allow} for common observable ancestors between outcome and treatment~\citep{miao2018identifying,wang2021proxy}.
Nevertheless, note that \cref{prop:informal} provides only sufficient conditions for identifiability, and there could exist identifiable queries which do not comply with the requirements of the proposition.

Finally, recall from \cref{subsec:disentanglement} that, similar to CNFs \citep{javaloy2024causal}, we can readily interpret the generative network of \ours as a parametric confounded SCM (\cref{def:scm}) of the form $\scm_\thetab \coloneqq (\flow^{-1}, \distribution[\varexo], \distribution[\varhidden])$.
This SCM induces the same causal graph as the underlying \scm  by design, \graph, and since the family of normalizing flows are universal density approximators, $\scm_\thetab$ can match the observational distribution $p_\scm(\varobs)$ given enough resources. %
We can then leverage the previous results to prove the following:
\begin{restatable}{corollary}{decafscm} \label{cor:decaf-cf-estimates}
	If \ours induces the same causal graph \graph as \scm and $p_\scm(\varobs) \aeeq p_\thetab(\varobs)$, then \ours provides correct estimates of any query identifiable in the sense of \cref{cor:do-calculus}.
\end{restatable}

In other words, \cref{cor:decaf-int-estimates} guarantees that, if we match the observational distribution with \ours, then the do-operator presented in \cref{subsec:disentanglement} provides a correct estimate of the identifiable query of interest.

\subsection{Counterfactual queries \label{subsec:cf-queries}}

Next, we focus on counterfactual queries of the form $\query(\scm) = p_\scm(\giventhat{\cfactual\varoutcome}{\doop(\cfactual\vartreat), \factual\varobs})$, where $\factual\varobs$ is the observed factual. Intuitively, this query represents \textit{the distribution the outcome would have had, had we intervened on the treatment variable}.
We demonstrate, for the first time to our knowledge, a one-to-one correspondence between proxy-identifiable interventional and counterfactual queries.
More specifically, we show that (all formal derivations can be found in \cref{app:subsec:cf-identifiablity}):

\begin{proposition}[Informal]
    If an interventional query $p(\giventhat{\varoutcome}{\doop(\vartreat)})$ is identifiable in the sense of \cref{prop:informal}, then its counterfactual counterpart, $p(\giventhat{\cfactual\varoutcome}{\doop(\cfactual\vartreat), \factual\varobs})$\equationPunctuation{,} is also identifiable.
    \label{prop:counterfactuals}
\end{proposition}

\begin{wrapfigure}[15]{R}{0.3\linewidth}
	\centering
    \vspace{-1.\baselineskip}
	\begin{tikzcd}[
		cells={nodes={main node}}, column sep=tiny, row sep=small
		]
		& \cfactual\vartreat \arrow[rr] & & \cfactual\varoutcome \\
		\cfactual\varproxyo \arrow[ur, secondcolor, "\tiny \text{\faCut}"{marking, rotate=90,xshift=-0.3mm, pos=0.6}] & &  & & \cfactual\varproxyt \arrow[ul] \\
		|[dashed]| \evarexo_\varproxyo \arrow[d, dashed] \arrow[u, dashed] & |[dashed]| \evarexo_\vartreat \arrow[dd, dashed] \arrow[uu, dashed, secondcolor, "\tiny \text{\faCut}"{marking, yshift=-0.3mm, rotate=90, pos=0.8}] & |[hidden]|\varhidden \arrow[hidden, dll] \arrow[hidden, drr] \arrow[hidden, ddl] \arrow[hidden, ddr] \arrow[hidden, ull] \arrow[hidden, urr] \arrow[hidden, uul, secondcolor, "\tiny \text{\faCut}"{marking, yshift=0.3mm, rotate=270, pos=0.8}] \arrow[hidden, uur] & |[dashed]| \evarexo_\varoutcome \arrow[dd, dashed] \arrow[uu, dashed] & |[dashed]| \evarexo_\varproxyt \arrow[d, dashed] \arrow[u, dashed] \\
		|[observed]| \factual\varproxyo \arrow[dr] & &  & & |[observed]| \factual\varproxyt \arrow[dl] \\
		& |[observed]| \factual\vartreat \arrow[rr] & & |[observed]| \factual\varoutcome
	\end{tikzcd}
	\caption{\textbf{Twin SCM} with observed factual nodes grayed out and edges severed to compute $p(\giventhat{\cfactual\varoutcome}{\doop(\cfactual\vartreat), \factual\varobs})$ in red.%
    }
	\label{fig:twin-network-cf}
\end{wrapfigure}
The proof of \cref{prop:counterfactuals} exploits the notion of twin SCM~\citep{Balke1994ProbabilisticEO}, which duplicates the structural equations for the factual and counterfactual worlds while sharing the exogenous variables, and the fact that \cref{app:prop:causal-query-identifiability} (the formal version of \cref{prop:informal}) allows for queries with additional covariates as long as they do not form colliders, which is always the case with $\factual\varobs$ in $p_\scm(\giventhat{\cfactual\varoutcome}{\doop(\cfactual\vartreat), \factual\varobs})$, as we show in the illustrative twin network of \cref{fig:twin-network-cf}.
We can then follow the same derivations from the previous section to show that:
\begin{corollary}
    If \ours induces the same causal graph \graph as \scm and $p_\scm(\varobs) \aeeq p_\thetab(\varobs)$, then \ours provides correct estimates of any counterfactual query which can be decomposed in a combination of (proxy-)identifiable queries using do-calculus.
    \label{prop:counterfactuals-estimation}
\end{corollary}

\Cref{prop:counterfactuals-estimation} implies that, if \ours correctly estimates an interventional query, %
then it also does for its counterfactual counterpart.
While the above results can look surprising at first, recall that we assume continuous endogenous variables and diffeomorphic causal generators (\cref{sec:deconfounding-causal-flows}).
Moreover, the correct estimation of counterfactual queries does not come without challenges: \itemi we need to accurately estimate $p_\thetab(\giventhat{\varhidden}{\varobs})$, which is why it is crucial to correctly design and train $q_\phib$; and \itemii given \varhidden and \varobs, we need to accurately perform the abduction step. Fortunately, the latter step is trivialized using CNFs as generative networks~\citep{javaloy2024causal}, since they are bijective given~\varhidden.

\paragraph{Remarks.} Whilst \ours can estimate \emph{any} causal query, \emph{this estimation can be incorrect for unidentifiable queries}. 
Therefore, we must verify the identifiability for each query of interest, which we aim to ease by providing algorithms to check the graphical requirements of \cref{prop:informal} in \cref{sec:app:path_identifiability}. 
Namely, \cref{alg:causal_query} checks if a query that involves a specific treatment-outcome pair, which includes average treatment effects and counterfactuals, is identifiable. If we were interested in a query on all variables, \eg, as samples from an interventional distribution, we should evaluate the identifiability of the causal query for all descendants of the treatment, as proposed in \cref{alg:interventional_gen_identifiability}.
Similarly, note that all results above rely on the assumption that \ours matches the observational distribution, and thus it is crucial to ensure that the training completed successfully.

\section{Empirical evaluation}
\label{sec:experiments}

In this section, we assess the performance of \ours relative to existing methods. Namely, we show that \ours accurately estimates interventional and counterfactual queries when the requirements of \cref{cor:decaf-int-estimates,cor:decaf-cf-estimates} are met. %
All experimental details are described in \cref{app:sec:additional-results}.

\paragraph{Common evaluation.}

For all experiments, we measure the estimation quality for  interventional and counterfactual queries using the mean absolute error (MAE) of, respectively, the average treatment effect (ATE) and the counterfactual (CF) samples, as we have access to the ground-truth values. 
We use as reference (or \emph{oracle}) a CNF~\citep{javaloy2024causal} that \emph{does observe} the hidden confounders. 
We also account for differences across observed variables by computing all errors over the standardized variables. Note that ATE and CF errors provide complementary measures of estimation quality, and their interpretation is best understood relative to the oracle performance in each metric.

\paragraph{Hyperparameter selection.} We choose hyperparameters based on the MMD~\citep{gretton2006kernel} over validation observational data, following our theoretical premise that \ours correctly estimates causal queries when $p_\scm(\varobs) =p_\thetab(\varobs)$; see \cref{app:sec:hyperparameters} for details on the hyperparameter grid.

\subsection{Ablation study and practical considerations}
\label{subsec:ablation}

In order to provide insights into practical limitations of \ours, we first
conduct an ablation study to understand the extent for which misspecifying the size of \varhidden affects \ours, as well as its sensitivity to the number of available proxies.
For additional details and results, refer to \cref{app:sec:additional-results-ablation}.

\paragraph{Experimental setup.}

\begin{wrapfigure}[9]{R}{.25\linewidth}  
    \vspace{-0.5\baselineskip}
    \centering
        \begin{tikzcd}[
        cells={nodes={main node}}, column sep=tiny, row sep=normal
    ]
         |[xshift=0mm]| \ervn_1 \arrow[d, leftarrow, hidden] \arrow[drr, leftarrow, hidden] \arrow[rr, phantom, "\ldots" description] & |[draw=none]| {} \arrow[dl, leftarrow, hidden] \arrow[dr, leftarrow, hidden] & |[xshift=-0mm]|\ervn_\proxysize \arrow[dll, leftarrow, hidden] \arrow[d, leftarrow, hidden]\\
         |[hidden]|\varhidden_1 \arrow[hidden, d] \arrow[hidden, drr] & & |[hidden]|\varhidden_2 \arrow[hidden, dll] \arrow[hidden, d] \\
         \vartreat\arrow[rr] & & \varoutcome  
    \end{tikzcd}
    
    \caption{Ablation graph.}
    \label{fig:ablation_graph}
\end{wrapfigure}

We consider two synthetic SCMs with linear and non-linear causal equations that follow the causal graph \graph depicted in 
\cref{fig:ablation_graph},
comprising two independent hidden confounders affecting every variable, and \proxysize null proxies. 
We evaluate how well \ours estimates $p(\giventhat{\varoutcome}{\doop(\vartreat)})$ as we change the number of proxy variables, \proxysize, and the specified latent dimensionality, \hiddensize.

\paragraph{Proxy informativeness.}

The completeness condition (\cref{app:prop:causal-query-identifiability}), \ie, that proxies yield ``\emph{enough information}'' (\cref{prop:informal}), is difficult to verify.
Fortunately, \cref{fig:cf_ablation} shows that using more proxies consistently improves the estimation of confounded causal queries in practice, as it is more likely to satisfy completeness. Thus, practitioners should aim to collect as many informative proxies as possible to ensure correct causal estimates. %

\begin{wrapfigure}[10]{R}{0.5\linewidth}
	\centering
	\vspace{-1\baselineskip}
	\includegraphics[width=\linewidth]{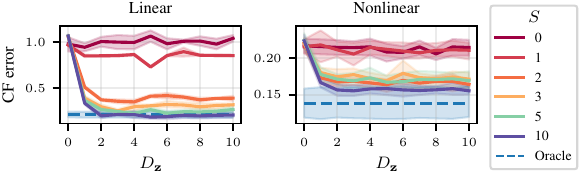}
	\caption{CF error as we increase the number of proxies, \proxysize, and the latent dimensionality, \hiddensize. Plots show mean and \SI{95}{\percent} CI over 5 realizations, intervening on the \nth{25}, \nth{50}, and \nth{75} percentiles of \vartreat.}
	\label{fig:cf_ablation}
    \vspace{\baselineskip}
\end{wrapfigure}

\paragraph{Latent dimensionality.}

When the dimensionality of the hidden confounders is unknown, we expect the entropy term in \cref{eq:elbo} to prevent modeling exogenous variables with \varhidden, as discussed in \cref{sec:decaf}. \Cref{fig:cf_ablation} corroborates our intuition, showing that \ours remains robust to over-specification of the latent dimension, \hiddensize, while error increases as we underestimate it. This suggests that, in practice, choosing a large latent space is preferable.

\paragraph{Other ablations.}

We summarize here other experiments of practical interest that can be found in the appendix. 
\Cref{app:sec:encoder_ablation,app:sec:encoder_factorization} corroborate our design choices for the deconfounding network, namely, the use of conditional normalizing flows and the posterior factorization in \cref{eq:factorization-encoder}.
Then, \cref{app:sec:ablation_train_size} assess the sample efficiency of \ours, showing that both \ours and the oracle monotonically improve their estimations as the training size increases, supporting that correct causal estimates are obtained when the model accurately fits the observational distribution.

\subsection{Semi-synthetic experiments}

\begin{figure}[t]
	\centering
	\begin{subfigure}{0.49\linewidth}
		\centering
		\includegraphics[width=\linewidth]{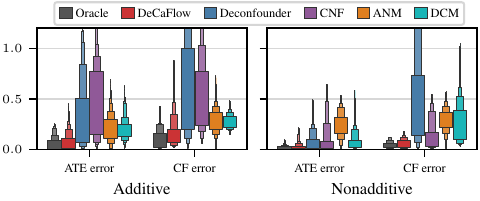}
		\caption{Sachs' dataset.}
		\label{fig:sachs_results}
	\end{subfigure}
	\begin{subfigure}{0.49\linewidth}
		\centering
		\includegraphics[width=\linewidth]{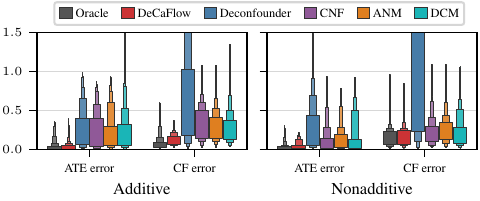}
		\caption{Ecoli70 dataset.}
		\label{fig:ecoli_results}
	\end{subfigure}
	\caption{ATE and CF error boxenplots~\citep{letter-value-plot} of different CGMs on the \captiona Sachs and \captionb Ecoli70 datasets, aggregating over all \textcolor{idcolor}{identifiable} direct effects (see  \cref{fig:sachs-graph,fig:ecoli_graph_custom}) after intervening on their \nth{25}, \nth{50}, and \nth{75} percentiles over 5 random initializations.} %
	\label{fig:results_sachs_and_ecoli}
\end{figure}

Next, we evaluate how \ours performs relative to existing approaches. To this end, we consider semi-synthetic datasets for which we have access to the ground-truth.
Additional details, results, and a justification on the use of semi-synthetic datacan be found in  \cref{app:sec:additional-results-sachs,app:sec:additional-results-ecoli}.

\paragraph{Baselines.}

We %
consider
three CGMs which assume causal sufficiency and are thus \emph{unaware} of the hidden confounders: CNFs~\citep{javaloy2024causal}; ANMs~\citep{hoyer2008nonlinear}; and DCMs~\citep{chao2023interventional}; as well as the Deconfounder~\citep{wang2019blessings}, which uses proxies to provide unbiased ATE estimates under hidden confounding, yet it requires to train one model per outcome. %
We take the oracle model as a lower bound of the error. 

\subsubsection{Protein-signaling networks}

\begin{wrapfigure}[10]{r}{0.25\linewidth}
	\centering
	\vspace{-1.25\baselineskip}
	    \begin{tikzcd}[
        cells={nodes={fairnessobs}}, column sep=small, row sep=small]
          &
        |[fairnesshidden]|\PKC \arrow[hidden, r]
                         \arrow[hidden, dl]
                         \arrow[hidden, ddl]
                         \arrow[hidden, drr]
                         \arrow[hidden, ddrr]
        & |[fairnesshidden]|\PKA \arrow[hidden, dll]
                         \arrow[hidden, ddll]
                         \arrow[hidden, ddl]
                         \arrow[hidden, dd]
                         \arrow[hidden, dr]
                         \arrow[hidden, ddr] & \\
         \Raf \arrow[identifiable, d] 
        & & &  \Jnk \\
        \Mek \arrow[identifiable, r] 
        & \Erk \arrow[identifiable, r] & \Akt & \Pte \\
        & \Plcg \arrow[r] \arrow[rr, bend right] & \pipthree \arrow[r] & \piptwo \\
    \end{tikzcd} 
	\caption{Sachs' graph. Green edges mark \textcolor{idcolor}{proxy-identifiable} effects.}
	\label{fig:sachs-graph}
\end{wrapfigure}
Following \citet{chao2023interventional}, we first experiment with the protein-signaling network dataset \citep{sachs05dataset}.
Namely, we randomly generate a non-linear SCM inducing the same causal graph as the original dataset, see \cref{fig:sachs-graph}, except for the root nodes, for which we use the original data.
As a result, we 
have a bidimensional hidden confounder, \PKC and \PKA, and three treatment variables to intervene upon, \Raf, \Mek, and \Erk. 
We consider additive and non-additive causal equations, measure the effect of interventions on the downstream nodes and, more importantly, ensure when generating the SCM that the randomized effect of the hidden confounder is perceptible. %

\paragraph{Results.} 

We present a visualization of the %
results in \cref{fig:sachs_results}, where we can %
observe that \ours outperforms every considered approach in all cases, for both ATE and counterfactual errors, \emph{staying on par with the oracle model}.
Moreover, we appreciate a great difference in performance between \ours and CNFs, which corroborates the importance of the architecture and variational training employed by \ours, since a CNF is equivalent to \ours with $\hiddensize = 0$\equationPunctuation{.}

\subsubsection{Gene networks} 

We next conduct a similar experiment as in the previous section, considering this time the causal graph of the Ecoli70 dataset~\citep{schafer2005ecoli}, depicted in \cref{fig:ecoli_graph_custom}, which represents a gene network extracted from E. coli data. This time, we replace root nodes with Gaussian variables. See \cref{app:sec:additional-results-ecoli}.

\paragraph{Results.}

Similar to the previous case, the results presented in \cref{fig:ecoli_results} demonstrate that \ours is indeed able to closely match the performance of the oracle model, outperforming existing approaches. 
It is worth-noting, however, that the non-additive case shows long-tailed error distributions for all models, including the oracle, highlighting unavoidable issues of any data-centric approach, also \ours, and that have to be considered during evaluation and deployment.

\begin{wrapfigure}[6]{r}{0.175\linewidth}  %
    \centering
    \vspace{-\baselineskip}
    \renewcommand{\figurename}{Fig.}%
    \includegraphics{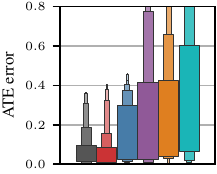}
\end{wrapfigure}
We feel compelled to explain that the striking performance of the Deconfounder is an artifact of evaluating on causal queries that it cannot correctly estimate. 
As we discuss in \cref{app:sec:additional-results}, the Deconfounder guarantees correct ATE estimation under more stringent assumptions than those from \cref{sec:theoretica_results}.
If we plot the ATE error evaluated on only those queries that meet Deconfounder's assumptions, we indeed observe that it achieves significantly lower error, %
as shown in the inset figure. %

This experiment highlights every strength of \ours, as it needs to: \itemi~model several hidden confounders affecting different sets of variables; %
\itemii~correctly estimate all causal queries for which we have proxy information; 
and \itemiii achieve the above in an agnostic manner, \ie, training the model out-of-the-box and \emph{one single time}, despite the graph \graph having 43 observed variables.

\subsection{Fairness real-world use case}
 \label{sec:usecase}

Taking inspiration from the experiments by \citet{kusnerCounterfactualFairness2017} and \citet{javaloy2024causal} we test whether, by modeling the confounded SCM with \ours, we can leverage it for more than causal-query estimation and, in particular, for counterfactual-fairness prediction.
See \cref{app:sec:additional-results-fairness} for further details.

\begin{wrapfigure}[8]{r}{0.3\linewidth}
    \centering
    \vspace{-1.5\baselineskip}
        \begin{tikzcd}[
        cells={nodes={fairnessobs}},
        column sep=small, 
        row sep=small,
        /tikz/execute at end picture={
            \draw[dashed, secondcolor] (U |- 0, -1.1) -- (U |- 0, 1.25);
          }
        ]
        \methodname{Sex}
        \arrow[r] \arrow[dr] \arrow[ddr] \arrow[drrr, bend left=30]& \methodname{GPA}  &\\
        \methodname{Race} \arrow[d] \arrow[ur] \arrow[r] \arrow[dr] \arrow[rrr, bend right=15]& \methodname{LSAT} & |[fairnesshidden]|\methodname{Know} \arrow[ul, hiddenarrow] \arrow[l, hiddenarrow] \arrow[dl, hiddenarrow] \arrow[r, hiddenarrow, ""{name=U, description}] &  \methodname{Decile3} \\
        \methodname{Fam} \arrow[uur] \arrow[ur] \arrow[r] \arrow[urrr, bend right=30]& \methodname{FYA} \arrow[urr, bend right=20] &  \\
    \end{tikzcd}
    \caption{Assumed causal graph in \cref{sec:usecase}. 
    Only the classifiers consider \decilethree.
    }
    \label{fig:fairness-causal-graph}
\end{wrapfigure}

\paragraph{Dataset and objective.}

Our aim is to train a gradient-boosted decision tree~\citep{friedman2001greedy} on the law school dataset~\citep{wightman1998lsac}, which comprises of \num{21790} law students, that remains accurate while being fair toward the sensitive attributes of the students.
In particular, we aim to predict the decile of a student in its \nth{3} year of university, given their undergraduate and \nth{1} year grades, family income, race, and sex.

\paragraph{Experimental setup.}

First, we train \ours assuming the causal graph in \cref{fig:fairness-causal-graph}, excluding \decilethree, where all grades are affected by a common ``knowledge'' hidden confounder~\citep{kusnerCounterfactualFairness2017}.
Then, we train a predictor using as input the hidden confounder and non-sensitive exogenous variables extracted from \ours.
If, as discussed in \cref{subsec:disentanglement}, \ours successfully recovers the exogenous variables, we expect the predictor to be fair yet slightly less accurate, since \decilethree is directly affected by the sensitive attributes. 

\clearpage

\paragraph{Results.}

\begin{wrapfigure}[7]{r}{0.5\linewidth}
    \centering
    \setlength{\tabcolsep}{4pt}
    \captionof{table}{Test RMSE on \decilethree prediction and MMD of inter-group predictive distributions.} 
    \label{tab:rmse_comparison}
    \resizebox{\linewidth}{!}{
        \begin{tabular}{ccccccc} \toprule
            & {\small Unfair} & {\small Unaware} & {\small \ours} & {\small Fair $K$} & {\small Fair Add} & {\small Mean} \\ \midrule
            RMSE & 1.413 & 1.419 & 1.604 & 2.817 & 2.826 & 2.83 \\
            MMD  & 0.163 & 0.147 & 0.0054 & $10^{-5}$ & $10^{-4}$ & 0 \\
            \bottomrule
        \end{tabular}
    }
\end{wrapfigure}

\Cref{tab:rmse_comparison} provides the prediction error (RMSE) and the difference between group distributions (MMD) for the proposed \ours-based predictor, comparing with an {Unfair} predictor that uses sensitive attributes; an {Unaware} predictor that excludes sensitive attributes, and two fair predictors, {Fair K} and {Fair Add}, as initially proposed by \citet{kusnerCounterfactualFairness2017}.
We see that the proposed predictor slightly increases the error, while significantly reducing the MMD between the predicted distributions between sensitive attributes. Moreover, the other fair classifiers behave just as a baseline always predicting the average prediction.

\begin{wrapfigure}[11]{r}{0.5\linewidth}
	\centering
	\vspace{-1.5\baselineskip}
	\includegraphics[width=\linewidth]{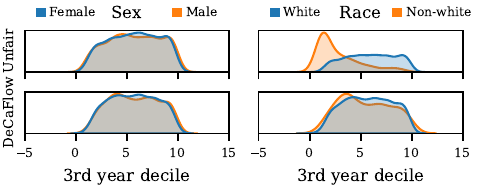}
	\captionof{figure}{\textbf{Distribution of predicted \decilethree}. A fair predictor yields similar distributions across the considered groups per attribute (\sex and \race).}
	\label{fig:fairness_dist}
\end{wrapfigure}

To provide a better intuition on the differences between predictors, we plot in \cref{fig:fairness_dist} the distributions predicted by the gradient-boosted decision tree stratified by the sensitive classes, for the Unfair and \ours-based classifiers. 
\Cref{fig:fairness_dist} shows that, while both classifiers provide similarly-distributed predictions for both sex classes, female and male, we find a qualitatively significant difference between the race classes, white and non-white, with the predictions of the Unfair classifier clearly skewed, predicting much lower deciles for the non-white population.
In contrast, the \ours-based classifier provides much similar predictions for all sensitive classes at the expense of a slightly higher prediction error.

\section{Concluding remarks}  \label{sec:conclusion}

In this work we introduced \ours, a CGM that can enable accurate estimation of interventional and counterfactual queries under hidden confounding. 
\Ours expands CNFs, preserving their properties while offering several key advantages over prior works. Namely, \ours can be applied out-of-the-box to any given causal graph and, training once per dataset, it correctly estimates a broad class of (potentially hidden-confounded) interventional \emph{and counterfactual} queries over continuous endogenous variables.
Moreover, we theoretically characterized all queries that \ours correctly estimates as those for which do-calculus yields observational or proximal-identifiable queries, extending prior results \citep{miao2018identifying, wang2021proxy} to include counterfactuals and observed common ancestors. 
Finally, we showed that \ours outperforms existing methods across a variety of settings, accurately recovering more hidden-confounded causal effects and enabling fair downstream predictions.

\paragraph{Limitations.} While \ours relaxes \emph{causal sufficiency}, it still relies on the existence of sufficiently-informative proxies. %
This condition is untestable since we do not observe the hidden confounder, but collecting additional proxies can help satisfy it~\citep{andrews2011examples}, as shown in \cref{subsec:ablation}.  
Similarly, \ours works with continuous random variables by assumption (see \cref{sec:deconfounding-causal-flows}), although \cref{sec:usecase}, \citet{javaloy2024causal} and \citet{de2024marginal} show that CNFs effectively approximate discrete distributions in practice.
Another limitation is assuming perfect knowledge of the causal graph \graph. In practice, \graph may be partially available or noisy. When graph misspecification does not involve children of the hidden confounders, \ours inherit from CNFs the ability to operate with a known causal ordering or with partially specified graphs where variables are grouped, see \citep[App.~A.2.2]{javaloy2024causal}. However, if the assumed graph incorrectly specifies the relations involving hidden confounders---and thus violating the assumptions in \cref{prop:informal}---our identifiability results under hidden confounding no longer hold, and estimates for confounded causal queries may become inaccurate. 
Alternatively, \ours could be combined with methods that jointly perform causal identification and estimation for individual queries \citep{xia2023neural}, which handle cases identifiable beyond our theory but trade scalability for flexibility.

\paragraph{Future work.} 
We believe \ours opens many intriguing venues we are excited to explore, such as expanding the range of queries it can estimate using instrumental variables~\citep{hartford2017deep}, 
applying it to settings with time-varying treatments or where multiple interventions take place,
investigating the empirical sensitivity of \ours to noisy or misspecified causal graphs \citep{oates2017repair}, or extending our framework to support soft (stochastic) interventions with dedicated evaluation protocols \citep{correa2020calculus}. We are also excited to see \ours 
applied to real-world problems such as decision support systems %
\citep{sanchez2022causal}, educational analysis \citep{murnane2010methods}, or policy making \citep{fougere2021policy}, yet always validating them with interventional data.

\begin{ack}

The authors would like to thank Luigi Gresele for useful discussions and comments which helped improving the quality of this work.
This work has been supported by the project 
\textit{``Society-Aware Machine Learning: The paradigm shift demanded by society to trust machine learning,"} funded by the European Union and led by IV (\href{https://machinelearning.uni-saarland.de/society-aware-ml/}{ERC-2021-STG, SAML}, 101040177); and the  Deutsche Forschungsgemeinschaft (DFG) grant number 389792660 as part of the Transregional Collaborative Research
Centre TRR 248: Center for Perspicuous Computing (CPEC) \href{https://perspicuous-computing.science}{(TRR~248 -- CPEC)}.
Members of Universidad Polit\'ecnica de Madrid (AA, JP and SZ) have received the funding from the \href{https://cordis.europa.eu/project/id/101172872}{SYNTHIA} project. SYNTHIA (Synthetic Data Generation framework for integrated validation of use cases and Al healthcare applications) is supported by the Innovative Health Initiative Joint Undertaking (IHI JU) under grant agreement No 101172872. Funded by the European Union, the private members, and those contributing partners of the IHI JU. 
In addition, \textit{Programa Propio UPM} funded the stay of AA  at Saarland University.
Moreover, AJ has received funding from the \textit{``UNREAL: a Unified Reasoning Layer for Trustworthy ML''} project (EP/Y023838/1) selected by the ERC and funded by UKRI EPSR. 
Views and opinions expressed are, however, those of the author(s) only and do not necessarily reflect those of the aforementioned funding agencies. Neither of the aforementioned parties can be held responsible for them. %

\end{ack}

    \bibliography{references.clean}
    \bibliographystyle{myplainnat}

    \newpage
    \onecolumn
    \restoregeometry %

    \clearpage
	\FloatBarrier
	\newpage
	\section*{NeurIPS Paper Checklist}

	\begin{enumerate}
		
		\item {\bf Claims}
		\item[] Question: Do the main claims made in the abstract and introduction accurately reflect the paper's contributions and scope?
		\item[] Answer: \answerYes{} %
		\item[] Justification: The abstract and introduction assert that \ours provides correct estimates for every do-calculus and proximal identifiable causal query, including counterfactuals, under hidden confounding and outperforms prior models. These points are rigorously proven in the theory sections (\cref{sec:theoretica_results}) and demonstrated by comprehensive empirical results (\cref{sec:experiments}).
		
		\item {\bf Limitations}
		\item[] Question: Does the paper discuss the limitations of the work performed by the authors?
		\item[] Answer: \answerYes{} %
		\item[] Justification: A dedicated ``Limitations'' paragraph at the end of \cref{sec:conclusion} explicitly states that \ours depends on sufficiently informative variables---an untestable assumption---and on a $C^1$-diffeomorphic confounded SCM, thereby clarifying when the method may fail and why.
		
		\item {\bf Theory assumptions and proofs}
		\item[] Question: For each theoretical result, does the paper provide the full set of assumptions and a complete (and correct) proof?
		\item[] Answer: \answerYes{} %
		\item[] Justification: Every theorem, proposition and corollary in \cref{sec:theoretica_results} is stated with explicit numbered assumptions and the correspoding formal proofs are given in \cref{app:sec:proofs} and cross referenced. As an example, \cref{app:prop:causal-query-identifiability} is the formal version of \cref{prop:informal}, which includes a list of independence, completeness and regularity conditions and then supplies a step-by-step proof.
		
		\item {\bf Experimental result reproducibility}
		\item[] Question: Does the paper fully disclose all the information needed to reproduce the main experimental results of the paper to the extent that it affects the main claims and/or conclusions of the paper (regardless of whether the code and data are provided or not)?
		\item[] Answer: \answerYes{} %
		\item[] Justification: The main paper explicitly refers to ``all experiments details'' to \cref{sec:decaf} and \cref{app:sec:implementation}, \cref{app:sec:additional-results} and \cref{app:sec:do-operator}, where it specified the data-generation pipelines, the causal graphs, the architecture, the ELBO training objective with its regularization process, the do-operator, the metrics and the evaluation protocol In addition, the authors commit to releasing the full codebase, together with the hyper-parameter search and seeds.

		\item {\bf Open access to data and code}
		\item[] Question: Does the paper provide open access to the data and code, with sufficient instructions to faithfully reproduce the main experimental results, as described in supplemental material?
		\item[] Answer: \answerYes{} %
		\item[] Justification:
        he datasets employed are publicly available or fully described, a user-friendly implementation of \ours is available in the link of the introduction of this paper (\href{https://github.com/aalmodovares/DeCaFlow}{\methodname{github.com/aalmodovares/DeCaFlow}}), as well as examples of use,
        identifiability check algorithms and visualizations. In addition, we offer the whole training, hp tuning and
        validation pipeline, for reproducibility, under request to correspondence authors.

		\item {\bf Experimental setting/details}
		\item[] Question: Does the paper specify all the training and test details (e.g., data splits, hyperparameters, how they were chosen, type of optimizer, etc.) necessary to understand the results?
		\item[] Answer: \answerYes{} %
		\item[] Justification:  \cref{app:sec:additional-results} is entirely devoted to experimental details, describing for each dataset the generation pipeline, train-test splits, the number of runs, the interventions percentiles and exact metrics. In addition, \cref{app:sec:implementation} supplements with implementation hyper-parameters, warm-up schedule, posterior factorization and masking strategy.
		
		\item {\bf Experiment statistical significance}
		\item[] Question: Does the paper report error bars suitably and correctly defined or other appropriate information about the statistical significance of the experiments?
		\item[] Answer: \answerYes{} %
		\item[] Justification: Every key figure and table includes statistical uncertainty. \cref{fig:results_sachs_and_ecoli} plots boxenplots (with all percentiles), and tables provide mean and standard deviation of all metrics across all seeds and all evaluated causal effects. Tables also include significantly better results related with statistical tests and intervals included in the captions.
		
		\item {\bf Experiments compute resources}
		\item[] Question: For each experiment, does the paper provide sufficient information on the computer resources (type of compute workers, memory, time of execution) needed to reproduce the experiments?
		\item[] Answer: \answerYes{} %
		\item[] Justification: Details about the execution resources and times are provided in \cref{app:sec:implementation},
		
		\item {\bf Code of ethics}
		\item[] Question: Does the research conducted in the paper conform, in every respect, with the NeurIPS Code of Ethics \url{https://neurips.cc/public/EthicsGuidelines}?
		\item[] Answer: \answerYes{} %
		\item[] Justification: All experiments rely on publicly available or semi-synthetic benchmark data (Sachs protein signalling, Ecoli70 gene network, and the anonymised LSAC law-school dataset) and introduce no new personal data collection, human-subject interaction, or high-risk model release; the work explicitly addresses fairness (\cref{sec:experiments}) and reports moderate compute usage, thereby avoiding the privacy, discrimination, security, or environmental concerns enumerated in the NeurIPS Code of Ethics.

		\item {\bf Broader impacts}
		\item[] Question: Does the paper discuss both potential positive societal impacts and negative societal impacts of the work performed?
		\item[] Answer: \answerYes{} %
		\item[] Justification: We explain how DeCaFlow can enable better decision-making in domains such as healthcare and education while cautioning that causal assumptions must be validated, especially in sensitive applications, and state that the method introduces no additional ethical risks beyond those already known for causal-inference models.
		
		\item {\bf Safeguards}
		\item[] Question: Does the paper describe safeguards that have been put in place for responsible release of data or models that have a high risk for misuse (e.g., pretrained language models, image generators, or scraped datasets)?
		\item[] Answer: \answerNA{}{} %
		\item[] Justification: The release comprises only the DeCaFlow training and evaluation code plus lightweight models trained on publicly available or semi-synthetic benchmarks; because no large pretrained generative models or scraped datasets with dual-use potential are distributed, special safeguards are unnecessary.
		
		\item {\bf Licenses for existing assets}
		\item[] Question: Are the creators or original owners of assets (e.g., code, data, models), used in the paper, properly credited and are the license and terms of use explicitly mentioned and properly respected?
		\item[] Answer: \answerYes{} %
		\item[] Justification: The licenses and all the copyright information will be included in every asset in the code.
		
		\item {\bf New assets}
		\item[] Question: Are new assets introduced in the paper well documented and is the documentation provided alongside the assets?
		\item[] Answer: \answerYes{} %
		\item[] Justification: The paper introduces a new, fully self-contained DeCaFlow repository released under the General Public License. No new datasets or personal data are created, so consent and privacy disclosures are unnecessary.
		
		\item {\bf Crowdsourcing and research with human subjects}
		\item[] Question: For crowdsourcing experiments and research with human subjects, does the paper include the full text of instructions given to participants and screenshots, if applicable, as well as details about compensation (if any)? 
		\item[] Answer: \answerNA{} %
		\item[] Justification: The study does not involve any crowdsourcing or prospective research with human participants; it relies exclusively on pre-existing, publicly available or synthetic datasets (Sachs, Ecoli70, LSAC), so participant instructions, screenshots, and compensation details are not applicable.
		
		\item {\bf Institutional review board (IRB) approvals or equivalent for research with human subjects}
		\item[] Question: Does the paper describe potential risks incurred by study participants, whether such risks were disclosed to the subjects, and whether Institutional Review Board (IRB) approvals (or an equivalent approval/review based on the requirements of your country or institution) were obtained?
		\item[] Answer: \answerNA{} %
		\item[] Justification: The work uses only pre-existing benchmark datasets—Sachs protein-signalling, a semi-synthetic Ecoli70 gene network, and the publicly released LSAC law-school dataset—without recruiting new participants or collecting personal data, so human-subjects review and IRB disclosure are not applicable.
		
		\item {\bf Declaration of LLM usage}
		\item[] Question: Does the paper describe the usage of LLMs if it is an important, original, or non-standard component of the core methods in this research? Note that if the LLM is used only for writing, editing, or formatting purposes and does not impact the core methodology, scientific rigorousness, or originality of the research, declaration is not required.
		\item[] Answer: \answerNA{} %
		\item[] Justification: The authors have not used LLMs for important tasks of the paper.
		
	\end{enumerate}

    \clearpage
    \renewcommand{\partname}{}  %
	\part{Appendix} %
	\parttoc %
    
    \clearpage
    \appendix

    \counterwithin{table}{section}
    \counterwithin{figure}{section}
    \renewcommand{\thetable}{\thesection.\arabic{table}}
    \renewcommand{\thefigure}{\thesection.\arabic{figure}}

\section{Causal identifiability}
\label{app:sec:proofs}

\subsection{Model identifiability}
\label{app:subsec:uncounded-identifiability}

    We briefly discuss the identifiability (in the sense of \citet{pmlr-v206-xi23a}) of those variables that are indirectly confounded by \varhidden or not confounded at all, \ie, of those variables that are not children of any hidden confounder.
    As we discuss now, we can reduce our SCM (\cref{def:scm}) to a conditional one that only models these aforementioned variables, recovering the identifiability guarantees from \citet{javaloy2024causal}.
        To prove model identifiability, we resort to what we call the induced conditional SCM, which intuitively represents the original SCM where we restrict our view to a subset of variables, and assume the rest of the variables are given.
        
        \begin{definition}[Induced conditional SCM]
            Given a SCM $\scm = (\funcb, \distribution[\varexo], \distribution[\varhidden])$, and a subset of observed variables $\varobs' \subset \varobs$, we define the \emph{induced conditional SCM of \scm given $\varobs'$}, denoted by $\scm_{\cond{\varobs'}}$, to the SCM result of having observed $\varobs'$, and where causal generators and exogenous variables are restricted to only those associated with the unconditioned variables, \ie, $\varobs \setminus \varobs'$.
        \end{definition}

    \begin{figure}[h]
        \centering
        \begin{subfigure}[b]{0.45\linewidth}
            \centering
            \begin{tikzcd}[
                cells={nodes={main node}}, column sep=tiny, row sep=small
            ]
                \evarobs_1 \arrow[dr] \arrow[dd] & & |[hidden]|\varhidden \arrow[hidden, ll] \arrow[hidden, rr] \arrow[hidden, dd] & & \evarobs_2  \arrow[dl] \arrow[dd] \\
                & \evarobs_3 \arrow[rr] & & \evarobs_5 \arrow[dr] \\
                \evarobs_6 \arrow[ur] & & \evarobs_7 \arrow[ul] \arrow[ur] \arrow[ll] \arrow[rr] & & \evarobs_8
            \end{tikzcd}
            \caption{Confounded SCM.}
        \end{subfigure}
        \begin{subfigure}[b]{0.45\linewidth}
            \centering
            \begin{tikzcd}[
                cells={nodes={main node}}, column sep=tiny, row sep=small
                ]
                |[given variable=\evarobs_1]| \bullet \arrow[dr] \arrow[dd] & & & & |[given variable=\evarobs_2]| \bullet  \arrow[dl] \arrow[dd] \\
                & \evarobs_3 \arrow[rr] & & \evarobs_5 \arrow[dr] \\
                \evarobs_6 \arrow[ur] & & |[given variable=\evarobs_7]| \bullet \arrow[ul] \arrow[ur] \arrow[ll] \arrow[rr] & & \evarobs_8
            \end{tikzcd}
            \caption{Conditional unconfounded SCM.}
        \end{subfigure}
        \caption{Example of: \captiona a confounded SCM \scm; and \captionb its induced conditional counterpart, $\scm_{\cond{\varobs'}}$ where the children of the hidden confounder are observed and fixed, $\varobs' = \varchildren{\varhidden} = \{\evarobs_1, \evarobs_2, \evarobs_7\}$. Note that $\scm_{\cond{\varobs'}}$ does not exhibit hidden confounding.}
        \label{fig:example-conditionally-unconfounded graph}
    \end{figure}

        We provide a visual depiction of this idea in \cref{fig:example-conditionally-unconfounded graph}.
        Using this definition, we can observe that, if we were to condition on the children of the hidden confounder, we would be left with a (conditional) \emph{unconfounded SCM}, as the influence of the hidden confounder has been completely blocked by conditioning on its children.
        Now, if we have two models that perfectly match their marginal distributions, this means that they perfectly match their induced conditional SCM, no matter which value we observed for \varchildren{\varhidden}, and we can thus leverage existing results from \citet{javaloy2024causal} for unconfounded SCMs. More specifically:

\begin{corollary} \label{app:cor:extend-flow}
    Assume that we have two SCMs $\scm \coloneqq (\funcb, \distribution[\varexo], \distribution[\varhidden])$ and $\pred\scm \coloneqq (\pred\funcb, \distribution[\pred\varexo], \distribution[\pred\varhidden])$ that are Markov-equivalent---\ie, they induce the same causal graph---and which coincide in their marginal distributions, $p(\varobs) \aeeq \pred p(\varobs)$. 
    Then, both SCMs, restricted to every variable other than \varchildren{\varhidden}, are equal up to an element-wise transformation of the exogenous distributions.
\end{corollary}
\begin{proof}
    The proof follows almost directly from \citep[Theorem 1]{javaloy2024causal}. 
    First, note that the two induced conditional SCMs are no longer influenced by \varhidden once that we have observed a specific realization of \varchildren{\varhidden}, so that we can drop \varhidden from their structure, \ie, we can rewrite them instead as unconfounded SCMs, $\scm_{\cond{\varchildren{\varhidden}}} = (\funcb[\cond{\varchildren{\varhidden}}], \distribution[\varexo\cond{\varchildren{\varhidden}}])$ and $\pred\scm_{\cond{\varchildren{\varhidden}}} = (\predfuncb[\cond{\varchildren{\varhidden}}], \distribution[\pred\varexo\cond{\varchildren{\varhidden}}])$\,.
    To ease notation, let us call $\varobs^\mathsf{c} \coloneqq \varobs \setminus \varchildren{\varhidden}$ the variables that are not children of \varhidden.
    
    Next, note that for almost every realization of \varchildren{\varhidden}, we have that $p(\giventhat{\varobs^\mathsf{c}}{\varchildren{\varhidden}}) \aeeq \pred p(\giventhat{\varobs^\mathsf{c}}{\varchildren{\varhidden}})$ since $p(\varobs) \aeeq \pred p(\varobs)$ by assumption and $p(\varobs) = p(\giventhat{\varobs^\mathsf{c}}{\varchildren{\varhidden}}) p(\varchildren{\varhidden})$\,.
    As a result, for each realization of \varchildren{\varhidden} we can apply Theorem 1 of \citet{javaloy2024causal}, which yields that the two induced conditional SCMs are equal up to an element-wise transformation of the exogenous distribution.
    
    Finally, since the causal generators and exogenous distributions of the induced SCMs are, for almost every \varchildren{\varhidden}, identical to their counterparts in the original SCMs (as we have just discarded those components associated with \varchildren{\varhidden}), we get that, those elements in both SCMs associated with $\varobs^\mathsf{c}$, are identical up to said (possibly \varchildren{\varhidden}-dependent) component-wise transformation.
\end{proof}

\subsection{Query identifiability}
\label{app:subsec:general-case-proof}

    We now prove the identifiability of the causal queries considered in the main text. 
    To this end, one key property that we will use in the following is that of completeness (as, \eg, in the work of \citet{wang2021proxy}). 
    Intuitively, we say that a random variable \varhidden is complete given another random variable \varproxyo if ``any infinitesimal change in \varhidden is accompanied by variability in \varproxyo'' \citep{miao2023identifying}, yielding enough information to recover the posterior distribution of \varhidden. This concept is similar in spirit to that of variability in the case of discrete random variables \citep{nasr2023counterfactual}. In practice, completeness is more likely to be achieved the more proxies we measure \citep{andrews2011examples}.  %

    \begin{definition}[Completeness] \label{def:completeness}
        We say that a random variable \varhidden is complete given \varproxyo for almost all \varcov if, for any square-integrable function $g(\cdot)$ and almost all \varcov,
        $\int g(\varhidden, \varcov) p(\giventhat{\varhidden}{\varcov, \varproxyo}) \dif\varhidden = 0$
        for almost all \varproxyo, if and only if $g(\varhidden, \varcov) = 0$ for almost all \varhidden.
    \end{definition}

The following proposition (informally simplified in \cref{prop:informal}) is a generalization of the results previously presented by \citet{miao2018identifying} and \citet{wang2021proxy}, where we include an additional covariate \varcov to the causal query, and make no implicit assumptions on the causal graph allowing, \eg, for the treatment and outcome variables to hame some observed parents in common. 
However, note that \varcov cannot be a collider (\eg, forming a subgraph of the form $\varproxyo\rightarrow\varcov\leftarrow\varoutcome$). Otherwise, conditioning on \varcov would make independent variables dependent (in the example, \varoutcome and \varproxyo), and the causal effect of \vartreat on \varoutcome would not be identifiable.

\begin{proposition}[Query identifiability]  \label{app:prop:causal-query-identifiability}
    Given two SCMs $\scm \coloneqq (\funcb, \distribution[\varexo], \distribution[\varhidden])$ and $\pred\scm \coloneqq (\pred\funcb, \distribution[\pred\varexo], \distribution[\pred\varhidden])$, assume that they are Markov-equivalent---\ie, they induce the same causal graph---and which coincide in their marginal distributions, $p(\varobs) \aeeq \pred p(\varobs)$.
    Then, they compute the same causal query, $p(\giventhat{\varoutcome}{\doop(\vartreat), \varcov}) = \pred p(\giventhat{\varoutcome}{\doop(\vartreat), \varcov})$, where $\varoutcome, \vartreat, \varcov \subset \varobs$%
    , if there exists two proxies $\varproxyt, \varproxyo \subset \varobs$ and $\varblock \subset \varobs$, none of them overlapping nor containing variables from the previous subsets, \st:
    \begin{enumerate}[i)]
        \item \varproxyt is conditionally independent of $(\vartreat, \varproxyo)$ given \varblock, \varhidden and \varcov. That is, $\giventhat{\varproxyt \indep (\vartreat, \varproxyo)}{\varblock, \varhidden, \varcov}$\,.
        
        \item \varproxyo is conditionally independent of \varoutcome given \vartreat, \varblock, \varhidden and \varcov. That is, $\giventhat{\varoutcome \indep \varproxyo}{\vartreat, \varblock, \varhidden, \varcov}$\,.

        \item $(\varblock, \varhidden)$ forms a valid adjustment set for the query $p(\giventhat{\varoutcome}{\doop(\vartreat), \varcov})$. That is, %
        given \varcov, they are independent of \vartreat after severing any incoming edges to it, $\giventhat{\vartreat \indep_{\graph_{\overline{\vartreat}}} (\varblock, \varhidden)}{\varcov}$\,, and they block every backdoor path from \vartreat to \varoutcome.
        
        \item \varhidden is complete given \varproxyo for almost all \vartreat, \varblock, and \varcov,
        \item \pred\varhidden is complete given \varproxyt for almost all \varblock and \varcov, %
    \end{enumerate}
    and the following regularity conditions also hold:
    \begin{enumerate}[i)]
        \setcounter{enumi}{5}
        \item $\iint \pred p(\giventhat{\pred\varhidden}{\varproxyt, \varblock, \varcov}) \pred p(\giventhat{\varproxyt}{\pred\varhidden, \varblock, \varcov}) \dif\pred\varhidden\dif\varproxyt < \infty$ for all \varblock, \varcov, and
        \item $\int \pred p(\giventhat{\varoutcome}{\vartreat, \varblock, \pred\varhidden, \varcov})^2 \pred p(\giventhat{\pred\varhidden}{\varblock, \varcov}) \dif \pred\varhidden < \infty$ for all \vartreat, \varblock, and \varcov.
    \end{enumerate}
\end{proposition}

\begin{proof}
    First, note that the first three independence assumptions hold for both models, \scm and $\pred\scm$, as they induce the same causal graph.
    Following the same arguments as \citet[Proposition 1]{miao2018identifying}, we have that assumptions \itemv, \itemvi, and \itemvii guarantee the existence of a function $\pred h$ such that it solves the integral equation over $\pred\scm$,
    \begin{equation}
        \pred p(\giventhat{\varoutcome}{\vartreat, \varblock, \pred\varhidden, \varcov}) = \int \pred h(\varoutcome, \vartreat, \varblock, \varproxyt, \varcov) \pred p(\giventhat{\varproxyt}{\varblock, \pred \varhidden, \varcov}) \dif \varproxyt \,,\label{app:eq:assumed-solution}
    \end{equation}
    since assumption \itemvi ensures that the conditional expectation operator is compact \citep{carrasco2007linear}, assumption \itemv that all square-integrable functions are in the image of the operator (\ie, the operator is surjective), and assumption \itemvii that $\pred p(\giventhat{\varoutcome}{\vartreat, \varblock, \pred\varhidden, \varcov})$ is indeed part of the image.
    
    We can show that $\pred h$ also solves a similar integral equation, this time over the other SCM, \scm, as follows:
    \begin{align}
        p(\giventhat{\varoutcome}{\vartreat, \varblock, \varproxyo, \varcov}) 
        &= \pred p(\giventhat{\varoutcome}{\vartreat, \varblock, \varproxyo, \varcov})  && \eqcomment{equal marginals} \\
        &= \int \pred p(\giventhat{\varoutcome}{\vartreat, \varblock, \varproxyo, \pred\varhidden, \varcov}) \pred p(\giventhat{\pred\varhidden}{\vartreat, \varblock, \varproxyo, \varcov}) \dif \pred\varhidden && \eqcomment{augment with $\pred\varhidden$} \\
        &= \int \pred p(\giventhat{\varoutcome}{\vartreat, \varblock, \pred\varhidden, \varcov}) \pred p(\giventhat{\pred\varhidden}{\vartreat, \varblock, \varproxyo, \varcov}) \dif \pred\varhidden && \eqcomment{assumption \itemii} \\
        &= \iint \pred h(\varoutcome, \vartreat, \varblock, \varproxyt, \varcov) \pred p(\giventhat{\varproxyt}{\varblock, \pred\varhidden, \varcov}) \pred p(\giventhat{\pred\varhidden}{\vartreat, \varblock, \varproxyo, \varcov}) \dif \pred\varhidden \dif\varproxyt && \eqcomment{plug \cref{app:eq:assumed-solution}} \\
        &= \iint \pred h(\varoutcome, \vartreat, \varblock, \varproxyt, \varcov) \pred p(\giventhat{\varproxyt}{\varblock, \pred\varhidden, \vartreat, \varproxyo, \varcov}) \pred p(\giventhat{\pred\varhidden}{\vartreat, \varblock, \varproxyo, \varcov}) \dif \pred\varhidden \dif\varproxyt && \eqcomment{assumption \itemi} \\
        &= \int \pred h(\varoutcome, \vartreat, \varblock, \varproxyt, \varcov) p(\giventhat{\varproxyt}{\vartreat, \varblock, \varproxyo, \varcov}) \dif \varproxyt \,. && \eqcomment{equal marginals} 
        \label{app:eq:relate-both-models}
    \end{align}

    Note that \cref{app:eq:relate-both-models} is a Fredholm equation of the first kind that is implicitly solved by modeling the observational data. 
    Similarly, we can relate the expression for the interventional distribution of both models:
    \begin{align}
        \pred p (\giventhat{\varoutcome}{\doop(\vartreat), \varcov}) 
        &= \int \pred p(\giventhat{\varoutcome}{\doop(\vartreat), \varblock, \pred\varhidden, \varcov}) \pred p(\giventhat{\varblock, \pred\varhidden}{\varcov}) \dif\varblock \dif \pred\varhidden && \eqcomment{augment and ass. \itemiii} \label{app:prop:proof:eq-augment-zb} \\
        &= \int \pred p(\giventhat{\varoutcome}{\vartreat, \varblock, \pred\varhidden, \varcov}) \pred p(\giventhat{\varblock, \pred\varhidden}{\varcov}) \dif\varblock \dif \pred\varhidden && \eqcomment{backdoor criterion} \label{app:eq:backdoor}\\
        &= \iint \pred h(\varoutcome, \vartreat, \varblock, \varproxyt, \varcov) \pred p(\giventhat{\varproxyt}{\varblock, \pred \varhidden, \varcov}) \pred p(\giventhat{\varblock, \pred\varhidden}{\varcov}) \dif\varblock \dif\varproxyt \dif \pred\varhidden && \eqcomment{plug \cref{app:eq:assumed-solution}} \\
        &= \int \pred h(\varoutcome, \vartreat, \varblock, \varproxyt, \varcov) p(\giventhat{\varblock, \varproxyt}{\varcov}) \dif\varblock \dif\varproxyt && \eqcomment{equal marginals}\\
        &= p(\giventhat{\varoutcome}{\doop(\vartreat), \varcov}) \,, 
        \label{app:eq:relating-both-interventions}
    \end{align}
    where the last equality is a consequence of \cref{app:eq:relate-both-models} as we will show now.
    More specifically, we have that
    \begin{align}
        p(\giventhat{\varoutcome}{\vartreat, \varblock, \varproxyo, \varcov}) 
        &= \int \pred h(\varoutcome, \vartreat, \varblock, \varproxyt, \varcov) p(\giventhat{\varproxyt}{\vartreat, \varblock, \varproxyo, \varcov}) \dif \varproxyt && \eqcomment{\cref{app:eq:relate-both-models}} \\
        &= \iint \pred h(\varoutcome, \vartreat, \varblock, \varproxyt, \varcov) p(\giventhat{\varproxyt}{\varblock, \varhidden, \vartreat, \varproxyo, \varcov}) p(\giventhat{\varhidden}{\vartreat, \varblock, \varproxyo, \varcov}) \dif \varproxyt \dif \varhidden \,, && \eqcomment{augment with \varhidden} \\
        &= \iint \pred h(\varoutcome, \vartreat, \varblock, \varproxyt, \varcov) p(\giventhat{\varproxyt}{\varblock, \varhidden, \varcov}) p(\giventhat{\varhidden}{\vartreat, \varblock, \varproxyo, \varcov}) \dif \varproxyt \dif \varhidden \,. && \eqcomment{assumption \itemi}
    \end{align}
    
    Similarly, we have that
    \begin{align}
        p(\giventhat{\varoutcome}{\vartreat, \varblock, \varproxyo, \varcov}) &= \int p(\giventhat{\varoutcome}{\vartreat, \varblock, \varproxyo, \varhidden, \varcov}) p(\giventhat{\varhidden}{\vartreat, \varblock, \varproxyo, \varcov}) \dif \varhidden && \eqcomment{augment with \varhidden} \\ 
        &= \int p(\giventhat{\varoutcome}{\vartreat, \varblock, \varhidden, \varcov}) p(\giventhat{\varhidden}{\vartreat, \varblock, \varproxyo, \varcov}) \dif \varhidden \,. && \eqcomment{assumption \itemii} 
    \end{align}
    Now, equating both expressions we have that
    \begin{align}
        0 &= \iint \left\{  p(\giventhat{\varoutcome}{\vartreat, \varblock, \varhidden, \varcov}) - \int \pred h(\varoutcome, \vartreat, \varblock, \varproxyt, \varcov) p(\giventhat{\varproxyt}{\varblock, \varhidden, \varcov}) \dif \varproxyt \right\} p(\giventhat{\varhidden}{\vartreat, \varblock, \varproxyo, \varcov}) \dif \varhidden \,,
    \end{align}
    which, due to assumption \itemiv, implies that
    \begin{align}
        p(\giventhat{\varoutcome}{\vartreat, \varblock, \varhidden, \varcov}) \aeeq \int \pred h(\varoutcome, \vartreat, \varblock, \varproxyt, \varcov) p(\giventhat{\varproxyt}{\varblock, \varhidden, \varcov}) \dif \varproxyt \,.
        \label{app:eq:w-substitute}
    \end{align}
    
    Finally, putting all together we see that we can write the interventional distribution of the original model using $\pred h$,
    \begin{align}
        p(\giventhat{\varoutcome}{\doop(\vartreat), \varcov}) 
        &= \iint p(\giventhat{\varoutcome}{\doop(\vartreat), \varblock, \varhidden, \varcov}) p(\giventhat{\varblock, \varhidden}{\varcov}) \dif\varblock\dif\varhidden && \eqcomment{augment and assumption \itemiii} \\
        &= \iint p(\giventhat{\varoutcome}{\vartreat, \varblock, \varhidden, \varcov}) p(\giventhat{\varblock, \varhidden}{\varcov}) \dif\varblock\dif\varhidden && \eqcomment{backdoor criterion} \\
        &= \iint \pred h(\varoutcome, \vartreat, \varblock, \varproxyt, \varcov) p(\giventhat{\varproxyt}{\varblock, \varhidden, \varcov}) p(\giventhat{\varblock, \varhidden}{\varcov}) \dif\varblock \dif\varhidden \dif \varproxyt && \eqcomment{\cref{app:eq:w-substitute}} \\
        &= \int \pred h(\varoutcome, \vartreat, \varblock, \varproxyt, \varcov) p(\giventhat{\varblock, \varproxyt}{\varcov}) \dif\varblock  \dif\varproxyt \,, && \eqcomment{equal marginals}
    \end{align}
    which justifies the last equality in \cref{app:eq:relating-both-interventions}.
\end{proof}

In \cref{app:prop:causal-query-identifiability},
assumptions \itemi-\itemiii regard the conditional independence of different variables in the causal graph, which can be directly verified given a faithful causal graph. Assumptions \itemiv and \itemv regard the information that \varproxyt and \varproxyo contain of the hidden confounders which, intuitively, means that the posterior of the hidden confounder varies enough as we change the values of \varproxyt and \varproxyo, in order to properly perform inference on it. This assumption is harder to verify but, as we show in \cref{app:sec:additional-results-ablation}, the more proxy variables we have, the better the estimation of the hidden confounder's effect and the more accurate the causal query estimation is. Finally, assumptions \itemvi and \itemvii are standard regularity conditions~\citep{miao2018identifying, wang2021proxy} that are (almost surely) fulfilled in practice, as long as the random variables are well behaved, \eg, having finite moments. Such conditions are typically satisfied by most continuous and discrete distributions used in probabilistic modeling, including Gaussian, exponential family, and bounded-support distributions, making them mild and non-restrictive assumptions. 

    \begin{wrapfigure}[10]{r}{.25\linewidth}
        \vspace{-.5\baselineskip}
        \centering
            \begin{tikzcd}[
        cells={nodes={main node}}, column sep=tiny, row sep=small
        ]
        \varproxyo \arrow[dr] & & |[hidden]|\varhidden \arrow[hidden, ll] \arrow[hidden, rr] \arrow[hidden, dl] \arrow[hidden, dr] \arrow[hidden, dd] & & \varproxyt \arrow[dl] \\
        & \vartreat \arrow[rr] & & \varoutcome \\
        & & \varblock \arrow[ul] \arrow[ur]
    \end{tikzcd}
        \caption{Example for which \cref{app:prop:causal-query-identifiability} applies, and where $\varblock\neq\emptyset$\equationPunctuation{.}}
        \label{app:fig:miao-block-graph}
    \end{wrapfigure}
    Using a causal graph similar to the one presented by \citet{miao2018identifying}, we now provide some intuition on the semantics of each random variable in \cref{app:prop:causal-query-identifiability}.
    More specifically, consider the causal graph that we depict in \cref{app:fig:miao-block-graph}, and say that we want to check if the causal query $p(\giventhat{\varoutcome}{\doop(\vartreat)})$ is identifiable (note that this the same query as in \cref{app:prop:causal-query-identifiability} but with $\varcov = \emptyset$). As it is common in the causal inference literature~\citep{peters2017elements, spirtes2001causation},
    \vartreat and \varoutcome represent the treatment and outcome random variables. 
    More specific to \cref{app:prop:causal-query-identifiability} are \varproxyt and \varproxyo. 
    Here, \varproxyt is a proxy variable whose role is that of distinguishing the information from \varhidden and other variables, to reconstruct the information of \varhidden and block the backdoor path that \varhidden would usually block. 
    Similarly, the variable \varproxyo is another proxy variable which, in this case, serves the purpose of verifying that the substitute formed with \varproxyt is indeed a good one.
    Finally, the variable \varblock serves the purpose of blocking all the remaining backdoor paths that \varhidden may not block, so that we can apply the backdoor criterion. %

Moreover, note that for all interventional queries we let \varcov be the empty set, similar to the results proved by \citet{miao2018identifying} and \citet{wang2021proxy}. 
We will consider cases when \varcov is not empty later in \cref{app:subsec:cf-identifiablity} to prove counterfactual identifiability. 
Note also that \cref{app:prop:causal-query-identifiability} reduces to previous results when $\varcov = \varblock = \emptyset$\,.

We now turn our attention towards proving \cref{cor:do-calculus}, \ie, towards broadening the concept of query identifiability by introducing \cref{app:prop:causal-query-identifiability} as a base case of do-calculus. 
To this end, we introduce the concept of a \emph{hedge} which will be use later, but we still strongly recommend reading the work by \citet{shpitser2006identification}.

\begin{definition}[Hedge, {\citep[Def. 6]{shpitser2006identification}}]
    Let $\varoutcome, \vartreat \subset \varobs$ be disjoint sets of variables in \graph. Let $F$, $F'$ be $\vr$-rooted C-forests (see \citep[Def. 5]{shpitser2006identification}) such that $F \cap \vartreat \neq \emptyset$, $F' \cap \vartreat = \emptyset$, $F' \subset F$, and $\vr$ is a subset of the ancestors of \varoutcome after severing the incoming edges of \vartreat. Then $F$ and $F'$ form a hedge for $p(\giventhat{\varoutcome}{\doop(\vartreat)})$ in \graph.
\end{definition}

\queryreduction*
\begin{proof}
	With the additional notion of proxy-identifiability provided by \cref{app:prop:causal-query-identifiability} (informally presented in \cref{prop:informal}), the result is just a consequence of applying the identifiability algorithm provided by \citet{shpitser2006identification}. See also \citep{huang2012pearl,tikka2017identifying} for other references.

    Since the do-calculus rules are complete in the classical sense of identifiability, a query is not identifiable if the aforementioned algorithm yields a \texttt{FAIL} status (\ie, it executes line 5 of Figure 3 in \citep{shpitser2006identification}).
    If that is the case, then it means that, at the specific recursive call for which the algorithm failed, the local graph \graph contains a hedge and the interventional query $p(\giventhat{\varoutcome}{\doop(\vartreat)})$ is not identifiable in the classical sense.

    Crucially, this hedge $(F, F')$ expresses the inability of identifying an interventional query of the form $p(\giventhat{\vr}{\doop(\vartreat')})$ where the root $\vr$ is a subset of ancestors of $\varoutcome' \subseteq \varoutcome$ and $\vartreat' \subseteq \vartreat$\equationPunctuation{.}
    Then, this local query can still be proxy-identifiable if \cref{app:prop:causal-query-identifiability} can be applied, and thus we can continue running the identification algorithm.
    
    The stated result is then a consequence of successfully applying the logic above each time we find a \texttt{FAIL} status, yielding a final \texttt{FAIL} status otherwise.
\end{proof}

To be even more explicit regarding the identifiability of the queries proven in corollary above,
let us call \scm the original SCM as usual, and $\pred\scm$ another SCM inducing the same causal graph as \scm and which matches the observational marginal distribution of \scm, \ie, $p(\varobs) \aeeq \pred p(\varobs)$. Then, the output of the identifiability algorithm from the corollary above \textit{for both SCMs} will be two identical expressions \texttt{EXP} composed of sum, integrals, and products of observational quantities (\ie, marginals and conditionals of subsets of \varobs) as well as proxy-identifiable queries of the form $p(\giventhat{\varoutcome}{\doop(\vartreat)})$ as in \cref{app:prop:causal-query-identifiability}. Therefore,
\begin{equation}
    \query(\scm) = \texttt{EXP}(\scm) = \texttt{EXP}(\pred\scm) = \query(\pred\scm) \equationPunctuation{,}
\end{equation}
where the second equality is a consequence of both SCMs having equal observational distributions (and thus any other quantity than can derived exclusively from $p(\varobs)$) and of applying \cref{app:prop:causal-query-identifiability} for any interventional query that appears in the expression.

\decafscm*
\begin{proof}
    The proof is a direct consequence of the corollary above and the fact that we can interpret \ours as a dense parametric family of confounded SCMs inducing the same causal graph as \scm (similar to the interpretation of \citet{javaloy2024causal} as bijective SCMs) by considering the triplet $\scm_\thetab \coloneqq (\flow^{-1}, \distribution[\varexo], \distribution[\varhidden])$, where $\flow^{-1}$ is the inverse of the generative network that transforms \varexo into \varobs given \varhidden. This family being dense is a consequence of the generative networks forming a family of universal density approximators \citep{papamakarios2021normalizing,javaloy2024causal}.
\end{proof}

To be completely exhaustive, in the following we explore the general proposition \cref{app:prop:causal-query-identifiability} on all scenarios where \vartreat and \varoutcome may or may not be directly caused by the hidden confounder, as we show in the following subsections.

\subsubsection{Fully hidden-confounded case}

In the case where both variables are children of \varhidden, we must see whether we can apply do-calculus with \cref{app:prop:causal-query-identifiability} as an additional base case, as described in \cref{cor:do-calculus}. 

\subsubsection{Hidden-unconfounded case}

Assume the case where neither \vartreat nor \varoutcome are children of the hidden confounder, \ie, $\varoutcome,\vartreat \notin \varchildren{\varhidden}$\,. 
In this case, the proof of \cref{app:prop:causal-query-identifiability} can be simplified and drop the requirement of finding valid proxy variables.

\begin{corollary}%
\label{app:cor:unconfounded-case}
    Given two SCMs $\scm \coloneqq (\funcb, \distribution[\varexo], \distribution[\varhidden])$ and $\pred\scm \coloneqq (\pred\funcb, \distribution[\pred\varexo], \distribution[\pred\varhidden])$, assume that they are Markov-equivalent---\ie, they induce the same causal graph---and coincide in their marginal distributions, $p(\varobs) \aeeq \pred p(\varobs)$.
    If $\varoutcome, \vartreat \notin \varchildren{\varhidden}$\,,
    then, $p(\giventhat{\varoutcome}{\doop(\vartreat), \varcov}) = \pred p(\giventhat{\varoutcome}{\doop(\vartreat), \varcov})$, where $\varoutcome, \vartreat, \varcov \subset \varobs$\,.
\end{corollary}
\begin{proof}
    The proof follows directly by applying \cref{app:prop:causal-query-identifiability} with the minimal subset $\varblock \subset \varparents{\vartreat} \setminus \{ \varcov \}$ that blocks all the backdoor paths, and by noticing that in this case there is no need to use the variables \varhidden and $\pred\varhidden$. 
    That is, we can go from \cref{app:prop:proof:eq-augment-zb} to \cref{app:eq:relating-both-interventions} directly by using only \varblock and the equal-marginals assumption:
    \begin{align}
        \pred p (\giventhat{\varoutcome}{\doop(\vartreat), \varcov}) 
        &= \int \pred p(\giventhat{\varoutcome}{\doop(\vartreat), \varblock, \varcov}) \pred p(\giventhat{\varblock}{\varcov}) \dif\varblock \\
        &= \int \pred p(\giventhat{\varoutcome}{\vartreat, \varblock, \varcov}) \pred p(\giventhat{\varblock}{\varcov}) \dif\varblock \\
        &= \int p(\giventhat{\varoutcome}{\vartreat, \varblock, \varcov}) p(\giventhat{\varblock}{\varcov}) \dif\varblock \\
        &= p(\giventhat{\varoutcome}{\doop(\vartreat), \varcov}) \,.
    \end{align}
\end{proof}

    Even though we can leverage and simplify \cref{app:prop:causal-query-identifiability} as shown above, it is worth remarking that, for this particular case, the model identifiability results described in \cref{app:subsec:uncounded-identifiability} are stronger, as it provides results on the identifiability of the causal generators and exogenous distributions, and therefore of any causal query derived from them.

\subsubsection{Confounded outcome case}

For the case where only the outcome variable is a child of the hidden confounder, we can apply a similar reasoning as we did in the previous case, although this time we cannot leverage the stronger results from \citet{javaloy2024causal}. 
More specifically: 

\begin{corollary}%
    Given two SCMs $\scm \coloneqq (\funcb, \distribution[\varexo], \distribution[\varhidden])$ and $\pred\scm \coloneqq (\pred\funcb, \distribution[\pred\varexo], \distribution[\pred\varhidden])$, assume that they are Markov-equivalent---\ie, they induce the same causal graph---and coincide in their marginal distributions, $p(\varobs) \aeeq \pred p(\varobs)$.
    Assume that $\vartreat \notin \varchildren{\varhidden}$\,. 
    Then, $p(\giventhat{\varoutcome}{\doop(\vartreat), \varcov}) = \pred p(\giventhat{\varoutcome}{\doop(\vartreat), \varcov})$, where $\varoutcome, \vartreat, \varcov \subset \varobs$\,.
\end{corollary}
\begin{proof}
    The proof is identical to that of \cref{app:cor:unconfounded-case}.
\end{proof}

    \paragraph{Front-door example.}
    
    \begin{wrapfigure}[7]{r}{.25\linewidth}
        \centering
        \vspace{-\baselineskip}
        \includestandalone[width=0.8\linewidth]{figs/graphs/front-door}
        \caption{Example of a front-door causal.}
        \label{app:fig:frontdoor}
    \end{wrapfigure}
    While the proof above is trivial given the previous results, it is worth stressing the importance of modeling the hidden confounder as we do in this work with \ours. %
    As an example, consider the SCM depicted in \cref{app:fig:frontdoor}, where we have that the outcome is directly confounded by \varhidden, while \vartreat is not.
    In this case, \ours can correctly estimate the causal effects of \varblock and \vartreat on \varoutcome, \ie, to correctly estimate $p(\giventhat{\varoutcome}{\doop(\vartreat)})$ and $p(\giventhat{\varoutcome}{\doop(\varblock)})$, using $\pred\varhidden$ to model the influence of \varblock onto \varoutcome that is not explained through \vartreat.
    Other models that do not model \varhidden---\eg, an unaware CNF~\citep{javaloy2024causal}---would be able to match the observed marginal distribution (as they are universal density approximators) and therefore to estimate $p(\giventhat{\varoutcome}{\doop(\varblock)})$ (as it is identifiable through the mediator \vartreat using the front-door criterion), yet they would necessarily fail to estimate $p(\giventhat{\varoutcome}{\doop(\vartreat)})$, since they assume that $\giventhat{\varoutcome \indep \varblock}{\vartreat}$ yet we know that $\giventhat{\varoutcome \notindep \varblock}{\vartreat}$ in the true model.
    In other words, an unaware CNF would hold that $p(\giventhat{\varoutcome}{\doop(\vartreat)}) = p(\giventhat{\varoutcome}{\vartreat})$ which is clearly false by looking at \cref{app:fig:frontdoor}.

    To be even more explicit, in this case we would have a data-generating process that factorizes as
    \begin{equation}
        \pred p(\varblock, \vartreat, \varoutcome, \pred\varhidden) = \pred p(\pred\varhidden) \pred p(\giventhat{\varblock}{\pred\varhidden}) \pred p(\giventhat{\vartreat}{\varblock}) \pred p(\giventhat{\varoutcome}{\vartreat, \pred\varhidden}) \,,
        \label{app:eq:graph_fact}
    \end{equation}
    and hence the estimated interventional distribution from \ours matches the true one: %
        \begin{align}
        p(\giventhat{\varoutcome}{\doop(\vartreat)}) 
        &= \int p(\giventhat{\varoutcome}{\vartreat, \varblock}) p(\varblock) \dif \varblock && \eqcomment{\varblock forms a valid adjustment set} \\
        &= \int \left \{ \int \pred p(\giventhat{\varoutcome}{\vartreat, \varblock, \pred\varhidden}) \pred p(\giventhat{\pred\varhidden}{\vartreat, \varblock}) \dif \pred\varhidden \right \} \pred p(\varblock)\dif\varblock && \eqcomment{Factorization and eq. marginals} \\
        &= \iint \pred p(\giventhat{\varoutcome}{\vartreat, \pred\varhidden})\pred p(\giventhat{\pred\varhidden}{\varblock}) \pred p(\varblock) \dif\varblock \dif\pred\varhidden  && \eqcomment{Factorization in \cref{app:eq:graph_fact}} \\
        &= \int \pred p(\giventhat{\varoutcome}{\vartreat, \pred\varhidden}) \pred p(\pred\varhidden)  \dif\pred\varhidden && \eqcomment{marginalize \varblock} \label{eq:marginalize_b} \\
        &= \pred p(\giventhat{\varoutcome}{ \doop(\vartreat)}) \,.  && 
        \label{eq:frontdoor_step_1}
    \end{align}

\subsubsection{Hidden-confounded treatment case}
\label{app:subsec:confounded-treatment}

\begin{wrapfigure}[7]{r}{.25\linewidth}
    \centering
    \vspace{-\baselineskip}
    \includestandalone[width=.5\linewidth]{figs/graphs/no-adjustment}
    \caption{Case with no valid adjustment set.}
    \label{app:fig:no-adjustment}
\end{wrapfigure}
When only the treatment variable \vartreat is a child of \varhidden, we can face two different scenarios: \itemi we find a valid adjustment set \varblock blocking all backdoor paths, in which case we can reason just as in the other partially hidden-confounded case, and \itemii we cannot, and then rely on do-calculus and the identifiability \wrt \varblock.
For example, if \varblock happens to be a parent of \varoutcome which is directly caused by the treatment variable \vartreat and the hidden confounder \varhidden as in \cref{app:fig:no-adjustment}, we cannot find a valid adjustment set for the causal query, but it may still serve us if we can identify the same query with the adjustment set as outcome variable.

\begin{corollary}%
    Given two SCMs $\scm \coloneqq (\funcb, \distribution[\varexo], \distribution[\varhidden])$ and $\pred\scm \coloneqq (\pred\funcb, \distribution[\pred\varexo], \distribution[\pred\varhidden])$\,, assume that they are Markov-equivalent---\ie, they induce the same causal graph---and coincide in their marginal distributions, $p(\varobs) \aeeq \pred p(\varobs)$.
    If $\varoutcome \notin \varchildren{\varhidden}$\, then, $p(\giventhat{\varoutcome}{\doop(\vartreat), \varcov}) = \pred p(\giventhat{\varoutcome}{\doop(\vartreat), \varcov})$, where $\varoutcome, \vartreat, \varcov \subset \varobs$\, if there exists $\varblock \subset \varobs$ not containing variables from the previous subsets, such that one of the following two conditions are true:
    \begin{enumerate}[i)]
        \item \varblock forms a valid adjustment set for the query $p(\giventhat{\varoutcome}{\doop(\vartreat), \varcov})$. %
        
        \item \varblock blocks all backdoor paths and the query $p(\giventhat{\varblock}{\doop(\vartreat), \varcov})$ is identifiable. %
    \end{enumerate}
\end{corollary}
\begin{proof}
    If condition \itemi holds, then we have a valid adjustment set, and the proof is identical to that of \cref{app:cor:unconfounded-case}.

    Otherwise, if condition \itemii holds, we have that the interventional query on \varoutcome equals the observational query when conditioned on \varblock, but that now \varblock is not independent of $\doop(\vartreat)$, \ie,

    \begin{align}
        \pred p (\giventhat{\varoutcome}{\doop(\vartreat), \varcov}) 
        &= \int \pred p(\giventhat{\varoutcome}{\doop(\vartreat), \varblock, \varcov}) \pred p(\giventhat{\varblock}{\doop(\vartreat), \varcov}) \dif\varblock \\
        &= \int \pred p(\giventhat{\varoutcome}{\vartreat, \varblock, \varcov}) \pred p(\giventhat{\varblock}{\doop(\vartreat), \varcov}) \dif\varblock \\
        &= \int p(\giventhat{\varoutcome}{\vartreat, \varblock, \varcov}) p(\giventhat{\varblock}{\doop(\vartreat), \varcov}) \dif\varblock \\
        &= p(\giventhat{\varoutcome}{\doop(\vartreat), \varcov}) \,,
    \end{align}
    where we needed to use that the query $p(\giventhat{\varblock}{\doop(\vartreat), \varcov})$ is identifiable in the third equality.
\end{proof}

\subsubsection{Napkin example}

\begin{wrapfigure}[9]{r}{0.35\linewidth}
    \centering
    \vspace{-\baselineskip}
        \begin{tikzcd}[
        cells={nodes={main node}}, column sep=small, row sep=small
    ]
        &   |[hidden]|\varhidden_1 \arrow[hidden, ddl] \arrow[hidden, ddrr] & &  \\
        & |[hidden]|\varhidden_2 \arrow[hidden, dl] \arrow[hidden, dr] & \\
        \varproxyt \arrow[r] & \varblock \arrow[r] & \vartreat \arrow[r]  & \varoutcome \\
    \end{tikzcd}
    \caption{Napkin causal graph~\citep{pearl2018book}.}
    \label{fig:napkin_graph}
\end{wrapfigure}
Finally, we want to show one last illustrative example where \ours provides correct estimates of a causal query that is identifiable by the do-calculus, but neither the backdoor nor the front-door criteria are applicable.
While redundant (as the query is identifiable in the classical sense, and then \cref{cor:do-calculus} applies), we believe it can be a good exercise to convince the reader.
Namely, the graph of \cref{fig:napkin_graph} appears as the napkin graph in \citet[Fig. 7.5]{pearl2018book}. What is particularly interesting in this graph is that \varproxyt is not a valid adjustment set since, despite blocking the backdoor path from \vartreat to \varoutcome through \varblock, it forms a collider of $\varhidden_1$ and $\varhidden_2$.

However, $\varhidden_1$ only affects the outcome and $\varhidden_2$ only affects the treatment. Following from our previous results, the causal effect from \vartreat to \varoutcome should be correctly estimated by \ours.
Here, we show that this is the case. First, let us express the causal query of interest in another form applying do-calculus:
\begin{align}
    p (\giventhat{\varoutcome}{\doop(\vartreat)}) &= p (\giventhat{\varoutcome}{\doop(\varoutcome), \doop(\vartreat)}) =  && \eqcomment{Rule 3 of do-calculus since $\varoutcome \indep_{\graph_{\bar{\vartreat}, \bar{\varblock}}} \giventhat{\varblock}{\vartreat}$}
    \\
    & = p (\giventhat{\varoutcome}{\vartreat, \doop(\varblock)}) = && \eqcomment{Rule 2 of do-calculus $\varoutcome \indep_{\graph_{\bar{\varblock}, \underline{\vartreat}}} \giventhat{\vartreat}{\varblock}$}
    \\
    & = \frac{ p (\giventhat{\varoutcome, \vartreat}{\doop(\varblock)})}{p (\giventhat{\vartreat}{\doop(\varblock)})} && \eqcomment{Conditional probability}
    \label{eq:napkin_frac}
\end{align}

Once we have this expression, let us work on the numerator, considering that \ours is Markov-equivalent with the graph in \cref{fig:napkin_graph}:

{%
\begin{align}
    &p(\giventhat{\varoutcome, \vartreat}{\doop(\varblock)})  = \int p(\giventhat{\varoutcome, \vartreat}{\varblock, \varproxyt}) p(\varproxyt) \dif \varproxyt && \eqcomment{Backdoor criterion} &\\
     & =  \iiint \pred p(\giventhat{\varoutcome, \vartreat, \pred \varhidden_1, \pred \varhidden_2}{\varblock, \varproxyt}) p(\varproxyt) \dif \varproxyt \dif \pred \varhidden_1 \dif \pred \varhidden_2 && \eqcomment{Eq. marginals} &\\
     & = \iiint \pred p (\giventhat[|]{\varoutcome}{\vartreat, \pred \varhidden_1, \pred \varhidden_2, \varblock, \varproxyt}) \pred p (\giventhat[|]{\vartreat}{\pred \varhidden_1, \pred \varhidden_2, \varblock, \varproxyt}) \pred p(\giventhat[|]{\pred \varhidden_1, \pred \varhidden_2}{\varproxyt}) p(\varproxyt) \dif \varproxyt \dif \pred \varhidden_1 \dif \pred \varhidden_2
    && \eqcomment{Factorization} &\\
     & =  \iiint \pred p (\giventhat{\varoutcome}{\vartreat, \pred \varhidden_2}) \pred p (\giventhat{\vartreat}{\pred \varhidden_2, \varblock}) \pred p(\giventhat{\pred \varhidden_1, \pred \varhidden_2}{\varproxyt}) p(\varproxyt) \dif \varproxyt \dif \pred \varhidden_1 \dif \pred \varhidden_2
    && \eqcomment{Do-calculus rule 1} & \\
    & = \int \int \pred p (\giventhat{\varoutcome}{\vartreat, \pred \varhidden_2}) \pred p (\giventhat{\vartreat}{\pred  \varhidden_2, \varblock}) \pred p(\pred \varhidden_1, \pred \varhidden_2) \dif \pred \varhidden_1 \dif \pred \varhidden_2 && \eqcomment{Marginalize \varproxyt}  & \\
     & = \int \int \pred p (\giventhat{\varoutcome}{\vartreat, \pred \varhidden_2}) \pred p (\giventhat{\vartreat}{\pred \varhidden_2, \varblock}) \pred p(\pred \varhidden_1) \pred p (\pred \varhidden_2) \dif \pred \varhidden_1 \dif \pred \varhidden_2 && \eqcomment{$\pred \varhidden_1 \indep_\graph \varhidden_2$} &\\
     & = 
     \int \pred p (\giventhat{\varoutcome}{\vartreat, \pred \varhidden_2})\pred p(\pred \varhidden_1)  \dif \pred \varhidden_1
     \int \pred p (\pred \varhidden_2) \pred p (\giventhat{\vartreat}{\pred \varhidden_2, \varblock}) \dif \pred \varhidden_2 &&  \eqcomment{Separate integrals} \\
      & = \pred p(\giventhat{\varoutcome}{\doop(\vartreat)}) \; \pred p(\giventhat{\vartreat}{\doop(\varblock)}) && \eqcomment{\ours estimate} &
     \label{eq:napkin_numerator}
\end{align}}

Note also that, as shown in \cref{eq:frontdoor_step_1}, \ours correctly estimates $p(\giventhat{\vartreat}{\doop(\varblock)})$.
Therefore, if we substitute \cref{eq:napkin_numerator} in \cref{eq:napkin_frac}, we have that
\begin{align}
    p (\giventhat{\varoutcome}{\doop(\vartreat)}) = \frac{\pred p(\giventhat{\varoutcome}{\doop(\vartreat)}) \; p(\giventhat{\vartreat}{\doop(\varblock)})}{p (\giventhat{\vartreat}{\doop(\varblock)})} = \pred p(\giventhat{\varoutcome}{\doop(\vartreat)}) \equationPunctuation{.}
\end{align}

That is, we have explicitly shown that \ours correctly estimates the true causal query $p(\giventhat{\varoutcome}{\doop(\vartreat)})$.

\subsection{Counterfactual query identifiability}
\label{app:subsec:cf-identifiablity}

    In this section, we show that counterfactual query identifiability is a direct result of the interventional query identifiability from the previous section.

    In order to formally define counterfactuals, in this section we introduce the concept of counterfactual SCMs in a rather untraditional fashion. 
    Namely, we combine the concepts of twin networks from \citet{pearl2009causality} (which replicates the data-generating process) and that of counterfactual SCMs from \citet{peters2017elements} (which defines a counterfactual \textit{prior} to the intervention) as follows:

    \begin{definition}[Counterfactual twin SCM] \label{def:twin-scm}
        Given a SCM $\scm = (\funcb, \distribution[\varexo], \distribution[\varhidden])$, we define its counterfactual twin SCM as a SCM $\cfactual\scm$ where all structural equations are duplicated, and the exogenous noise is shared across replications, and where additionally one of the halves is observed (``the factual world''), and the other half is unobserved (``the counterfactual world''). 
    \end{definition}

    We provide in \cref{app:fig:twin-construction} a more intuitive depiction on the construction of these counterfactual twin networks. 
    From this definition, one can recover the counterfactual SCM defined by \citet{peters2017elements} by just focusing on the replicated part of the counterfactual twin network, and conditioning the exogenous noise and hidden confounder on the observed half, \ie, $(\funcb, \distribution[\giventhat{\varexo}{\factual\varobs}], \distribution[\giventhat{\varhidden}{\factual\varobs}])$\,. 
    Similarly, one can compute the usual counterfactual query by performing an intervention on the counterfactual twin network, \ie, by replacing the intervened equations by the constant intervened value, and computing the query conditioned on the factual variables, $p(\giventhat{\cfactual\varoutcome}{\doop(\cfactual\vartreat), \factual\varobs})$. 
    This is visually represented in \cref{app:subfig:cf-twin}.

\begin{figure}[t]
    \centering
    \hfill
    \begin{subfigure}[b]{0.3\linewidth}
        \centering  
            \begin{tikzcd}[
        cells={nodes={main node}}, column sep=tiny, row sep=small
        ]
        \varproxyo \arrow[dr] & & |[hidden]|\varhidden \arrow[hidden, ll] \arrow[hidden, rr] \arrow[hidden, dl] \arrow[hidden, dr] & & \varproxyt \arrow[dl] \\
        & \vartreat \arrow[rr] & & \varoutcome
    \end{tikzcd}
        \caption{}
    \end{subfigure}
    \hfill
    \begin{subfigure}[b]{0.3\linewidth}
        \centering
            \begin{tikzcd}[
        cells={nodes={main node}}, column sep=tiny, row sep=small
        ]
        |[dashed]| \evarexo_\varproxyo \arrow[d, dashed] & |[dashed]| \evarexo_\vartreat \arrow[dd, dashed] & |[hidden]|\varhidden \arrow[hidden, dll] \arrow[hidden, drr] \arrow[hidden, ddl] \arrow[hidden, ddr] & |[dashed]| \evarexo_\varoutcome \arrow[dd, dashed] & |[dashed]| \evarexo_\varproxyt \arrow[d, dashed] \\
        \factual\varproxyo \arrow[dr] & &  & & \factual\varproxyt \arrow[dl] \\
        & \factual\vartreat \arrow[rr] & & \factual\varoutcome
    \end{tikzcd}
        \caption{}
    \end{subfigure}
    \hfill
    \begin{subfigure}[b]{0.3\linewidth}
        \centering
            \begin{tikzcd}[
        cells={nodes={main node}}, column sep=tiny, row sep=small
        ]
        & \cfactual\vartreat \arrow[rr] & & \cfactual\varoutcome \\
        \cfactual\varproxyo \arrow[ur, secondcolor, "\tiny \text{\faCut}"{marking, rotate=90,xshift=-0.3mm, pos=0.6}] & &  & & \cfactual\varproxyt \arrow[ul] \\
        |[dashed]| \evarexo_\varproxyo \arrow[d, dashed] \arrow[u, dashed] & |[dashed]| \evarexo_\vartreat \arrow[dd, dashed] \arrow[uu, dashed, secondcolor, "\tiny \text{\faCut}"{marking, yshift=-0.3mm, rotate=90, pos=0.8}] & |[hidden]|\varhidden \arrow[hidden, dll] \arrow[hidden, drr] \arrow[hidden, ddl] \arrow[hidden, ddr] \arrow[hidden, ull] \arrow[hidden, urr] \arrow[hidden, uul, secondcolor, "\tiny \text{\faCut}"{marking, yshift=0.3mm, rotate=270, pos=0.8}] \arrow[hidden, uur] & |[dashed]| \evarexo_\varoutcome \arrow[dd, dashed] \arrow[uu, dashed] & |[dashed]| \evarexo_\varproxyt \arrow[d, dashed] \arrow[u, dashed] \\
        |[observed]| \factual\varproxyo \arrow[dr] & &  & & |[observed]| \factual\varproxyt \arrow[dl] \\
        & |[observed]| \factual\vartreat \arrow[rr] & & |[observed]| \factual\varoutcome
    \end{tikzcd}
        \caption{}
        \label{app:subfig:cf-twin}
    \end{subfigure}
    \hfill
    \caption{Example of the transition from \captiona the regular depiction of a (confounded) SCM, to \captionb an explicit SCM where the exogenous variables are drawn, and \captionc a counterfactual twin SCM where the data-generating process is replicated in the ``factual and counterfactual worlds''. Figure \captionc also depicts which nodes are observed and which are severed in order to compute a counterfactual query of the type $p(\giventhat{\cfactual\varoutcome}{\doop(\cfactual\vartreat), \factual\varobs})$\,.}
    \label{app:fig:twin-construction}
\end{figure}

    In order to prove query identifiability in the counterfactual setting, we need to use the following technical result regarding the completeness of a random variable:

    \begin{lemma} \label{lemma:completeness}
        If a random variable \varhidden is complete given \varproxyo for almost all \varblock, as given by \cref{def:completeness}, then it is complete given \varproxyo for almost all \varblock and \varcov, where \varcov is another continuous random variable.
    \end{lemma}
    \begin{proof}
        We prove this result by contradiction.
        Assume that the result does not hold, then there must exist a non-zero measure subset of the space of $\varblock \times \varcov$ for which there exists a square-integrable function $g(\cdot)$ such that $\int g(\varhidden, \varblock, \varcov) p(\giventhat{\varhidden}{\varblock, \varcov, \varproxyo}) \dif\varhidden = 0$
        for almost all \varproxyo, but $g(\varhidden, \varblock, \varcov) \neq 0$ for almost all \varhidden.

        Since this subset has positive measure, there must contain an $\epsilon$-ball within. If we now focus on the \varblock-projection of this ball where we fix \varcov to its value on the center, we have that it is a subset of non-zero measure in the space of \varblock (as otherwise it would be zero-measure in the Cartesian-product measure), where the function $g(\cdot, \varcov)$ breaks our initial assumption of the completeness of \varhidden. Thus, we reach a contradiction.
    \end{proof}

    Given \cref{def:twin-scm}, it is rather intuitive that, if a causal query is identifiable in a SCM $\scm$, then it has to be identifiable in both halves of its induced counterfactual twin SCM $\cfactual\scm$, as they are identical.
    More importantly, we can now leverage again \cref{app:prop:causal-query-identifiability}, this time with $\varcov = \factual\varobs$, to prove counterfactual query identifiability whenever we have interventional query identifiability.

    \begin{proposition}[Counterfactual identifiability] \label{app:prop:cf-identifiability}
        Given two SCMs $\scm \coloneqq (\funcb, \distribution[\varexo], \distribution[\varhidden])$ and $\pred\scm \coloneqq (\pred\funcb, \distribution[\pred\varexo], \distribution[\pred\varhidden])$, assume that they are Markov-equivalent---\ie, they induce the same causal graph---and that they coincide in their marginal distributions, $p(\varobs) \aeeq \pred p(\varobs)$.
        Then, if a query $p(\giventhat{\varoutcome}{\doop(\vartreat)})$ is identifiable in the sense of \cref{app:prop:causal-query-identifiability}, where $\varoutcome, \vartreat \subset \varobs$, the query $p(\giventhat{\cfactual\varoutcome}{\doop(\cfactual\vartreat), \factual\varobs})$ is also identifiable in the induced counterfactual twin SCM as long as the regularity conditions still hold, \ie, if:
        \begin{enumerate}[i)]
        \item $\iint \pred p(\giventhat{\pred\varhidden}{\varproxyt, \varblock, \varcov}) \pred p(\giventhat{\varproxyt}{\pred\varhidden, \varblock, \varcov}) \dif\pred\varhidden\dif\varproxyt < \infty$ for almost all \varblock, \varcov, and
        \item $\int \pred p(\giventhat{\varoutcome}{\vartreat, \varblock, \pred\varhidden, \varcov})^2 \pred p(\giventhat{\pred\varhidden}{\varblock, \varcov}) \dif \pred\varhidden < \infty$ for almost all \vartreat, \varblock, and \varcov.
    \end{enumerate}
    \end{proposition}
    \begin{proof}
        We essentially need to prove that the independence and completeness assumptions keep holding when we add the factual covariate, $\varcov = \factual\varobs$\,.
        
        For the independence, we need to show that, if we have a set of variables that fulfill the independence conditions from \cref{app:prop:causal-query-identifiability}, then this set of variables keeps holding them if we include $\varcov = \factual\varobs$\,. 
        This is, however, easy to show since factual and counterfactual variables only have ``tail-to-tail'' dependencies, \ie, they are connected only through the shared exogenous variables. As a result, if two variables from the same half are conditionally independent given a third set of variables, conditioning on the other half cannot change this independence.

        For the completeness, we need to show that introducing the factual variable retains the completeness assumed in \cref{app:prop:causal-query-identifiability},  which is direct to show using \cref{lemma:completeness}. Specifically, it holds that
        \begin{enumerate}[i)]
            \item \varhidden is complete given \varproxyo for almost all \vartreat, \varblock, and \varcov, and
            \item \pred\varhidden is complete given \varproxyt for almost all \varblock and \varcov.
        \end{enumerate}

        Therefore, the requirements of \cref{app:prop:causal-query-identifiability} hold when we append a factual variable to the twin network, and thus we can reapply all the results from the previous sections to the counterfactual cases.
    \end{proof}

    Once proven the result above, proving \cref{cor:decaf-cf-estimates} is direct by following the exact same steps as we did in \cref{app:subsec:general-case-proof} to the counterfactual twin network instead of the original network.

    It is important to note that, while the results above provide counterfactual identifiability whenever we have interventional identifiability, we still rely on how much of a good approximation the encoder is to the inverse of the decoder in the proposed \ours model. 
    That is, the quality of the encoder determines how well we can perform the abduction step to compute counterfactuals.
    This consideration is unique to counterfactuals, as we just have to sample from the prior of \varhidden in the case of interventional queries.

    \section{Experimental details and additional results}
\label{app:sec:additional-results}

This section presents a series of ablation studies designed to answer practical questions about the behavior of \ours and to justify key design choices.
These analyses provide empirical guidance for practitioners, clarifying how model performance depends on factors such as training data size, latent dimensionality, and proxy quality.
Beyond validating theoretical claims, the results offer concrete recommendations for effectively applying \ours in realistic scenarios.

Finally, we include complementary experimental details and extended comparisons with baseline methods, covering dataset descriptions, data-generating processes, and additional quantitative results and visualizations that extend those presented in \cref{sec:experiments}.

\subsection{Ablation study on latent dimension and number of proxies}
\label{app:sec:additional-results-ablation}

\begin{wrapfigure}{r}{0.5\linewidth}
        \centering
    \includegraphics[width=\linewidth]{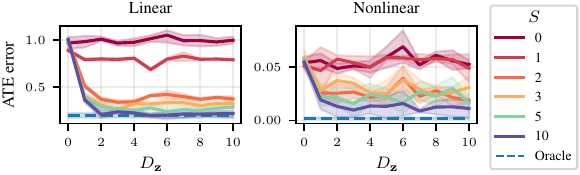}
    \caption{ATE absolute error as we change the number of proxy variables, \proxysize, and the latent dimensionality, \hiddensize. We plot mean and \SI{95}{\percent} CI over 5 realizations, intervening on the \nth{25}, \nth{50}, and \nth{75} percentile value of \vartreat. Oracle represents a causal normalizing flow that observes \varhidden.}
    \label{fig:ate_ablation}
\end{wrapfigure}

We include here additional results of the ATE error, complementary to those of \cref{subsec:ablation}. If we observe \cref{fig:ate_ablation}, we extract the
same conclusion as observing counterfactual error,
the causal effect is not recoverable with less than two
proxies, and more proxies result in better estimates.
On the other hand, the selection of the dimension of
the latent space bigger than the true dimension of the
latent confounders does not affect the performance
negatively.

Overall, these findings indicate that \ours is robust to latent space over-specification, thanks to KL regularization, and that, in practice, providing more and better proxies leads to more accurate estimation of causal effects even when confounding structure is unknown.

\paragraph{Details of the generative process.}
We show the equations that we have used for the ablation study. There exist two unobserved confounders, $\varhidden_1$ and $\varhidden_2$. Note that the proxies available in the nonlinear experiment are bounded or periodic, especially sigmoids and hyperbolic tangents saturate and $\max(0, \ervx)$ loses all the information about the confounder for negative values and sines and cosines are periodic functions. In other words, the distributions $p(\varhidden \mid \varproxyo_i)$ are not complete, we lose information about \varhidden when in the transformations to each $\varproxyo$. However, if we add more proxies of the confounders, the information that the proxies contain about the confounder is higher, and the causal effect of $\ervx_1$ on $\ervx_2$ becomes recoverable.

\begin{equation*}
\small
\begin{array}{c@{\hspace{.25cm}}c}
    \text{Linear} & \text{Nonlinear} \\
    \left\{
    \begin{aligned}
        &\varhidden_1 \sim P_{\varhidden_1} \\
        &\varhidden_2 \sim P_{\varhidden_2}\\
        &\vartreat= 1.5 \cdot \varhidden_1 + 0.5\cdot\varhidden_2 + 0.4 \cdot \ervu_\vartreat \\
        &\varoutcome = -0.75 \cdot \varhidden_1 + 0.6\cdot\varhidden_2+ 0.9 \cdot \vartreat + 0.3 \cdot \ervu_\varoutcome \\
        &\varproxyo_1 = -0.5 \cdot \varhidden_1 + 0.3\cdot\varhidden_2+ 0.5 \cdot \ervu_2 \\
        &\varproxyo_2 = 0.75 \cdot \varhidden_1 - 0.4\cdot\varhidden_2+ 0.4 \cdot \ervu_2 \\
        &\varproxyo_3 = -0.85 \cdot \varhidden_1 + 0.6\cdot\varhidden_2+ 0.6 \cdot \ervu_3 \\
        &\varproxyo_4 = 0.6 \cdot \varhidden_1 + 0.6\cdot\varhidden_2 + 0.55 \cdot \ervu_4 \\
        &\varproxyo_5 = -0.8 \cdot \varhidden_1 + 0.4\cdot\varhidden_2+ 0.4 \cdot \ervu_5 \\
        &\varproxyo_6 = 0.9 \cdot \varhidden_1 - 0.7\cdot\varhidden_2+ 0.6 \cdot \ervu_6 \\
        &\varproxyo_7 = -0.72 \cdot \varhidden_1 + 0.5\cdot\varhidden_2 + 0.56 \cdot \ervu_{8} \\
        &\varproxyo_{8} = 0.78 \cdot \varhidden_1 + 0.4\cdot\varhidden_2+ 0.58 \cdot \ervu_{8} \\
        &\varproxyo_{9} = -0.55 \cdot \varhidden_1 + 0.7\cdot\varhidden_2+ 0.6 \cdot \ervu_{9} \\
        &\varproxyo_{10} = 0.88 \cdot \varhidden_1 + 0.3\cdot\varhidden_2+ 0.4 \cdot \ervu_{10}
    \end{aligned}
    \right.
    &
    \left\{
\begin{aligned}
    &\varhidden_1 \sim P_{\varhidden_1} \\
    &\varhidden_2 \sim P_{\varhidden_2} \\
    &\vartreat = \frac{\varhidden_1^2}{4} \cdot \sin\left(\frac{\varhidden_2}{2}\right) + \varhidden_1 + 0.6 \cdot \ervu_\vartreat \\
    &\varoutcome = \frac{\varhidden_1 \cdot \vartreat}{4} + 0.8 \cdot \varhidden_2 + 0.5 \cdot \vartreat + \ervx_1 \cdot \ervu_2 \cdot 0.3 + 0.2 \cdot \ervu_\varoutcome \\
    &\varproxyo_1 = 0.6 \cdot \varhidden_1^2 + \left(\frac{\varhidden_2}{4}\right)^3 + 0.3 \cdot \sin\left(\frac{\varhidden_2}{2}\right) + 0.5 \cdot \ervu_1 \\
    &\varproxyo_2 = \sin\left(\frac{\varhidden_1}{2}\right) + \cos\left(\frac{\varhidden_2}{3}\right) + 0.4 \cdot \ervu_2 \\
    &\varproxyo_3 = \cos\left(\frac{\varhidden_1}{2}\right) - \text{tanh}\left(\frac{\varhidden_2}{3}\right) + 0.6 \cdot \ervu_3 \\
    &\varproxyo_4 = \text{tanh}\left(\frac{\varhidden_1}{2}\right) + \sigma\left(\frac{\varhidden_2}{2}\right) + 0.55 \cdot \ervu_4 \\
    &\varproxyo_5 = \sigma\left(\frac{\varhidden_1}{2}\right) + \max(0, -\varhidden_2) + 0.4 \cdot \ervu_5 \\
    &\varproxyo_6 = \max(0, \varhidden_1) - 0.5 \cdot \max(0, \varhidden_2) + 0.6 \cdot \ervu_6 \\
    &\varproxyo_7 = \max(0, -\varhidden_1) + 0.3 \cdot \max(0, -\varhidden_2) + 0.5 \cdot \varhidden_1 \cdot \ervu_7 \\
    &\varproxyo_{8} = 0.8 \cdot \max(0, \varhidden_1) + 0.3 \cdot \max(0, \varhidden_2) + 0.58 \cdot \ervu_{8} \\
    &\varproxyo_{9} = 0.75 \cdot \max(0, -\varhidden_1) + 0.5 \cdot \max(0, \varhidden_2) + 0.6 \cdot \ervu_{9} \\
    &\varproxyo_{10} = 0.3 \cdot \varhidden_1^3 + 0.5 \cdot |\varhidden_2| + 0.4 \cdot \ervu_{10}
\end{aligned}
    \right.
\end{array}
\end{equation*}

\subsection{Ablation study for encoder selection} \label{app:sec:encoder_ablation}
\begin{wrapfigure}[16]{R}{0.5\linewidth}
    \centering
    \vspace{-\baselineskip}
    \includegraphics[width=\linewidth]{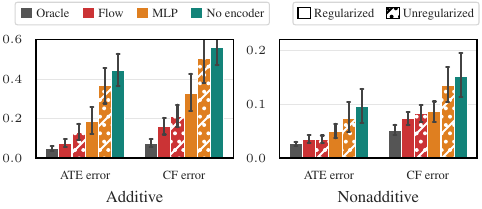}
    \caption{Ablation for encoder selection in Sachs' dataset. Metrics and $95\%$ CI over 5 realization and all confounded identifiable effects, intervening on percentiles 25, 50 and 75 of each intervened variable. Oracle represents a causal normalizing flow that observes all the confounders.}
\label{fig:bar_ablation_encoder}
\end{wrapfigure} 
We have performed an ablation study for selecting the encoder in the Sachs' dataset, where we evaluate the errors in the estimations of causal queries using a conditional normalizing flow (Flow) and a multilayer perceptron (MLP) as encoders. We also evaluate the impact of using the warm-up regularization \citep{vahdat2020nvae} in the KL term. We can observe in \cref{fig:bar_ablation_encoder} that we achieve lower errors when applying a regularized flow. This is able to model dependent latent variables and provides a more flexible representation. In addition, we can appreciate that applying the warm-up regularization term is useful and does not produce negative effects.

The improvement achieved by the flow is explained by the following practical aspects of the conditional normalizing flows. First, we can efficiently introduce the factorization proposed in \cref{eq:factorization-encoder}, taking advantage of the structure of the causal graph (see \cref{fig:scheme_decaf_napkin} for an example), while this factorization implies the use of several MLP. Second, normalizing flows are universal density approximators and do not need to assume specific posterior distributions (\ie Gaussians). Note that every continuous distribution can be modeled by a conditional normalizing flow, following the Kn\"othe-Rosenblatt transport \citep{Knothe1957ContributionsTT, 10.2307/2236692}.

\subsection{Ablation on encoder factorization}
\label{app:sec:encoder_factorization}

Using a conditional normalizing flow as the encoder allows us to model the dependencies between the observations and the posterior of the latent variables as desired.

\begin{wrapfigure}[13]{R}{0.5\linewidth}
    \centering
    \vspace{-\baselineskip}
    \includegraphics[width=\linewidth]{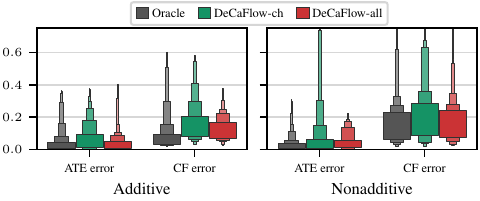}
    \caption{Ablation for posterior factorization in Ecoli70 dataset. Boxenplots of error metrics in the \textcolor{idcolor}{identifiable} edges of \cref{fig:ecoli_graph_custom}. \ours-ch uses \cref{eq:factorization-encoder-ch} and \ours-all uses \cref{eq:factorization-encoder-expanded} for posterior factorization.}
    \label{app:fig:ablation_factorization}
\end{wrapfigure}

We propose in \cref{eq:factorization-encoder} (extended in \cref{eq:factorization-encoder-expanded}) a factorization in which each hidden confounder depends on its parents (other hidden confounders), its children and the parents of its children, avoiding cycles. If a child of an unobserved confounder, $c$, has other parents, then that child is a collider between the hidden confounders and the other parents of $c$. Therefore, conditioned on $c$, the hidden confounder is dependent of the other parents of $c$, given $c$. That is the reason because we consider sensible to include the other parents of $c$ in the factorization of the hidden confounder, \varhidden.

However, we also provide an ablation study on the Ecoli70 dataset, where we show that this factorization indeed helps to the estimation of causal queries. Note that in the Ecoli70 dataset, \methodname{lacY} is a collider between \methodname{eutG} and \methodname{cspG}. Therefore, conditioned on \methodname{lacY}, the two hidden confounders \methodname{eutG} and \methodname{cspG} become dependent. The factorization of \cref{eq:factorization-encoder-expanded} implies that the posterior of \methodname{cspG} is modeled employing all the children of \methodname{cspG} and also the parents of its children, with \methodname{eutG} among them. This dependency can be modeled by our encoder in an autoregressive manner.

This factorization incorporates more variables to approximate the posterior of the hidden confounders, compared with a simpler approach that consist in modeling only children dependencies:

\begin{equation}
    \small
    q_\phi (\giventhat{\varhidden}{\varobs}) = 
     \prod_{\indexthree=1}^\hiddensize q_\phi\left(\varhidden_\indexthree \mid
     \text{ch}(\varhidden_\indexthree)
     \right)
    \label{eq:factorization-encoder-ch}
\end{equation}

As shown in \cref{app:fig:ablation_factorization}, leveraging the factorization of \cref{eq:factorization-encoder-expanded} reduces the errors estimating causal queries in complex graphs, where colliders and dependent hidden confounders are present.

\subsection{Ablation on train size}
\label{app:sec:ablation_train_size}
We have proven theoretically that \ours is able to produce correct estimates of the identifiable causal queries, having that \ours achieves a perfect matching of the observational distribution, $p_\scm (\varobs)$. 

Although normalizing flows are universal density approximators \citep{papamakarios2021normalizing}, as a machine learning method, its performance improves as we increase the size of the dataset.

\begin{wrapfigure}{r}{0.5\linewidth}
    \centering
    \includegraphics[]{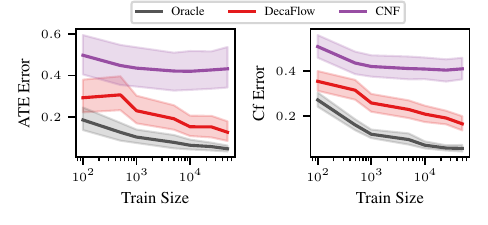}
    \caption{ATE and Counterfactual error in Sachs' additive dataset, varying the number of train samples. Test size are the same for all realizations. Metrics and 95$\%$ CI over 10 realizations and all confounded identifiable effects, intervening in percentiles 25 and 75 of each intervened variable. Oracle represents a CNF that observes the confounders. }
    \label{fig:ablation_train_size}
\end{wrapfigure}

Therefore, to further investigate the behavior of \ours under varying data availability, we conducted an ablation study on the \emph{training data size}. This analysis allows us to assess how the model's ability to estimate causal queries evolves as the number of observed samples increases. Since the objective of \ours is to recover the underlying causal mechanisms by matching the observational distributions, it is crucial to understand how data scarcity affects this matching process and, consequently, the accuracy of downstream tasks such as ATE and counterfactual estimation.

\cref{fig:ablation_train_size} reports both ATE and CF estimation errors as a function of the training set size. As expected, the errors systematically decrease when more data are available, since the model obtains a more accurate approximation of the data distribution. Notably, \ours exhibits a similar trend to the oracle, with both ATE and counterfactual errors monotonically decreasing as the number of training samples grows.

In contrast, the CNF (unaware of confounders) also benefits from larger datasets, but shows a slower improvement rate and an earlier plateau, since it does not have guarantees of correct causal estimation even if it matches the observational distribution. These results empirically validate our theoretical claims: as the training distribution approaches to the true observational distribution, the guarantees of \ours hold, leading to vanishing estimation errors. 

\subsection{Semi-synthetic Sachs' dataset}
\label{app:sec:additional-results-sachs}

This dataset represents a network of protein-signaling in human T lymphocytes. Every variable, except \methodname{PKA} and \methodname{Plcg} can be intervened upon; therefore, there is not only one causal query of interest, but tens of possible causal queries can arise in this setting. This highlights one of the strengths of \ours, because we only need a single trained model to answer all identifiable causal queries.

The original data contains a total of 853 observational samples; however, we have decided to evaluate our model on semi-synthetic data because of the following reasons:

\begin{itemize}
    \item The original network of \citet{sachs05dataset} contains cycles, which is a violation of one of our assumptions. However, we have found different versions of the causal graph \citep{pmlr-v202-kaltenpoth23a, luo11_bayesian_hierarchical} that do not contain cycles. Therefore, we have decided to employ the causal graph that appears in the library \textit{bnlearn} \citep{bnlearn}---a recognized library for Bayesian Network learning---as ground truth causal graph. The best way to ensure that the causal graph used is the ground truth is by generating samples according to the causal graph. In addition, that causal graph is the one used by \citet{chao2023interventional}.
    \item We can compare our model with one of the baseline models, DCM, with the same dataset as \citet{chao2023interventional} used.
    \item Semi-synthetic data allow us to compute all metrics to evaluate causal queries, having the ground truth.
    \item The interventions made in the real world dataset are \emph{soft interventions}, \ie, an external factor is used that modifies one of the variables, changing. On the other hand, \ours performs \emph{hard interventions}, making it unclear how to compare the two causal queries.
\end{itemize}

For generating the data in this experiment, we have followed the procedure proposed by \citet{chao2023interventional}, where they take the causal graph of \citet{sachs05dataset} and the empirical distribution of the root nodes, and generate the rest of the variables with random non-linear mechanisms. In addition, exogenous variables have been included in an additive and non-additive manner, respectively.

In the following, we complement the figures presented in \cref{sec:experiments} with a table that summarizes all the interesting metrics, evaluated on the \textcolor{idcolor}{confounded identifiable} causal queries shown in \cref{fig:sachs-graph}. Interventional distributions and counterfactuals have been computed intervening in percentiles 25, 50 and 75 of the intervened variable.

Since observational MMD is computed only once, the statistics given in \cref{tab:metrics_sachs_both} are calculated \textit{only} over 5 runs. On the other hand, we have as many interventional MMDs per run as interventions have been made. However, the statistics of interventional MMD are computed over all the interventions of all intervened variables and 5 runs (5 runs $\times$ $3$ intervened variables = 15 samples). Finally, statistics over counterfactual error and ate error aggregate all the intervention-outcome pairs over the five runs. For example, in this case we intervene in 3 variables, performing 3 different interventions and evaluate in 3, 2, and 1 variable, respectively, for each intervened variable, and we have a total of (3+2+1)$\times$3$\times$5 = 90 different measurements to compute the statistics.

\begin{table*}[t]
\caption{Performance metrics on Sachs datasets. Mean$_\text{std}$ over five runs and all causal queries of interest. Interventions on \methodname{Raf}, \methodname{Mek} and \methodname{Akt} and evaluating on \textcolor{idcolor}{confounded identifiable} effects. Bold indicates significantly better results ($95\%$ CI from a Mann-Whitney U test). Lower error values indicate better performance.}
\label{tab:metrics_sachs_both}
\scriptsize
\setlength{\tabcolsep}{2pt}
\centering
\begin{tabular}{clcccccccc}
\toprule
& & \multicolumn{4}{c}{Additive} & \multicolumn{4}{c}{Non-additive} \\
\cmidrule(lr){3-6} \cmidrule(lr){7-10}
& Model & MMD obs & MMD int & $|$ATE err$|$ & $|$CF err$|$ & MMD obs & MMD int & $|$ATE err$|$ & $|$CF err$|$ \\
& & $\times$$10^{4}$ & $\times$$10^{4}$ & $\times$$10^{2}$ & $\times$$10^{2}$ & $\times$$10^{4}$ & $\times$$10^{4}$ & $\times$$10^{2}$ & $\times$$10^{2}$ \\
\midrule
{ Oracle  } & CNF & $4.84_{1.84}$ & $7.50_{6.17}$& $6.05_{6.83}$ &  $10.03_{10.29}$ & $5.96_{2.37}$ & $6.71_{2.97}$& $2.34_{2.02}$ &  $4.84_{3.43}$ \\
\midrule
\multirow{2}{*}{Aware} 
& \ours & $\mathbf{2.15_{0.54}}$ & $\mathbf{7.04_{3.87}}$ &  $\mathbf{4.49_{6.76}}$ &  $\mathbf{12.95_{8.00}}$ & $5.12_{2.42}$ &$7.58_{16.92}$& $\mathbf{5.16_{5.61}}$ &  $1.83_{1.65}$ \\
& Deconfounder & -- & $-$& $34.34_{33.45}$ &  $71.13_{86.98}$ &
-- & $-$& $8.14_{10.69}$ &  $63.15_{79.12}$ \\
\midrule
\multirow{3}{*}{Unaware} 

& CNF & $5.80_{1.58}$ & $73.94_{88.78}$& $44.49_{39.12}$ &  $56.09_{38.89}$ &
$5.11_{1.90}$ &  $12.79_{20.73}$& $9.74_{15.71}$ &  $15.15_{15.37}$ \\
& ANM & $83.86_{13.41}$ &$110.28_{112.43}$& $22.42_{14.06}$ &  $29.40_{12.22}$ &
$81.90_{7.21}$ & $60.40_{144.08}$& $23.88_{13.94}$ &  $28.97_{12.44}$\\
& DCM & $87.80_{2.95}$ &$125.59_{118.20}$& $21.21_{11.34}$ &  $28.25_{6.96}$ &
$14.23_{4.57}$ &$69.74_{390.81}$& $8.44_{7.96}$ &  $27.50_{23.71}$ \\
\bottomrule
\end{tabular}
\end{table*}

The metrics in \cref{tab:metrics_sachs_both} indicate that \ours outperforms all baselines across all interventional and counterfactual causal queries in both settings of the semi-synthetic datasets. However, as discussed in \cref{sec:conclusion}, a limitation of our empirical approach is that the differences in observational MMD, the selection criterion for CGMs, are marginal between the \textit{oracle}, \ours, and CNF. Notably, \ours even achieves a lower MMD than the \textit{oracle}. This discrepancy arises because the number of variables is large, and the MMD differences are on the order of $10^{-4}$.

\subsection{Semi-synthetic Ecoli70 dataset}
\label{app:sec:additional-results-ecoli}

The Ecoli70 dataset represent the gene expression of 46 genes of the RNA sequence of the \textit{Escherichia coli} bacteria. The assumed causal graph comes from the study of \citep{schafer2005ecoli}, which provides insight into the regulatory mechanisms governing \textit{E. coli} gene expression. Examples of interventions in these networks are gene knockout and gene over-expression \citep{long2014metabolic}. A priori, there could be several variables in which intervening can be interesting in evaluating the effects in the cell. 

For this experiment, we have generated the data in the same way as done with Sachs' dataset with random mechanisms, but in this case, since we do not have enough samples, root nodes follow standard Gaussian distributions. We have included an additive and a non-additive ways of including exogenous variables. In this case, we have used a semi-synthetic dataset because the real dataset available in \textit{bnlearn} \citep{bnlearn} contains only 9 samples.

In \cref{fig:ecoli_graph_custom} is presented the causal graph of this setting.
In addition, note that \cref{fig:ecoli_graph_custom} has been extracted from our \cref{alg:causal_query} of causal effect identifiability. That is, we have specified the causal graph and the variables that are unmeasured, and our Algorithm returns (in green) all the paths that are identifiable by \ours. Consider that black arrows are also identifiable, not only by \ours, but also for any CGM that approximates the observed data. In red, arrows that are not identifiable by \ours because there are not enough proxies to infer an unbiased causal effect.

A table summarizing the results obtained in the estimation \textcolor{idcolor}{confounded identifiable} causal queries are presented in \cref{tab:metrics_ecoli_both}. The statistics have been computed in the same way as in Sachs' dataset. In the case of ATE and CF error, they have been computed only on the \textit{direct} confounded identifiable paths, \ie, the green paths in \cref{fig:ecoli_graph_custom}.

\begin{table*}[t]
\caption{Performance metrics on Ecoli70 dataset. ATE and CF error statistics computed aggregating all causal queries and 5 runs. Intervened and evaluated on the direct \textcolor{idcolor}{confounded identifiable} causal effects of \cref{fig:ecoli_graph_custom}. Bold indicates significantly better results ($95\%$ CI from a Mann-Whitney U test). Lower error values indicate better performance.}
\label{tab:metrics_ecoli_both}
\scriptsize
\setlength{\tabcolsep}{2pt}
\centering
\begin{tabular}{clcccccccc}
\toprule
& & \multicolumn{4}{c}{Additive} & \multicolumn{4}{c}{Non-additive} \\
\cmidrule(lr){3-6} \cmidrule(lr){7-10}
& Model & MMD obs & MMD int & $|$ATE err$|$ & $|$CF err$|$ & MMD obs & MMD int & $|$ATE err$|$ & $|$CF err$|$ \\
& & $\times$$10^{4}$ & $\times$$10^{4}$ & $\times$$10^{2}$ & $\times$$10^{2}$ & $\times$$10^{4}$ & $\times$$10^{4}$ & $\times$$10^{2}$ & $\times$$10^{2}$ \\
\midrule
{ Oracle  } & CNF & $2.34_{0.62}$ & $6.05_{5.28}$& $5.04_{7.42}$ &  $9.91_{12.46}$ &
$1.49_{0.57}$ & $4.05_{8.22}$& $3.51_{4.84}$ &  $1.67_{1.64}$\\
\midrule
\multirow{2}{*}{Aware} 
& \ours & $2.42_{0.82}$ & $\mathbf{7.04_{3.87}}$& $\mathbf{4.49_{6.76}}$ &  $\mathbf{12.95_{8.00}}$ &
$1.58_{0.65}$ & $9.22_{22.38}$& $\mathbf{8.79_{17.91}}$ &  $2.15_{2.10}$ \\
& Deconfounder & -- & $-$& $27.35_{26.17}$ &  $82.15_{116.90}$ &
-- & $-$& $30.00_{33.24}$ &  $9.90_{9.47}$ \\
\midrule
\multirow{3}{*}{Unaware} 
& CNF & $2.98_{1.15}$ & $10.25_{12.13}$& $23.91_{25.16}$ &  $34.02_{23.90}$ &
$1.95_{0.77}$ & $10.20_{20.87}$& $12.72_{19.21}$ &  $2.45_{2.06}$  \\
& ANM & $32.80_{2.81}$ & $44.33_{17.62}$& $21.88_{23.89}$ &  $31.33_{20.64}$  &
 $13.17_{3.95}$ & $27.56_{31.57}$& $15.04_{18.18}$ &  $2.71_{1.88}$\\
& DCM & $31.65_{0.27}$& $49.50_{36.83}$& $24.45_{33.31}$ &  $30.22_{24.83}$ &
$18.78_{6.01}$ & $33.37_{36.14}$& $15.07_{22.37}$ &  $2.36_{2.08}$ \\
\bottomrule
\end{tabular}
\end{table*}

\ours significantly outperforms the baselines in ATE and counterfactual estimation in the additive setting and in ATE estimation in the non-additive setting. The MMD differences, both observational and interventional, are negligible between the \textit{oracle}, \ours, and CNF, likely due to the high number of variables diluting estimation bias. Counterfactual differences in the non-additive setting are also insignificant. However, compared to the \textit{oracle}, the gap between the \textit{oracle} and \textit{unaware} CGMs is smaller than in the additive case. While \ours reaches an intermediate point, the difference remains insignificant.

\subsubsection{Comment on the deconfounder results}

One may realize that the errors committed by the deconfounder of \citep{wang2019blessings, wang2021proxy} are greater than those from unaware models. First, we want to underline that, although the deconfounder allows us to predict counterfactual queries, the algorithm does not present any guarantees of a correct counterfactual estimation since it does not model the exogenous variables of the SCM. We hypothesize this to be the reason behind its performance in counterfactual estimation.

Moreover, let us explain some of the other paths where the errors of the deconfounder are greater than for unaware models. In Sachs' dataset, to model the causal effect \methodname{Ekt}$\rightarrow$\methodname{Akt}, the factorization model of the deconfounder uses \methodname{Raf, Mek, Jnk} and \methodname{P38} to extract the substitute confounder; the factorization model assumes that all those variables are independent conditioned to $\pred z$, while that is not the case in the true SCM and, therefore, this SCM violates the independence assumption of \citep{wang2019blessings}. The same argument is valid for the paths \methodname{yceP}$\rightarrow$\methodname{yfaD}, \methodname{lacA}$\rightarrow$\methodname{yaeM}, \methodname{yceP}$\rightarrow$\methodname{yfaD}, \methodname{ydeE}$\rightarrow$\methodname{pspA} and \methodname{pspB}$\rightarrow$\methodname{pspA}.

On the other hand, the paths \methodname{lacZ}$\rightarrow$\methodname{yaeM}, \methodname{asnA}$\rightarrow$\methodname{lacY} are frontdoor paths that \ours can identify because it models the hidden confounder following the true causal graph. However, the deconfounder is not designed to model this paths. To evaluate its performance for frontdoor paths, deconfounder uses the same variables as \ours to extract the substitute of the confounder. However, the deconfounder assumes independence conditioned to the substitute confounder and that is not the case; therefore, we are violating the independence assumption again.

\begin{wrapfigure}[11]{R}{0.5\linewidth}
    \centering
    \includegraphics[width=.8\linewidth]{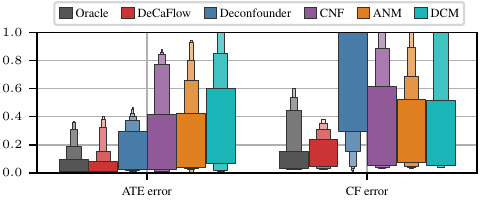}
    \caption{ATE and CF error evaluating only links where the deconfounder should work in the additive case.}
    \label{fig:deconfounder_links}
\end{wrapfigure}

The only two paths that meet the deconfounder assumptions in \cref{fig:ecoli_graph_custom} are \methodname{lacA}$\rightarrow$\methodname{lacY} and  \methodname{yedE}$\rightarrow$\methodname{pspB}. In consequence, we can observe in \cref{fig:deconfounder_links} that in those paths, the deconfounder performs at least as well as unaware methods. On the other hand, all the factor models used for the deconfounder implementation (PPCA, Deep exponential families and Variational autoencoder) assume additive noise. Therefore, interventional distributions in non-additive settings are not computable theoretically with these models.

\begin{table}[t]
\caption{Performance metrics on Ecoli70 dataset. Statistics computed an all samples over 5 runs, intervening and evaluating only in the causal effects that deconfounder should solve. Bold indicates significantly better results ($95\%$ CI from a Mann-Whitney U test). Lower error values indicate better performance.}
\label{tab:metrics_dataset}
\small
\setlength{\tabcolsep}{5pt}
\centering
\begin{tabular}{clcc}
\toprule
& Model & $|$ATE err$|$ $\times$$10^{2}$& $|$CF err$|$ $\times$$10^{1}$  \\
\midrule
{ Oracle } &  CNF & $8.31_{10.95}$ & $1.49_{1.86}$ \\
\midrule
\multirow{2}{*}{Aware} 
& \ours & $\mathbf{7.78_{7.30}}$ & $\mathbf{1.87_{1.50}}$ \\
& Deconfounder & $14.35_{15.24}$ & $12.03_{15.81}$ \\
\midrule
\multirow{3}{*}{Unaware} 
& CNF & $27.82_{30.17}$ & $4.01_{3.62}$ \\
& ANM & $27.63_{29.74}$ & $3.64_{3.15}$ \\
& DCM & $42.45_{54.23}$ & $4.08_{4.12}$ \\
\bottomrule
\end{tabular}
\end{table}

\subsubsection{Metrics on the other paths}
\begin{wrapfigure}[14]{R}{0.5\linewidth}
    \centering
    \vspace{-\baselineskip}
    \includegraphics[width=\linewidth]{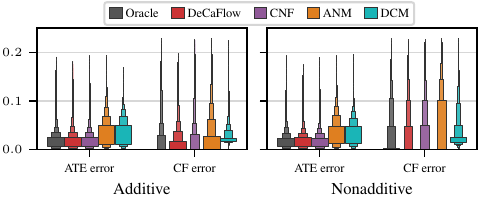}
    \caption{ Error boxenplots on the Ecoli70 dataset for different CGMs, averaged over all \textit{unconfounded} direct effects (see \cref{fig:ecoli_graph_custom}) after intervening in their \nth{25}, \nth{50}, and \nth{75} percentiles and 5 random realizations of the experiment.}
    \label{fig:black_paths}
\end{wrapfigure}
In this subsection we include a comparison between all the models in the \textit{unconfounded} and the \textcolor{secondcolor}{unidentifiable} effects. For \textit{unconfounded effects}, our expectation is to observe that all the CGMs achieve a performance comparable with the \textit{oracle}. On the other hand, we expect to have higher errors in \textcolor{secondcolor}{unidentifiable effects}, since we do not have theoretical guarantees.

\paragraph{Unconfounded Effects.}

The results for \textit{unconfounded effects} are summarized in \cref{fig:black_paths} and \cref{tab:black_paths}, considering only direct effects for ATE and counterfactual error computations. As expected, \ours and CNF achieve metrics comparable to the \textit{oracle} in both ATE and counterfactual estimations, particularly evident in \cref{fig:black_paths}, where error distributions are nearly identical. \ref{tab:black_paths} does not show statistically significative differences between \ours and CNF. Notably, architectures based on causal normalizing flows outperform ANM and DCM, which model each causal mechanism, $f_i$, with separate networks. This difference is crucial in settings with many variables and complex relations, where scalability is essential. Unlike ANM and DCM, which suffer from error propagation and limited scalability, causal normalizing flows leverage a single amortized model, making them more efficient in high-dimensional scenarios.

Finally, note that the deconfounder has not been included in these metrics because it is not designed for \textit{unconfounded queries} and there are many queries, while one deconfounder model is needed for each query.

\begin{table}[t]
    \caption{Performance metrics on Ecoli70 dataset. Statistics computed on all \textit{unconfounded} direct effects and 5 runs. Bold indicates significantly better results ($95\%$ CI from a Mann-Whitney U test). Lower error values indicate better performance.}
    \label{tab:black_paths}
    \small
    \setlength{\tabcolsep}{2pt}
    \centering
    \begin{tabular}{clcccccccc}
    \toprule
    & & \multicolumn{3}{c}{Additive} & \multicolumn{3}{c}{Non-additive} \\
    \cmidrule(lr){3-5} \cmidrule(lr){6-8}
    & Model & MMD int & $|$ATE err$|$ & $|$CF err$|$ & MMD int & $|$ATE err$|$ & $|$CF err$|$ \\
    & & $\times$$10^{4}$ & $\times$$10^{2}$ & $\times$$10^{2}$ & $\times$$10^{4}$ & $\times$$10^{2}$ & $\times$$10^{2}$ \\
    \midrule
    { Oracle } & CNF & $3.72_{3.73}$ & $2.00_{2.27}$ & $1.27_{3.49}$ & $1.94_{2.96}$ & $1.92_{1.99}$ & $1.76_{4.10}$ \\
    \midrule
    { Aware } & \ours & $4.53_{4.98}$& $2.00_{2.07}$ &  $1.31_{2.93}$ & $2.83_{6.36}$& $1.93_{1.95}$ &  $1.62_{3.87}$ \\
    \midrule
    \multirow{3}{*}{Unaware} 
    & CNF & $4.77_{6.09}$ & $2.02_{2.21}$ & $1.22_{3.18}$ & $2.97_{7.64}$ & $1.95_{1.92}$ & $1.71_{3.93}$ \\
    & ANM & $34.72_{8.56}$ & $3.57_{3.02}$ & $2.02_{4.09}$ & $15.13_{12.57}$ & $3.53_{3.15}$ & $2.64_{5.34}$ \\
    & DCM & $36.23_{14.29}$ & $3.48_{2.75}$ & $2.69_{2.30}$ & $21.22_{13.68}$ & $3.42_{2.63}$ & $3.00_{3.42}$ \\
    \bottomrule
    \end{tabular}
\end{table}

\paragraph{Unidentifiable Effects.}

\begin{figure}[t]
    \centering
    \includegraphics[width=.5\linewidth]{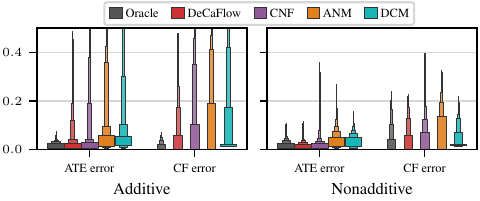}
    \caption{Error boxenplots on the Ecoli70 dataset for different CGMs, averaged over all \textcolor{secondcolor}{unidentifiable} direct effects (see \cref{fig:ecoli_graph_custom}) after intervening in their \nth{25}, \nth{50}, and \nth{75} percentiles and 5 random realizations of the experiment.}
    \label{fig:red-paths}
\end{figure}
The results for \textcolor{secondcolor}{unidentifiable effects}—causal queries that violate the assumptions in \cref{sec:theoretica_results}—are summarized in \cref{fig:red-paths} and \cref{tab:red-paths}. Notably, the \textit{oracle} performs significantly better than the other CGMs. As seen in \cref{fig:red-paths}, error distributions are highly skewed, with ATE and counterfactual errors reaching extreme values---considering that metrics are computed on the standardized variables. \cref{tab:red-paths} shows no significant differences between the metrics achieved by \ours and CNF.

\begin{table}[t]
\caption{Performance metrics on Ecoli70 dataset. Statistics computed on all \textcolor{secondcolor}{unidentifiable} direct effects and 5 runs. Bold indicates significantly better results ($95\%$ CI from a Mann-Whitney U test). Lower error values indicate better performance}
\label{tab:red-paths}
\small
\setlength{\tabcolsep}{2pt}
\centering
\begin{tabular}{clcccccccc}
\toprule
& & \multicolumn{3}{c}{Additive} & \multicolumn{3}{c}{Non-additive} \\
\cmidrule(lr){3-5} \cmidrule(lr){6-8}
& Model & MMD int & $|$ATE err$|$ & $|$CF err$|$ & MMD int & $|$ATE err$|$ & $|$CF err$|$ \\
& & $\times$$10^{4}$ & $\times$$10^{2}$ & $\times$$10^{3}$ & $\times$$10^{5}$ & $\times$$10^{2}$ & $\times$$10^{2}$ \\
\midrule
{ Oracle } &  CNF & $3.71_{3.52}$ & $1.79_{1.36}$ & $5.88_{15.16}$ & $16.98_{6.87}$ & $1.75_{1.59}$ & $1.62_{4.57}$ \\
\midrule
{Aware} 
& \ours & $3.80_{3.61}$& $3.95_{7.89}$ &  $33.62_{80.37}$ & $23.02_{21.96}$& $1.75_{1.66}$ &  $1.88_{4.97}$ \\
\midrule
\multirow{3}{*}{Unaware} 
& CNF & $4.54_{4.81}$ & $4.75_{10.65}$ & $44.76_{126.36}$ & $20.22_{6.68}$ & $2.32_{3.80}$ & $2.13_{6.25}$ \\
& ANM & $34.38_{5.17}$ & $7.43_{12.64}$ & $52.70_{137.99}$ & $130.71_{41.64}$ & $4.01_{3.82}$ & $2.93_{7.21}$ \\
& DCM & $35.49_{4.95}$ & $7.67_{13.93}$ & $67.46_{132.21}$ & $198.23_{58.62}$ & $3.43_{2.76}$ & $3.29_{3.92}$ \\
\bottomrule
\end{tabular}
\end{table}

\subsubsection{Hyper-parameters and splits}
\label{app:sec:hyperparameters}
We have performed a hyperparameter grid search over \emph{validation} data in both experiments on semi-synthetic datasets, exploring a large combination of hyperparameters for each model and dataset.

These are the parameters that were modified for each model:

\begin{itemize}
    \item CNF: the number of neurons and hidden layers of the single-layer flow, the type of flow (MAF, NSF). LR scheduler reducing on plateau and early stopping were applied with Adam optimizer \citep{kingma2014adam}.
    \item \ours: number of neurons and hidden layers of the single-layer causal flow (generative network), type of generative network architecture (MAF, NSF), number of neurons and hidden layers of the single-layer encoder flow (inference network), type of encoder architecture (MAF, NSF), KL regularization (True, False). LR scheduler reducing on plateau and early stopping was applied with the Adam optimizer \citep{kingma2014adam}.
    \item Deconfounder: type of factorization model (PPCA, VAE, Deep Exponential Families), number of neurons and hidden layers (in case of deep models), type of outcome model (MLP, random forest, linear regression), number of neurons and hidden layers of the outcome model (in case of deep models).
    \item DCM: number of neurons and hidden layers of each network, learning rate and number of iterations (we have not introduced early stopping or learning rate scheduler). The rest of hyperparameters were selected to the default value in the original code.
    \item ANM: an automatic search was performed across several models in the original DCM code. This search is performed with the DoWhy package \citep{dowhy_gcm}.
\end{itemize}

The selection was based on the matching of the observational for the causal generative models, using MMD. In the deconfounder, the factorization networks were selected by the likelihood of the observed variables and the outcome models with maximum likelihood. 

Although including all hyperparameters would be very extensive, we give here a sample of the hyperparameters selected for \ours in the Ecoli70 additive dataset:

\begin{itemize}
    \item Hidden neurons of causal flow (generative network): $3\times 128$
    \item Type of causal flow (generative network): neural spline flow (NSF) \citep{durkan2019neural}.
    \item Hidden neurons of encoder flow (inference network): $3\times 64$
    \item Type of normalizing flow (inference network): neural spline flow (NSF) \citep{durkan2019neural}.
    \item Regularize: True (warm-up: 30 epochs)
    \item Total number of parameters: 182k.
\end{itemize}

Both experiments were performed with 25,000 data, split into $80\%, 10\%, 10\%$ (train, validation, and test). All metrics are given over the \emph{test} dataset, and hyperparameter search was performed over the \emph{validation} dataset.

\subsubsection{Processing times}

All the experiments were conducted on CPU. Although the experiments were carried out on a cluster of different CPU, we include here two tables for the two semi-synthetic datasets (\cref{tab:timing_results_ecoli} and \cref{tab:timing_results_sachs}) with the processing times measured in a CPU Intel(R) Core(TM) i7-13650HX laptop, just to show that even in a laptop CPU, the training and inference times are sensible even for large datasets as the Ecoli70 dataset.

\begin{table}[t]
\centering
\caption{Computation times per model across training and evaluation regimes for Ecoli70  additive dataset. Mean and standard deviation of the training and inference time over 100 epochs in training and over 7 interventions in inference.}
\label{tab:timing_results_ecoli}
\resizebox{\linewidth}{!}{\begin{tabular}{lccc}
\toprule
Model & Epoch Tr. [s] (20000 samples) & Interventional [s] (2500 samples) & CF [s] (2500 samples) \\
\midrule
Oracle   & $0.64_{0.06}$ & $0.30_{0.02}$ & $0.36_{0.03}$ \\
\ours & $0.98_{0.10}$ & $0.28_{0.02}$ & $0.35_{0.04}$ \\
CNF      & $0.60_{0.07}$ & $0.26_{0.01}$ & $0.32_{0.05}$ \\
\bottomrule
\end{tabular}}
\end{table}

\begin{table}[t]
\centering
\caption{Computation times on the Sachs' Additive Dataset. Mean and standard deviation of the training and inference time over 100 epochs in training and over 3 interventions in inference.}
\label{tab:timing_results_sachs}
\resizebox{\linewidth}{!}{
\begin{tabular}{lccc}
\toprule
Model & Epoch Tr. [s] (20000 samples) & Interventional [s] (2500 samples) & CF [s] (2500 samples) \\
\midrule
Oracle   & $0.32_{0.06}$ & $0.08_{0.001}$ & $0.102_{0.010}$ \\
\ours & $0.75_{0.12}$ & $0.05_{0.004}$ & $0.086_{0.005}$ \\
CNF      & $0.33_{0.06}$ & $0.048_{0.003}$ & $0.065_{0.006}$ \\
\bottomrule
\end{tabular}
}
\end{table}

Note that \ours takes more time in training. This is because the network is more complex, due to the inference network, and that we have to sample from the posterior distribution. However, the difference in inference is not that relevant. In fact, \ours takes less time than the oracle in inference, even when they are sampling the same number of variables (hidden confounders + observed variables). The unaware causal normalizing flow (CNF) only samples from the observed variables. That is why the inference time is lower.

\subsection{Law school fairness use-case}
\label{app:sec:additional-results-fairness}

The experiment with real-world data was inspired by \citet{kusnerCounterfactualFairness2017} and \citet{javaloy2024causal}.
The goal is to find a fair estimator of the decile of the grades each student will occupy in their third year of university.

The dataset contains information on \num{27000} law students who were admitted by the Law School Admissions Council (LSAC) from 1991 to 1997. We have performed an experiment similar to that carried out by \citet{kusnerCounterfactualFairness2017}, where race and sex were treated as sensitive attributes. We have considered the following variables to include in our study:

\begin{itemize}
    \item \methodname{Race}: binary indicator of the race that distinguish between white and non-white.
    \item \methodname{Sex}: binary indicator of the sex that distinguish between male and female.
    \item \methodname{Fam}: family income.
    \item \methodname{LSAT}: the grade achieved in the Law School Admission Test (LSAT).
    \item \methodname{UGPA}: the undergraduate grade point average (GPA) of the student previous to the admission.
    \item \methodname{FYA}: first-year average grade.
    \item \methodname{Decile3}: the decile of the grades in the third year of university. This is the variable to predict.
\end{itemize}

We consider that an estimator $\hat{\varoutcome}$ is fair if it meets \textit{Demographic parity}, defined as follows~\citep[Def. 3]{kusnerCounterfactualFairness2017}: A predictor $\hat{\varoutcome}$ satisfies demographic parity if the predicted distributions for different values of a sensitive attribute are equal: $p(\hat{\varoutcome}\mid \vartreat=0)=p(\hat{\varoutcome}\mid \vartreat=1)$. We evaluate the difference between predicted distributions using Maximum Mean Discrepancy (MMD)~\citep{gretton2006kernel}, where a lower distance between the predictions of two sensitive groups denotes a fairer predictor.

The assumed causal graph is slightly different from that of \citet{kusnerCounterfactualFairness2017}, since their purpose is to make a fair prediction \methodname{FYA} accounting only for \methodname{Race}, \methodname{Sex}, \methodname{LSAT} and \methodname{UGPA}. However, we include \methodname{Fam} and \methodname{FYA} as predictors and the task is to predict \methodname{Decile3} and the assumed causal graph is the one of \cref{fig:fairness-causal-graph}. 

\paragraph{Proposed fair predictor with \ours.} We propose to model the confounded SCM presented in \cref{fig:fairness_scm_decaf}, where are explicitly shown the exogenous variables, that are independent of the other variables of the graph except of their associated endogenous variable. 

Afterwards, we predict the outcome, \methodname{Decile3} from the extracted latent variable that acts as substitute of the \methodname{knowledge} and the exogenous variables of \methodname{FYA} and \methodname{Fam}, following the causal graph of \cref{fig:fairness-causal-graph}, using a gradient-boosted decision tree~\citep{friedman2001greedy}: $\pred p(\methodname{Decile3}\mid \varexo_\methodname{FI}, \varexo_\methodname{FYA}, \varhidden)$. \ours models \varhidden and the exogenous variables as independent from \methodname{Race} and \methodname{Sex}. Therefore, the prediction of \methodname{Decile3} should be fair.

\begin{figure}
    \centering
        \begin{tikzcd}[
        cells={nodes={fairnessobs}},
        column sep=small, 
        row sep=small,
        ]
        |[fairnesshidden]| \varexo_{\methodname{S}}\arrow[r, hiddenarrow] &\methodname{Sex}
        \arrow[r] \arrow[dr] \arrow[ddr] & \methodname{GPA} & |[fairnesshidden]| \varexo_\methodname{G} \arrow[l, hiddenarrow] &\\
        |[fairnesshidden]| \varexo_{\methodname{R}}\arrow[r, hiddenarrow] &\methodname{Race} \arrow[d] \arrow[ur] \arrow[r] \arrow[dr] & \methodname{LSAT} & |[fairnesshidden]| \varexo_\methodname{L} \arrow[l, hiddenarrow] &|[fairnesshidden]| \varhidden \arrow[ull, hiddenarrow] \arrow[ll, hiddenarrow, bend right=18] \arrow[dll, hiddenarrow]    \\
        |[fairnesshidden]| \varexo_{\methodname{FI}}\arrow[r, hiddenarrow] &\methodname{Fam} \arrow[uur] \arrow[ur] \arrow[r] & \methodname{FYA} & |[fairnesshidden]| \varexo_\methodname{FYA} \arrow[l, hiddenarrow] & \\
    \end{tikzcd}
    \caption{Confounded SCM modeled by \ours.}
    \label{fig:fairness_scm_decaf}
\end{figure}

\paragraph{Baselines.} We consider as baselines the methods \textit{Fair K} and \textit{Fair add} proposed by \citet{kusnerCounterfactualFairness2017}.

\textit{Fair K} is a fair predictor categorized in Level 2 in \citet{kusnerCounterfactualFairness2017}, which postulates that the student's knowledge, \methodname{know} affects \methodname{GPA}, \methodname{LSAT}, \methodname{FYA} and \methodname{Decile 3}, following the distributions described below.
\begin{equation}
\begin{aligned}
    \methodname{Fam} &\sim \mathcal{N} \left( b_{Fam} + w_{Fam}^R \methodname{Race}, 1 \right), \\
    \methodname{GPA} &\sim \mathcal{N} \left( b_G + w_G^K \methodname{know} + w_G^R \methodname{Race} + w_G^S \methodname{Sex} + w_G^{Fam} \methodname{Fam}, \sigma_G^2 \right), \\
    \methodname{LSAT} &\sim \text{Poisson} \left( \exp(b_L + w_L^K  \methodname{know} + w_L^R \methodname{Race} + w_L^S \methodname{Sex} + w_L^{Fam} \methodname{Fam}) \right), \\
    \methodname{FYA} &\sim \mathcal{N} \left( w_F^K  \methodname{know} + w_F^R \methodname{Race} + w_F^S \methodname{Sex} + w_F^{Fam} \methodname{Fam}, 1 \right), \\
    \methodname{Decile3} &\sim \text{Poisson} \left( \exp(w_D^K  \methodname{know} + w_D^R  \methodname{Race} + w_D^S \methodname{Sex} + w_D^{Fam} \methodname{Fam}) \right), \\
    \methodname{know} &\sim \mathcal{N}(0,1).
\end{aligned}
\end{equation}
Then, the posterior distribution \methodname{know} is inferred using Monte Carlo with the probabilistic programming language Pyro \citep{bingham2019pyro}. The outcome is predicted using the inferred \methodname{know} using a gradient-boosted decision tree~\citep{friedman2001greedy}: $\pred p(\methodname{Decile3}\mid \methodname{know})$.

On the other hand, \textit{Fair Add} predicts the outcome from the residuals of predicting each variable with each parent, which guarantees that these residuals are independents of \methodname{Race} and \methodname{Sex}. That is, the predictor estimates the distribution $p(\methodname{Decile3} \mid \mathbf{r}_\methodname{Fam}, \mathbf{r}_\methodname{UGPA}, \mathbf{r}_\methodname{LSAT}, \mathbf{r}_\methodname{FYA})$, where these residuals are computed as:

\begin{equation}
    \begin{aligned}
        \mathbf{r}_{\methodname{Fam}} & = \methodname{Fam} - \E [\methodname{Fam} \mid \methodname{Sex}, \methodname{Race}] \\
        \mathbf{r}_{\methodname{UGPA}} & = \methodname{UGPA} - \E [\methodname{GPA} \mid \methodname{Sex}, \methodname{Race}, \methodname{Fam}] \\
        \mathbf{r}_{\methodname{LSAT}} & = \methodname{LSAT} - \E [\methodname{LSAT} \mid \methodname{Sex}, \methodname{Race}, \methodname{Fam}] \\
        \mathbf{r}_{\methodname{FYA}} & = \methodname{FYA} - \E [\methodname{FYA} \mid \methodname{Sex}, \methodname{Race}, \methodname{Fam}] \\
    \end{aligned}
\end{equation}

All predictors used are gradient-boosted decision trees~\citep{friedman2001greedy}.

\paragraph{Discussion of Results.} Although the \textit{fair} methods proposed by \citet{kusnerCounterfactualFairness2017} achieve significantly better \textit{demographic parity} than our approach using \ours (as indicated by a much lower MMD), their predictive performance is substantially inferior. Specifically, their performance is comparable to predicting the outcome using only the mean of the distribution, which serves as a baseline in \cref{tab:rmse_comparison}. In contrast, \ours achieves a $98\%$ reduction in MMD while incurring only an $11\%$ increase in RMSE, as illustrated in \cref{fig:fairness_dist}.

These experiments demonstrate that leveraging \ours to model confounded Structural Causal Models is beneficial beyond causal query estimation, leading to improved overall performance.

    \section{Implementation details} \label{app:sec:implementation}

\subsection{Posterior factorization of the deconfounding network}

\ours is capable of modeling confounded SCMs that contain several hidden confounders, $\varhidden = \{\varhidden_\indexthree\}_{\indexthree=1}^{\hiddensize}$, as in the Sachs' dataset (\cref{fig:sachs-graph}), Ecoli70 dataset (\cref{fig:ecoli_graph_custom}) or the Napkin graph (\cref{fig:napkin_graph}).  In such cases, the posterior over latent variables factorizes. We propose a factorized posterior in which each hidden confounder is conditioned on its children and on the parents of its children.

\begin{equation}
    \small
    q_\phi (\giventhat{\varhidden}{\varobs}) = 
     \prod_{\indexthree=1}^\hiddensize q_\phi\left(\varhidden_\indexthree \mid
     \text{pa}(\varhidden_\indexthree) 
     \cup
     \text{ch}(\varhidden_\indexthree)
    \cup
    \bigcup_{c \in \text{ch}(\varhidden_k)}
    \left( \text{pa}(c) \setminus \{\varhidden_\indextwo : \indextwo  \geq \indexthree\} \right) \right)
    \label{eq:factorization-encoder-expanded}
\end{equation}

Since we propose to use a conditional normalizing flow as the encoder, the dependencies between hidden confounders are modeled in an autoregressive manner. 
The rightmost part of the conditioning set accounts for collider-induced associations: conditioning on a child of $\varhidden_\indexthree$, $c$, makes $\varhidden_\indexthree$ dependent on other parents of $c$. Other parents of $c$ can also be hidden confounders. To model this, a causal ordering of the \varhidden components is assumed to avoid cycles in factorization, but it does not affect estimation, as collider associations have no inherent causal direction.

\subsection{Regularization of the Kullback-Leibler term in ELBO}

We propose the implementation of a warm-up adaptive regularization term that weights the contribution of the Kullback-Leibler term in the ELBO, to avoid posterior collapse \citep{vahdat2020nvae}.
During training, if the current epoch is lower than the predefined warm-up parameter, the KL term is weighted by $\beta$, which we define as $\beta = \min (1, \KLop{q_\phib(\giventhat{\varhidden}{\varobs})}{p(\varhidden)})$, as shown in \cref{alg:regularization}.

\begin{algorithm}[t]
\caption{KL regularization term in the training loop}
\label{alg:regularization}
\begin{algorithmic}[1]
\STATE \textbf{function} \textsc{ELBO computation}(epoch, warmup, $\theta, \phi$)
 \begin{ALC@g}
    \STATE \textbf{if} epoch $<$ warmup:
    \STATE $\;\;\; \ELBO(\phi, \theta) = \Expect[q_\phib]{\log p_\thetab(\giventhat{\varobs}{\varhidden})} - \beta \cdot \KLop{q_\phib(\giventhat{\varhidden}{\varobs})}{p(\varhidden)}$
    \STATE \textbf{else}:
    \STATE $\;\;\; \ELBO(\phi, \theta) = \Expect[q_\phib]{\log p_\thetab(\giventhat{\varobs}{\varhidden})} - \KLop{q_\phib(\giventhat{\varhidden}{\varobs})}{p(\varhidden)}$
    \STATE \textbf{return} \ELBO
\end{ALC@g}
\STATE \textbf{end function}
\end{algorithmic}
\end{algorithm}

In this way, we encourage the model to focus on data reconstruction on the first epochs, ignoring the KL term if the posterior is very similar to the prior, \ie, if $KL\approx 0$, then $\beta \approx 0$ and $\ELBO(\phi, \theta)\approx \Expect[q_\phib]{\log p_\thetab(\giventhat{\varobs}{\varhidden})}$. After the warm-up epoch, the loss is equivalent to the usual expression for the ELBO.
We have tested in the ablation study of \cref{app:sec:encoder_ablation} that the inclusion of the regularization term is useful in the Sachs' dataset. On the other hand, when posterior collapse does not occur, the $\beta$ term will be upper bounded by 1, therefore, not affecting the training process.

\subsection{Structural inductive bias} \label{sec:structural_example}
As presented in the original paper by \citet{javaloy2024causal}, the \textbf{adjacency matrix} that represents the causal graph is used to build the normalizing flow. In practice, this is implemented following the usual implementation of autoregressive normalizing flows using a Masked Autoencoder for Distribution Estimation (MADE) hypernetwork~\citep{germain2015made} that uses the causal graph for masking. In this case, we introduce the structural constrains between \itemi exogenous and endogenous variables and \itemii conditional variables and endogenous variables. 

As a result, our deconfounding network factorizes the posterior distribution as shown in \cref{eq:factorization-encoder}, modeling each hidden confounder as a function of its children, its parents and the parents of its children.
Similarly, the structural information in the generative network allows us to model each endogenous variable exclusively from its parents, whether these are other endogenous variables or hidden confounders, following \cref{eq:factorization-decoder}. 

We include in \cref{fig:scheme_decaf_napkin} an expanded version of \cref{fig:model} for the Napkin causal graph (\cref{fig:napkin_graph}), where it is shown in detail how its structural constraint is introduced in each conditional normalizing flow. %
Finally, note that the do-operator is inherited from the CNFs \citep{javaloy2024causal}, and details on its extension for \ours can be found in \cref{app:sec:do-operator}.

	\begin{figure}[t]
		\centering
		\includegraphics[width=0.8\linewidth, trim={7.2cm 0 4.6cm 0}, clip]{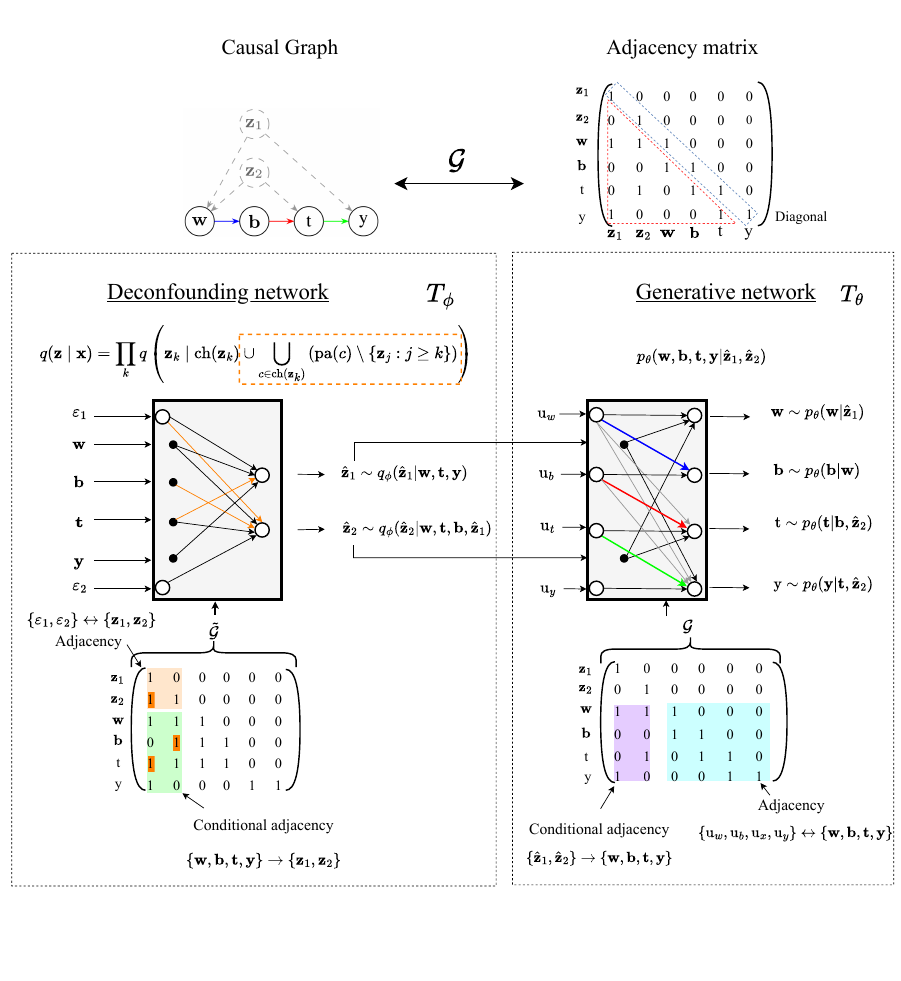}
		\caption{Complete illustration of \ours architecture, expanding \cref{fig:model} applied to the specific graph of \cref{fig:napkin_graph}. Both the deconfounding and the generative networks are conditional normalizing flows that factorize the distributions of the posterior and endogenous variables following \cref{eq:factorization-encoder} and \cref{eq:factorization-decoder}, respectively. Within these networks, functional dependencies are represented following the compacted version from \citet[Fig. 4(c)]{javaloy2024causal}. The \textcolor{orange}{orange} edges of the encoder corresponds to the collider association in the posterior factorization, and $\pred \graph$ encodes that associations.}
		\label{fig:scheme_decaf_napkin}
	\end{figure}

    \section{Do-operator \label{app:sec:do-operator}}

We introduce now the algorithms that \ours employ to generate interventional and counterfactual samples. 
First, we include those of \citet{javaloy2024causal}.
Note that the notation applied for \ours is slightly different from the that used for CNFs by \citet{javaloy2024causal}, naming the intervened variable as \vartreat, instead of $\ervx_i$, in order to be consistent with the notation used in \cref{sec:background,sec:theoretica_results}. %
\subsection{Do-operator in causal normalizing flows}

\begin{algorithm}[h]
\caption{Algorithm to sample from $P(\varobs \mid \doop(\ervx_i = \alpha))$. From \citet{javaloy2024causal}.}
\label{alg:interventional}
\begin{algorithmic}[1]
\STATE \textbf{function} \textsc{SampleIntervenedDist}$(i, \alpha)$
 \begin{ALC@g}
    \STATE $\varexo \sim P_{\varexo}$ \hfill $\triangleright$ Sample a value from the observational distribution.
    \STATE $\varobs \gets T_\theta^{-1}(\varexo)$
    \STATE $\ervx_i \gets \alpha$ \hfill $\triangleright$ Set $x_i$ to the intervened value $\alpha$.
    \STATE $\ervu_i \gets T_\theta(\varobs)_i$ \hfill $\triangleright$ Change the $i$-th value of $\varexo$.
    \STATE $\varobs \gets T_\theta^{-1}(\varexo)$
    \STATE \textbf{return} $\varobs$ \hfill $\triangleright$ Return the intervened sample.
\end{ALC@g}
\STATE \textbf{end function}

\end{algorithmic}
\end{algorithm}

Traditionally the computation of counterfactual samples follows the \textit{abduction, action and prediction} steps postulated by \citet{pearl19causalinferencestatistics}. The \textit{abduction} step consists of using the observations to determine the value of the exogenous variables. Then, the \textit{action} step computes the intervention, modifying the causal mechanism of the intervened variable and \textit{prediction} consist of using the exogenous variables and the modified SCM to compute the counterfactual. The computation of interventional samples follows a similar pattern, yet the exogenous values are directly sampled, \ie, skipping the abduction step. \Citet{javaloy2024causal} proposed an alternative implementation where, instead of modifying the causal mechanisms in the action step, the distribution of the exogenous variable associated with the intervened variable is changed instead, as described in \cref{alg:interventional,alg:counterfactual}.

\begin{algorithm}[h]
\caption{Algorithm to sample from $P(\cfactual{\varobs} \mid \doop(\ervx_i = \alpha), \factual{\varobs})$. From \citet{javaloy2024causal}.}
\label{alg:counterfactual}
\begin{algorithmic}[1]
\STATE \textbf{function} \textsc{GetCounterfactual}$(\factual{\varobs}, i, \alpha)$
 \begin{ALC@g}
    \STATE $\varexo \gets T_\theta(\factual{\varobs})$ \hfill $\triangleright$\textbf{Abduction:}  Get $\varexo$ from the factual sample.
    \STATE $\factual{\ervx_i} \gets \alpha$ \hfill $\triangleright$\textbf{Action:}  Set $x_i$ to the intervened value $\alpha$.
    \STATE $\ervu_i \gets T_\theta(\factual{\varobs})_i$ \hfill $\triangleright$\textbf{Action:}  Change the $i$-th value of $\varexo$.
    \STATE $\cfactual{\varobs} \gets T_\theta^{-1}(\varexo)$\hfill $\triangleright$\textbf{Prediction:} Get counterfactual
    \STATE \textbf{return} $\cfactual{\varobs}$ \hfill $\triangleright$ Return the counterfactual value.
    \end{ALC@g}
\STATE \textbf{end function}
\end{algorithmic}
\end{algorithm}

\subsection{Do-operator in interventional distributions with \ours}

The sampling process consists of first sampling from the prior distribution of the latent variables and from the exogenous distribution. Then, one can use the generative network (\flow) to generate interventional sampling, changing the components of \varexo associated with \vartreat as described in the previous section for CNFs.
Note that \varhidden is not an input of the normalizing flow, but a condition (or \textit{context}). Therefore, \varhidden is transformed neither in the forward nor reverse pass of the normalizing flow.
\begin{algorithm}[H]
\caption{Algorithm to sample from the interventional distribution, $P(\varobs \mid \doop(\vartreat = \alpha))$ with \ours.}
\label{alg:interventional_ours}
\begin{algorithmic}[1]
\STATE \textbf{function} \textsc{SampleIntervenedDist}$(\vartreat, \alpha)$
 \begin{ALC@g}
        \STATE $\varhidden \sim P_\varhidden$  \hfill $\triangleright$ Sample a value from the prior of \varhidden.
        \STATE $\varexo \sim P_{\varexo}$ \hfill $\triangleright$ Sample a value from the observational distribution.
        \STATE $\varobs \gets$ $\flowz^{-1}(\varexo)$
        \STATE $\vartreat \gets \alpha$ \hfill $\triangleright$ Set $\vartreat$ to the intervened value $\alpha$.
        \STATE $\ervu_\vartreat \gets$ $\flowz(\varobs)_\vartreat$ \hfill $\triangleright$ Change the component of \varexo associated with \vartreat.
        \STATE $\varobs \gets$ $\flowz^{-1}(\varexo)$
        \STATE \textbf{return} $\varobs$ \hfill $\triangleright$ Return the intervened sample.
    \end{ALC@g}
\STATE \textbf{end function}
\end{algorithmic}
\end{algorithm}

\begin{figure}[h]
    \centering
    \includegraphics[width=\linewidth, trim={2.95cm 0 4.8cm 0}, clip]{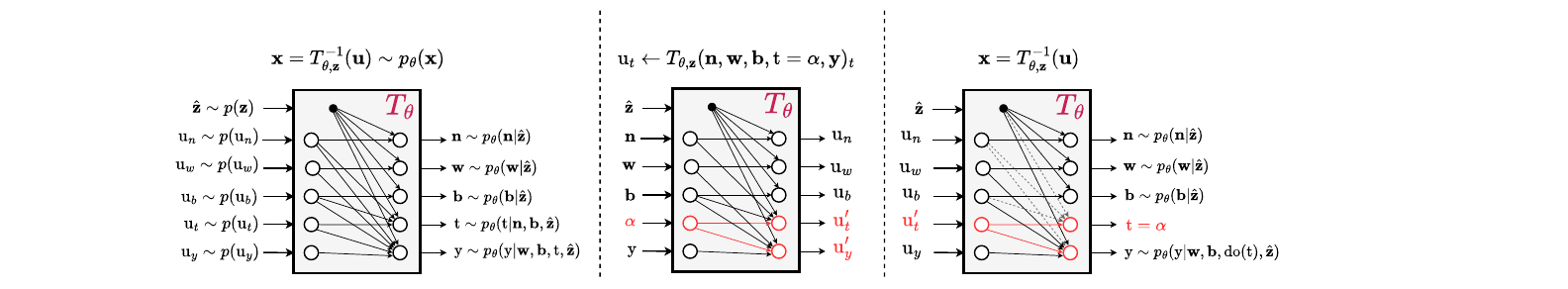}
    \caption{Schematic of the sampling process for an interventional distribution using the graph from \cref{fig:example-confounded-scm} and intervening in \vartreat. By sampling from the prior of the hidden confounders, $p(\varhidden)$, and the exogenous distribution, $p(\varexo)$, we obtain samples of the interventional distribution by appropriately setting $\ervu_\vartreat$, \ie, samples from $p_\theta(\giventhat{\varoutcome}{\doop(\vartreat)})$. Note that sampling from the interventional distribution only requires the generative network, \flow. Dashed gray arrows represent the cancellation of causal effects due to the intervention.}
    \label{fig:do-operator-interventional}
\end{figure}

As we can easily sample from interventional distributions, we compute the average treatment effect (ATE) via Monte Carlo. For example, to compute the ATE comparing two interventions ($\alpha_1, \alpha_2)$ in the variable \vartreat, we would generate samples of both interventional distributions, $p(\varobs \mid \doop(\vartreat = \alpha_1)), p(\varobs \mid \doop(\vartreat = \alpha_1))$, and approximate their expectations by taking the sample average:

\begin{align}
    \text{ATE}_\varobs(\alpha_1, \alpha_2) &= \E[\varobs \mid \doop(\vartreat = \alpha_2)] - \E[\varobs \mid \doop(\vartreat=\alpha_1)] \\
    &\approx \left(\frac{1}{N} \sum_{\varobs \sim \distribution(\giventhat{\varobs}{\doop(\vartreat = \alpha_2)})} \varobs \right) - \left( \frac{1}{N} \sum_{\varobs \sim \distribution(\giventhat{\varobs}{\doop(\vartreat = \alpha_1)})} \varobs \right)
\end{align}

If we were interested in the ATE of a subset of variables, \eg, \varoutcome, we would simply need to generate samples of \varobs and take only those from the variable of interest, \varoutcome.

\subsection{Do-operator in counterfactuals with \ours}

As part of the abduction step, our model estimates the posterior distribution of hidden confounders given a factual datapoint, $q_\phi(\varhidden \mid \factual{\varobs})$. Therefore, we can sample from the inferred posterior of the hidden confounders, and use those samples as the context for the generative network.

\begin{algorithm}[H]
\caption{Algorithm to sample from the counterfactual distribution, $P(\varobs \mid \doop(\vartreat = \alpha))$ with \ours.}
\label{alg:counterfactual_ours}
\begin{algorithmic}[1]
\STATE \textbf{function} \textsc{GetCounterfactual}$(\factual{\varobs}, \vartreat, \alpha)$
     \begin{ALC@g}
        \STATE $q_\phi (\varhidden \mid \factual{\varobs}) \gets$ Deconfounding network$(\factual{\varobs})$ \hfill $\triangleright$ \textbf{Abduction:} Get \varhidden from the factual sample.
        \STATE $\varhidden \sim q_\phi(\varhidden \mid \factual{\varobs})$\hfill $\triangleright$\textbf{Abduction}: Sample the posterior distribution.
        \STATE $\varexo \gets$ $\flowz(\factual{\varobs})$ \hfill $\triangleright$\textbf{Abduction:} Get $\varexo$ from the factual sample.
        \STATE $\factual{\vartreat} \gets \alpha$ \hfill $\triangleright$\textbf{Action:} Set $\vartreat$ to the intervened value $\alpha$.
        \STATE $\ervu_\vartreat \gets$ $\flowz(\factual{\varobs})_\vartreat$ \hfill $\triangleright$\textbf{Action:} Change the component of \varexo associated with \vartreat.
        \STATE $\cfactual{\varobs} \gets$ $\flowz^{-1}(\varexo)$ \hfill $\triangleright$ \textbf{Prediction:} compute the counterfactual
        \STATE \textbf{return} $\cfactual{\varobs}$ \hfill $\triangleright$ Return the counterfactual value.
    \end{ALC@g}
\STATE \textbf{end function}
\end{algorithmic}
\end{algorithm}

\begin{figure}[h]
    \centering
    \includegraphics[width=\linewidth, trim={3.1cm 0 3.1cm 0}, clip]{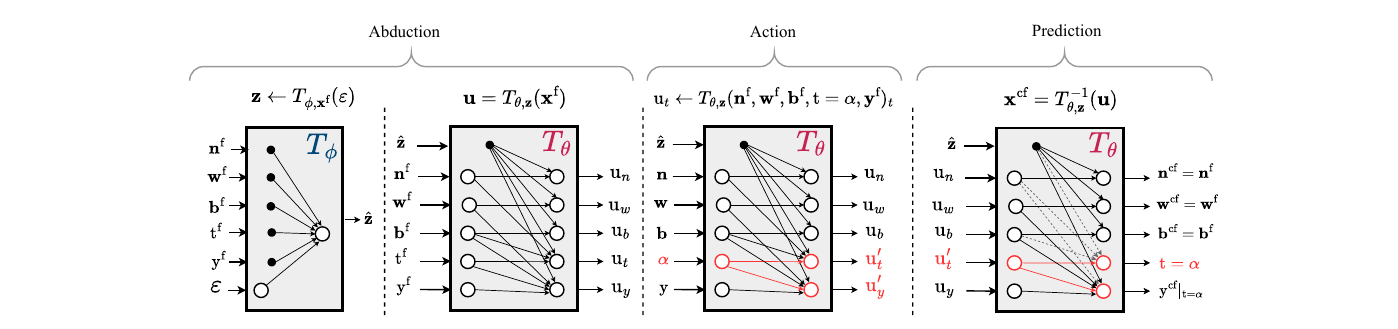}
    \caption{Schematic of the process of performing counterfactual inference with the causal graph from \cref{fig:example-confounded-scm} intervening in \vartreat. Both the deconfounding network, \encoder, and the generative network, \flow, are needed to generate counterfactual samples. Dashed gray arrows represent the cancellation of causal effect due to the intervention.}
    \label{fig:do-operator-counterfactual}
\end{figure}

    \section{Additional details on related work of causal inference with hidden confounders} \label{app:sec:related-work}

\subsection{Methods tailored to graph and query}
First, we want to remark that all the following methods are designed to address causal inference queries in specific causal graphs (or sub-graphs), therefore they can be used when these causal relationships hold. We summarize the causal graphs assumed by each work in \cref{fig:related_work_graphs}.
In the following, we assume the notation introduced in \cref{sec:background}, where \varhidden is the hidden confounder, \vartreat is the intervened variable (or treatment) and \varoutcome is the outcome, \ie, the variable where we want to evaluate the causal effects.

We have classified the different approaches depending on the graph that they are designed to address. However, there are two considerations that are common for all these approaches.
First, these methods follow a two-stage process: \itemi extracting a substitute of the unobserved confounder using variables affected by the confounder or instrumental variables, $ \pred \varhidden$, and \itemii estimating the outcome given this substitute, $\pred \varoutcome \sim p(\varoutcome\mid \pred \varhidden, \vartreat) $. In this case, one predictor must be trained per outcome, as well as one extractor per independent confounder.
Second, none of these methods estimate \textit{counterfactual distributions}, since they do not model exogenous variables.

\paragraph{Presence of null proxies independent of \vartreat (\cref{fig:one-proxy}).} We say \varproxyo to be a null proxy of \varhidden if it is a child of \varhidden independent of the outcome, \varoutcome, given \varhidden, \ie, $\varproxyo \indep \varoutcome \mid \varhidden$.
When null proxies of the confounder are available and they are also independent of the intervened variable, $\varproxyo \indep \vartreat \mid \varhidden$, these proxies can be used to infer a substitute of the hidden confounder. Among these works, \citet{allman09} and \citet{kuroki2014measurement} study the case in which the confounder is categorical, and use matrix factorization to extract a substitute when, either, there exist three Gaussian proxies \citep{allman09}, when the conditional distribution of the confounder given the proxy is known~\citep{kuroki2014measurement}, or when other proxies are available \citep{kuroki2014measurement}. \citet{kallus2018causal} also employ matrix factorization for cases where the confounder is continuous and the relation with the covariates and treatment (but not with the outcome) is linear. Similarly, \citet{kallus2019interval} uses kernel functions to extract the substitute confounder when the generators are nonlinear. The most relevant method based on deep generative methods is the one proposed by \citet{louizos2017causal}, where a variational autoencoder (VAE) is used to extract the substitute confounder when several null proxies are available, although no theoretical guarantees were provided and it was later shown to struggle in practice with complex distributions \citep{rissanen2021critical}. Finally, \citet{miao2023identifying} offer a regression-based approach to estimate the unobserved confounder under \textit{equivalence}, which assumes that any model of the joint achieves element-wise transformations of the latent variables, something that is not feasible to check: $\pred p(\vartreat, \varhidden \mid \varproxyo) = p(\vartreat, V(\varhidden) \mid \varproxyo)$.

\paragraph{Presence of two proxies: null and not null (\cref{fig:two-proxy}).} When the null proxies affect treatment (notice that in \cref{fig:two-proxy} the proxy $\varproxyo$ affects the treatment $\vartreat$), \citet{miao2018identifying} offer theoretic guarantees of causal identifiability in the presence of another proxy, $\varproxyt$, and completeness conditions. The proxy \varproxyt can be active, that is, it can directly affect \varoutcome. Then, \citet{tchetgen2020introduction} introduced the two-stage proximal least squares (P2SLS), which infers the substitute confounder from $p(\varproxyt \mid \vartreat, \varproxyo)$. P2SLS can be implemented using neural networks to achieve greater flexibility. 
Several works have followed-up the ideas introduced by \citet{miao2018identifying}, aiming to estimate the bridge function, \ie, finding an explicit form for the function $\tilde h$ shown in \cref{app:eq:relate-both-models}. For example, \citet{cui2024semiparametric} designed a doubly-robust estimator of the ATE by estimating the bridge function semiparametrically, and \citet{mastouri2021proximal} and \citet{kompa2022deep} applied moment restrictions to estimate the bridge function using deep neural networks. Other works have proposed multiple-robust methods when the confounders are categorical \citep{shi2020multiply}.

\paragraph{Instrumental variable (\cref{fig:instrumental_variable}).} Another condition that enables causal inference is the presence of instrumental variables (IVs), \ie variables that affect only the treatment and are independent of both the unobserved confounder and the outcome, given the treatment (in \cref{fig:instrumental_variable}, \varproxyo is an IV). In the linear case, \citet{pearl2009causality} and \citet{angrist2009mostly} demonstrated how a two-stage regression process can mitigate the confounding bias, as the only effect that occurs from the IV to the outcome is through the treatment variable. A substitute of the confounder is then extracted by computing the conditional distribution of the treatment given the IV, \ie, $\pred \varhidden \sim p(\vartreat \mid \varproxyo)$. Furthermore, \citet{hartford2017deep} extended this idea to include arbitrarily complex nonlinear data-generating processes, designing a two-step deep approach based on neural networks.

\paragraph{Multitreatment affected by a common confounder (\cref{fig:deconfounder}).} Finally, the multitreatment scenario has been studied by \citet{wang2019blessings} and \citet{ranganath2018multiple}, where it is called multitreatment since all covariates can be seen as treatments over the outcome variable, \varoutcome. Here, it is assumed that in the true causal model there exist several covariates that are independent given the unobserved confounder. Therefore, \citet{wang2019blessings} proposed to use a factorization model to infer the substitute confounder, such as probabilistic PCA or Poisson matrix factorization. In short, a factorization model assumes that the distribution of all the treatments factorizes as follows: $p(\mathbf{t}, \rvz) = p(\rvz)\prod_{i=1}^d p(\vartreat_i \mid \rvz)$, which should allow to construct a substitute of the confounder from the posterior of \varhidden: $\pred \varhidden \sim \pred p(\pred \varhidden \mid \mathbf{t})$.
Later, \citet{pmlr-v89-d-amour19a} provided counterexamples showing that the deconfounder does not achieve nonparametric identification without additional assumptions and, notably, one of the alternatives proposed by \citet{pmlr-v89-d-amour19a} highlights the use of proxy variables, which is the approach adopted by \ours.

Similar to \citet{wang2019blessings},  \citet{ranganath2018multiple} proposed to use a VAE as the factorization model, adding a regularization term to reduce the additional mutual information between the estimated confounder and the treatment $\vartreat_j$, given the rest of treatments, $\mathbf{t}_{-j}$. However, the theoretical guarantees of this approach require an infinite number of treatments to achieve unbiased estimates of the causal effects.
\citet{wang2021proxy} connect the ideas of \citet{miao2018identifying} and \citet{wang2019blessings} ensuring causal identification in the multitreatment setting when we know that some of the treatments \textit{can act as null proxies}, that is, when they do not affect the outcome. This assumption allows them to provide theoretical guarantees when the number of treatments does not tend to be infinite.  In despite of that, a factorization model such as the one \citet{wang2021posterior} propose can only model independent treatments given the hidden confounder, which greatly limits its practical utility.

\textbf{What is the relation of the deconfounder \citet{wang2019blessings, wang2021proxy} with \ours?} Similar to \ours, the deconfounder infers the posterior distribution of the confounder substitute from observational data using a generative model. However, the application of a factorization model restricts the structural dependencies that the it can model. For example, the deconfounder cannot model the structural dependencies of \cref{fig:two-proxy}, since the factorization model assumes $\varproxyo \indep \vartreat \indep \varproxyt \mid \varhidden$. In contrast, \ours leverages CNFs which can model these dependencies since the causal graph is encoded in the normalizing flow architecture.
It is also important to stress that \ours models the whole confounded SCM, including the exogenous variables. This allows us to compute \textit{counterfactuals} and train in a query-agnostic manner. In contrast, the deconfounder cannot compute counterfactuals and needs of a separate model per causal query.

\begin{figure}[H]
    \centering
    \begin{subfigure}{0.2\textwidth}
        \centering
            \begin{tikzcd}[
        cells={nodes={main node}}, column sep=tiny, row sep=small
        ]
        \varproxyo  & & |[hidden]|\varhidden \arrow[hidden, dl]  \arrow[hidden, ll] \arrow[hidden, dr]  &  \\
        & \vartreat \arrow[rr]  & & \varoutcome 
    \end{tikzcd}
        \caption{One proxy.}
        \label{fig:one-proxy}
    \end{subfigure}
    \begin{subfigure}{0.24\textwidth}
        \centering
            \begin{tikzcd}[
        cells={nodes={main node}}, column sep=tiny, row sep=small
        ]
        \varproxyo \arrow[dr] & & |[hidden]|\varhidden \arrow[hidden, ll] \arrow[hidden, rr] \arrow[hidden, dl] \arrow[hidden, dr]  & & \varproxyt \arrow[dl] \\
        & \vartreat \arrow[rr] & & \varoutcome 
    \end{tikzcd}
        \caption{Two proxies.}
        \label{fig:two-proxy}
    \end{subfigure}
    \begin{subfigure}{0.24\textwidth}
    \centering
    \includestandalone[width=0.7\linewidth]{figs/graphs/instrumental_variable}
    \caption{Instrumental variable.}
    \label{fig:instrumental_variable}
    \end{subfigure}
    \begin{subfigure}{0.3\textwidth}
        \centering
            \begin{tikzcd}[
        cells={nodes={main node}}, column sep=tiny, row sep=small
        ]
         & & |[hidden]|\varhidden \arrow[hidden, dd] \arrow[hidden, drr] \arrow[hidden, dr]
         \arrow[hidden, dl] \arrow[hidden, dll]\\
         \vartreat_1 \arrow[drr]
         & \vartreat_2 \arrow[dr] &
         & \vartreat_3 \arrow[dl]
         & \vartreat_4 \arrow[dll]\\
        &  & \varoutcome & & 
    \end{tikzcd}
        \caption{Multitreatment.}
        \label{fig:deconfounder}
    \end{subfigure}
    \caption{Graphs assumed by prior works. \captiona \citet{kuroki2014measurement, louizos2017causal, miao2023identifying, kallus2018causal, kallus2019interval, allman09} address the case where \varproxyo is independent of t. \captionb \citet{miao2018identifying} assumes the case where there exist two proxies. \captionc Graph with an instrumental variable. \captiond \citet{wang2019blessings, wang2021proxy, ranganath2018multiple} work with the multitreatment setting.}
    \label{fig:related_work_graphs}
\end{figure}

\subsection{CGM with unobserved confounders}

There exist several works that employ causal generative models (CGMs) in the presence of hidden confounders. We explain here the differences with our proposal, highlighting the practical advantages of \ours.

\paragraph{Neural Causal Models (NCMs).} \citet{xia2021causal} proposed a class of sequential causal generative models where each structural equation---\ie, the functional relationship between a variable and its parents in the causal graph---is modeled by a different neural network. The model is trained end-to-end to jointly learn all structural mechanisms. Beyond estimation, NCMs aim to determine whether a given causal query is identifiable from the data-generating process.
To assess identifiability, two NCMs are trained: one that maximizes the causal query, subject to a perfect observational fitting, and one that minimizes it. If both NCMs yield the same outcome, the query is deemed identifiable. Interestingly, this approach formalizes identifiability as an empirical condition based on optimization agreement.

However, the framework presents significant practical constraints: \itemi it only supports finite discrete variables, typically binary and low-dimensional, due to tractability constraints; \itemii it assumes that the true observational distribution is available for training; \itemiii two NCMs are trained per query, leading to high computational cost; and \itemiv identifiability is only revealed post-training, offering no guidance before the model is trained.
To perform counterfactual inference, \citet{xia2023neural} extended NCMs to estimate queries involving latent exogenous variables. However, their approach relies on rejection sampling to perform the abduction step, which is inefficient and unsuitable for continuous or high-dimensional settings, thus limiting its applicability in real-world scenarios.

In contrast, \ours addresses these limitations. First, we provide a principled criterion to estimate the identifiability of a query \emph{prior} to model training. Second, our framework supports continuous variables and scales to high-dimensional settings. Third, we train a single model that jointly estimates all causal mechanisms and enables efficient inference of counterfactual queries. Fourth, we use variational inference to approximate the posterior of hidden confounders, avoiding the inefficiency of rejection-based methods. Finally, we guarantee the identifiability of unconfounded exogenous variables (in the sense of \citet{pmlr-v206-xi23a}) by leveraging the theoretical framework of CNFs \citep{javaloy2024causal}. As a result, \ours is substantially more efficient and suited for  real-world applications.

\paragraph{Modular Causal Generative Models.} \citet{rahman2024modular} introduce a modular framework for high-dimensional causal inference, where variables influenced by the same hidden confounder are modeled jointly in end-to-end submodules. A key advantage of this approach is the ability to incorporate pretrained models into submodules, enabling flexible modeling of complex or structured variables when the modular criterion holds. The method supports continuous and discrete variables and uses adversarial training to match observational distributions. Symbolic identifiability is computed using the algorithm of \citet{jaber2022causal}, and they prove that identifiable queries remain estimable under their modular decomposition. However, the framework does not support counterfactual inference nor proximal learning, and it relies on adversarial optimization.

In comparison, \ours trains a single end-to-end model, estimates both observational and counterfactual distributions also in proximal settings, %
and enables efficient inference with broad applicability to real-world settings.

\paragraph{Counterfactual Identifiability of Bijective Causal Models.} \citet{nasr2023counterfactual} propose a sequential causal model using conditional normalizing flows to map exogenous to endogenous variables. The model focuses on counterfactual inference under backdoor and instrumental variable (IV) settings, with identifiability proven only for discrete variables. Proxy variables are not considered, and the use of invertible mappings over discrete domains makes theoretical claims less robust. Although the model claims support for continuous data, guarantees are restricted to discrete IV scenarios. Moreover, it does not model observational nor interventional distributions, and lacks parameter amortization due to its sequential structure.

In contrast, \ours supports continuous variables, models both observational and interventional distributions, and enables counterfactual inference under general confounding and proxy settings. It also requires a single end-to-end model and scales efficiently to real-world data.

\paragraph{Learning Functional Causal Models with Generative Neural Networks.} \citet{goudet2018learning} propose a method for causal discovery rather than causal inference under unobserved confounding. Given a Markov equivalence class (or graph skeleton), their approach uses generative neural networks to model each causal direction, selecting the graph that best matches the observational distribution evaluated via maximum mean discrepancy (MMD). The model is trained sequentially and assumes no hidden confounders.
While not directly comparable to our work, such causal discovery tools may serve as a preprocessing step when the causal graph is unknown, enabling downstream application of models---such as ours---that assume a known and correct structure.

    \section{Algorithms for causal query identification\label{sec:app:path_identifiability}}

As explained in \cref{subsec:cf-queries}, we can ask \ours to estimate any causal query, but we do not have the guarantee that the estimation \ours does is correct unless the query is identifiable. Therefore, we provide the practitioner with algorithms to check the identifiability of causal queries.

\paragraph{Specific treatment-outcome pair.}

We start presenting in \cref{alg:causal_query} an algorithm to identify a causal query, given a pair of treatment and outcome variables, which is valid for estimating the interventional distribution of the outcome, $p(\giventhat{\varoutcome}{\doop(\vartreat), \varcov})$, and the counterfactual one, $p(\giventhat{\cfactual{\varoutcome}}{\doop(\vartreat), \factual{\varobs}})$, as we postulated in \cref{sec:theoretica_results} that the latter is identifiable if the former is.

\begin{algorithm}[h]
\caption{Identification of causal queries that include intervention and outcome (\vartreat, \varoutcome)}
\label{alg:causal_query}
\begin{algorithmic}[1]
\REQUIRE Graph $\graph$, intervention variable $\vartreat$, outcome variable $\varoutcome$, covariates $\varcov$, hidden variables $\varhidden$
\ENSURE Boolean indicating if query is identifiable
\STATE $\varhidden \gets$ hidden variables that are parents of both $\vartreat$ and $\varoutcome$
\STATE \textbf{return} True \textbf{if} \varhidden is $\emptyset$  \hfill $\triangleright$ Unconfounded is identifiable
\FORALL{$\varhidden_\indexthree \in \varhidden$}
    \STATE \textbf{Comment:} Each $\varhidden_\indexthree$ is an independent component of $\varhidden$
    \STATE $\varproxyo$-proxies $\gets$ children of $\varhidden_\indexthree$ $d$-separated from $\vartreat$ given $(\varhidden, \varcov)$
    \STATE $\varproxyt$-proxies $ \gets$ children of $\varhidden_\indexthree$ $d$-separated from $\varoutcome$ given $(\varhidden, \varcov)$
    \IF{there exist $\varproxyo \in \varproxyo$-proxies and $\varproxyt \in \varproxyt$-proxies such that $\varproxyo$ is $d$-separated from $\varproxyt$ given $(\varhidden, \varcov)$}
        \STATE $\varhidden_\indexthree$ is deconfounded
    \ENDIF
\ENDFOR
\STATE \textbf{return} all $\varhidden_\indexthree$ are deconfounded
\end{algorithmic}
\end{algorithm}

We have employed \cref{alg:causal_query} on all direct paths of the Sachs and Ecoli70 datasets to check their identifiability, in order to get a visual representation of the queries that \ours can estimate in such complex graphs. %
If one is interested in evaluating a query which involves several outcomes, $\{\varoutcome_1, \varoutcome_2, \dots, \varoutcome_O \}$, one causal query per outcome variable should be evaluated.

\paragraph{Evaluation on all the variables.} Although \cref{alg:interventional_gen_identifiability} consists of iteratively applying \cref{alg:causal_query}, we also find it interesting to include the extension to identify causal queries evaluated on all variables in the dataset, which is useful for the case where we \ours as a generative model for the joint interventional distribution, $p(\varobs \mid \doop(\vartreat))$, or to generate joint counterfactual samples intervening in a specific variable, $\vartreat \subset\varobs$, $p(\cfactual{\varobs} \mid \doop(\vartreat), \factual{\varobs})$.

\begin{algorithm}[h]
\caption{Identification of causal queries, intervening in \vartreat and evaluating in all variables}
\label{alg:interventional_gen_identifiability}
\begin{algorithmic}[1]
\REQUIRE Graph $\graph$, intervention variable $\vartreat$, hidden variables $\varhidden$
\ENSURE Boolean indicating if the interventional distribution is identifiable 
\STATE $\varhidden \gets$ hidden variables that are parents of $\vartreat$
\FORALL{$\ervx_i \in$ descendants of $\vartreat$}
     \STATE \textbf{Comment:} Evaluate only on descendants of the intervention
    \STATE Check $(\vartreat, \ervx_i)$ identifiability with \cref{alg:causal_query}
\ENDFOR
\STATE \textbf{return} all $(\vartreat, \varobs_i)$ are identifiable
\end{algorithmic}
\end{algorithm}

\clearpage

\subsection{Pipeline for using \ours}

\begin{wrapfigure}{R}{0.4\linewidth}
    \centering
    \begin{tikzpicture}[node distance=2cm]

    \node (dataset) at (-4,5) {Dataset $\mathcal{D}$};
    \node (graph) at (0,5) {Graph $\mathcal{G}$};
    \node (queries) at (4,5) {Queries $\{\query_i\}_{i=1}^N$};

    \node (foreach) at (4,3.5)[hexa, align=center] {for each $\query_i$ do};

    \node (query) at (4,2) {Query $\query_i$};

    \node (train) at (-2,3.5) [process] {Train \ours};

    \node (trained) at (-2,0) {\ours trained};
    \node (check) at (2,0) [process, align=center]{{Check Query  identifiability} \\{\cref{alg:causal_query} and  \cref{alg:interventional_gen_identifiability}}};

    \node (decision) at (2,-2) [decision] {{\small Is $\query_i$ identifiable?}};

    \node (estimate) at (-2,-2) [process, align=center] {{Estimate $\query_i(\scm)$} \\ {\cref{alg:interventional_ours} and  \cref{alg:counterfactual_ours}}};
    \node (result) at (-2,-4) {$\hat{\query_i}(\scm)$};

    \node (fail) at (4.19,-4) {\huge \textbf{\textcolor{black}{\faFrown}}};

    \node (phantom) at (result.south) [yshift=-0.5cm] {}; 

    \begin{scope}[on background layer]
        \node[draw, dashed, rounded corners, fit=(query) (phantom) (estimate), label={[anchor=south east, xshift=5mm]below right:$N$}] (queryGroup) {};
    \end{scope}

    \draw [arrow] (dataset) -- (train);
    \draw [arrow] (graph) -- (train);
    \draw [arrow] (graph) -- (check);
    \draw [arrow] (query) -- (check);

    \draw[arrow] (queries) -- (foreach);
    \draw[arrow] (foreach) -- (query);

    \draw [arrow] (train) --(trained);
    \draw[arrow] (trained) -- (estimate);
    \draw [arrow] (check) -- (decision);

    \draw [arrow] (decision.east) -- ++(0.3,0) node[midway,above] {No} -- (fail);
    \draw [arrow] (decision.west) -- ++(-0.3,0) node[midway,above] {Yes}  -- (estimate); %

    \draw [arrow] (estimate) -- (result);

\end{tikzpicture}
    \caption{\textbf{Block diagram of our pipeline.}}
    \label{fig:pipeline}
\end{wrapfigure}
Our framework provides a systematic approach to estimating causal queries by integrating \ours, a model trained on observational data, with algorithms designed for query identifiability analysis.

As depicted in the pipeline, the framework takes as input a dataset $\gD$, a causal graph \graph, and a set of $N$ interesting queries $\{ \query_i\}_{i=1}^N$. The process begins by training \ours on $\gD$ and \graph, enabling it to learn the confounded SCM, \scm.

Simultaneously, the identifiability of each causal query $\query_i$ is assessed using dedicated algorithms (\cref{alg:causal_query} and \cref{alg:interventional_gen_identifiability}). If $\query_i$ is identifiable, the trained \ours is used to estimate $\query_i(\scm)$ (\cref{alg:interventional_ours} and \cref{alg:counterfactual_ours}), yielding the estimated causal effect $\hat{\query_i}(\scm)$. If $\query_i$ is not identifiable, the framework indicates that answering the query is not feasible given the available data and causal structure. Other causal queries can be answered by the model without retraining, provided that their identifiability is verified beforehand.

This workflow ensures a principled approach to causal inference, leveraging both data-driven modeling and theoretical guarantees on identifiability.

\paragraph{Validation with interventional data.} As a final step in the pipeline for real-world scenarios, especially in sensitive applications, we encourage practitioners to validate the framework with interventional data. Causal queries such as the \textit{average treatment effects} (ATEs) can be validated if a randomized experiment is available in which interventions are carried out on the treatment variable.

\end{document}